\PassOptionsToPackage{unicode=true}{hyperref} 
\PassOptionsToPackage{hyphens}{url}
\documentclass{article}

\usepackage{graphicx} 
\usepackage{titlesec} 

\usepackage{lmodern}

\usepackage{booktabs}
\usepackage{ltablex}
\usepackage{caption}
\usepackage{capt-of}
\usepackage{textcomp}
\usepackage{array}
\usepackage{ragged2e}
\usepackage{adjustbox}
\keepXColumns

\usepackage{placeins}
\usepackage{epstopdf}
\usepackage{listings} 
\usepackage{xcolor}   
\usepackage{longtable}
\usepackage[a4paper, margin=1in]{geometry} 
\usepackage{amssymb,amsmath}
\usepackage{ifxetex,ifluatex}
\usepackage[T1]{fontenc}
\usepackage[utf8]{inputenc}
\usepackage{textcomp}
\IfFileExists{upquote.sty}{\usepackage{upquote}}{}
\IfFileExists{microtype.sty}{%
\usepackage[]{microtype}
\UseMicrotypeSet[protrusion]{basicmath} 
}{}
\IfFileExists{parskip.sty}{%
\usepackage{parskip}
}{
\setlength{\parindent}{0pt}
\setlength{\parskip}{6pt plus 2pt minus 1pt}
}
\usepackage[colorlinks=true, linkcolor=blue, urlcolor=blue, citecolor=blue]{hyperref}
\hypersetup{
            pdfborder={0 0 0},
            breaklinks=true}
\usepackage{color}
\usepackage{fancyvrb}

\DefineVerbatimEnvironment{Highlighting}{Verbatim}{commandchars=\\\{\}}
\usepackage{framed}
\definecolor{shadecolor}{RGB}{42,33,28}
\newenvironment{Shaded}{\begin{snugshade}}{\end{snugshade}}

\newcommand{\AttributeTok}[1]{\textcolor[rgb]{0.74,0.68,0.62}{#1}}

\newcommand{\BuiltInTok}[1]{\textcolor[rgb]{0.74,0.68,0.62}{#1}}

\newcommand{\CommentTok}[1]{\textcolor[rgb]{0.00,0.40,1.00}{\textbf{\textit{#1}}}}

\newcommand{\DecValTok}[1]{\textcolor[rgb]{0.27,0.67,0.26}{#1}}

\newcommand{\ExtensionTok}[1]{\textcolor[rgb]{0.74,0.68,0.62}{#1}}
\newcommand{\FloatTok}[1]{\textcolor[rgb]{0.27,0.67,0.26}{#1}}
\newcommand{\FunctionTok}[1]{\textcolor[rgb]{1.00,0.58,0.35}{\textbf{#1}}}

\newcommand{\NormalTok}[1]{\textcolor[rgb]{0.74,0.68,0.62}{#1}}
\newcommand{\OperatorTok}[1]{\textcolor[rgb]{0.74,0.68,0.62}{#1}}

\newcommand{\StringTok}[1]{\textcolor[rgb]{0.02,0.61,0.04}{#1}}
\newcommand{\VariableTok}[1]{\textcolor[rgb]{0.74,0.68,0.62}{#1}}

\usepackage{longtable,booktabs}
\IfFileExists{footnote.sty}{\usepackage{footnote}\makesavenoteenv{longtable}}{}
\providecommand{\tightlist}{%
  \setlength{\itemsep}{0pt}\setlength{\parskip}{0pt}}
\ifx\paragraph\undefined\else
\let\oldparagraph\paragraph
\renewcommand{\paragraph}[1]{\oldparagraph{#1}\mbox{}}
\fi
\ifx\subparagraph\undefined\else
\let\oldsubparagraph\subparagraph
\renewcommand{\subparagraph}[1]{\oldsubparagraph{#1}\mbox{}}
\fi

\newcommand{\codeline}[1]{\texttt{\NormalTok{\$ }\BuiltInTok{\ExtensionTok{#1}}}}
\newenvironment{codeblock}
{
\begin{scriptsize}
\begin{Shaded}
}
{
\end{Shaded}
\end{scriptsize}
}

\makeatletter
\def\fps@figure{htbp}
\makeatother

\title{pySLAM: An Open-Source, Modular, and Extensible Framework for SLAM}
\author{Luigi Freda}
\date{\small \today}  

\begin{document}

\maketitle
\begin{center}
  \href{https://github.com/luigifreda/pyslam}{github.com/luigifreda/pyslam}
\end{center}

\begin{abstract}
pySLAM is an open-source framework for Visual SLAM that supports monocular, stereo, and RGB-D cameras. The framework integrates a broad range of classical and learning-based local features, multiple loop closure strategies, depth prediction models, semantic segmentation and mapping, a volumetric reconstruction pipeline, and a unified interface to feed-forward multi-view 3D reconstruction models. Multiple camera models are supported. Its sparse SLAM core is implemented in both Python and C++, which lets users trade flexibility for speed without changing the rest of the pipeline.
A unified dataset interface, automated evaluation scripts, and a regression test suite support reproducible experiments across more than fifteen dataset types. Designed for both beginners and experienced researchers, pySLAM emphasizes rapid prototyping, extensibility, and reproducibility, and encourages research that bridges traditional and deep learning-based approaches.
This document\footnote{You may find an updated version of this document at:\\ \href{https://github.com/luigifreda/pyslam/blob/master/docs/tex/document.pdf}{github.com/luigifreda/pyslam/blob/master/docs/tex/document.pdf}} presents the architecture of the framework, its main components, and the tools it provides for evaluation and testing.
\end{abstract}

\begin{figure}[htbp]
  \centering
  \includegraphics[width=0.5\textwidth]{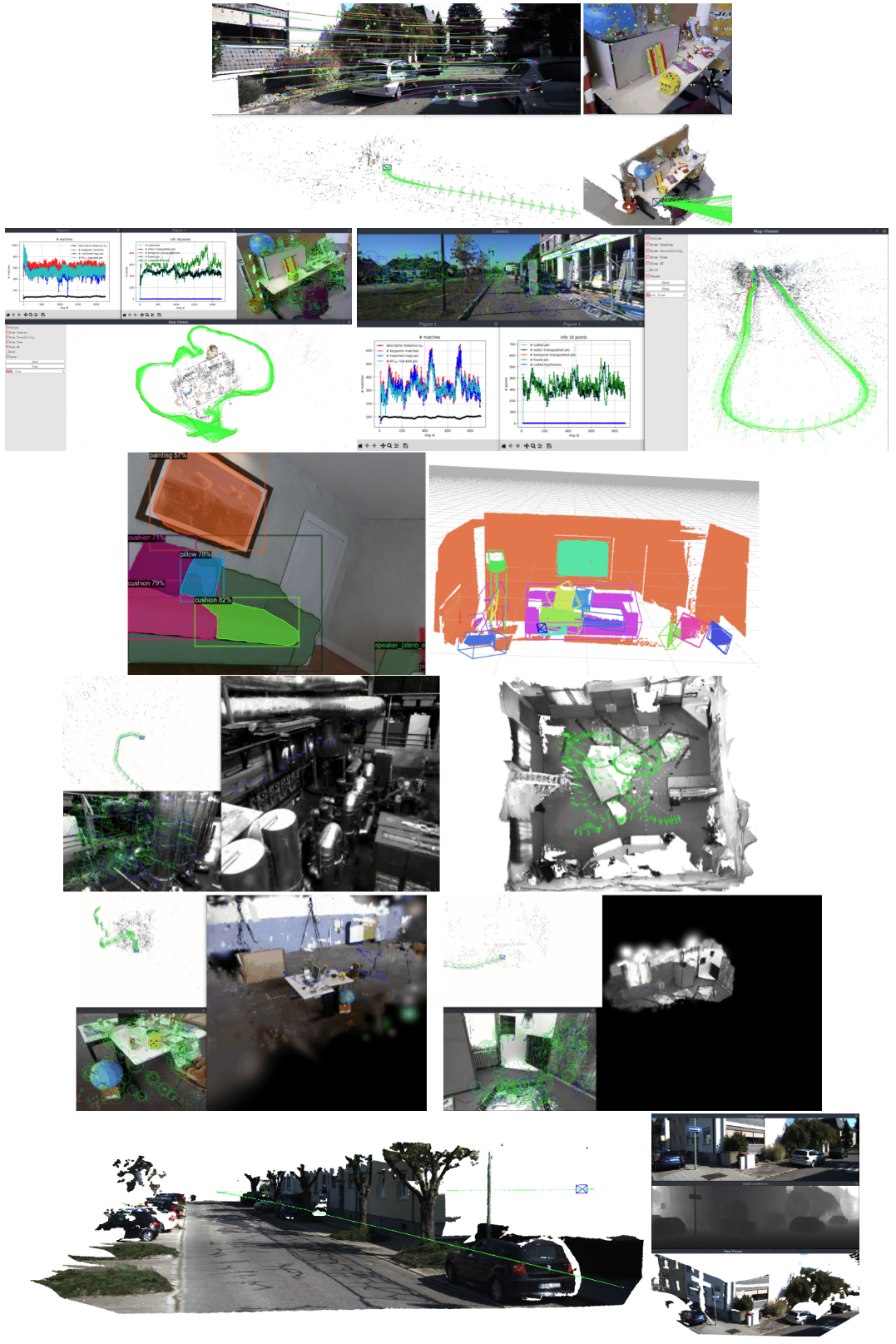}
  \label{fig:teaser}
\end{figure}

\pagebreak

\tableofcontents

\pagebreak

\section{Introduction}

\textbf{pySLAM} is an open-source framework designed to accelerate research and development in Visual SLAM (Simultaneous Localization and Mapping)\cite{cadena2017past} and Spatial AI\cite{davison2018futuremapping}. It originated from a practical need identified during the development of a SLAM/Spatial AI course: The  absence of a unified, accessible, and extensible SLAM ecosystem in C++ and Python. The project was driven by several foundational motivations, including the desire to rapidly prototype and evaluate both \textit{classical} and \textit{learning-based} features within complete VO/SLAM pipelines, ensure reproducibility across a wide range of datasets, and significantly lower the entry barrier for newcomers to the field.

A key decision behind pySLAM development was the choice of \textit{Python} as its primary language. Compared to C++, Python enables faster experimentation and benefits from a mature and expanding ecosystem of deep learning tools and research-oriented APIs. This choice aligns with current trends in computer vision and robotics research, where rapid prototyping and flexible experimentation are essential.

However, designing a modern SLAM framework in Python presents several challenges. The SLAM landscape is highly fragmented -- across feature types, datasets, programming languages (Python vs. C++), and development environments (virtual environments, Docker, versioning issues). Deep learning integration further complicates matters due to backend fragmentation (e.g., PyTorch vs. TensorFlow). These inconsistencies often lead to reproducibility bottlenecks and high barriers to experimental research, where setting up a functional baseline can become more laborious than the research itself.

To address these issues, pySLAM is built on several key design principles. \textit{Modularity} is central: components such as local/global features\cite{tuytelaars2008local,joshi2020recent,ma2021image}, loop closure mechanisms\cite{schubert2023visual}, semantic mapping\cite{georgevichsemantic,chen2025semantic,schmid2024khronos}, and volumetric integration\cite{dong2022ash,freda2023plvs} can be easily swapped or extended. This architecture supports the use of interchangeable baselines, enabling researchers to plug-and-play with different configurations depending on their goals. \textit{Extensibility} is facilitated through a base-class architecture that allows new components to be integrated with minimal friction, making pySLAM a flexible platform for benchmarking and algorithm development.

In terms of usability, pySLAM emphasizes simplicity in the build and deployment process. It supports a unified Python environment compatible with \textit{conda}, \textit{pyenv}, and \textit{pixi}, and offers a one-command installation process that covers setup and dataset acquisition. Cross-platform compatibility (Linux, macOS, Windows with WSL2) and full source accessibility further enhance its utility for both academic and industrial use.

Another cornerstone of the framework is standardized dataset handling. By providing a unified interface to multiple datasets, pySLAM encourages \textit{reproducibility} and \textit{reusability}. It supports consistent data loading, map state saving/loading, and automatic generation of evaluation metrics -- including final and online trajectory estimation -- thus enabling robust comparison between algorithms.

Functionally, pySLAM supports the full visual SLAM pipeline with monocular, stereo, and RGB-D configurations. It accommodates both traditional local features and deep-learning-based methods, and includes loop closure detection integrated with optimization backends like g2o\cite{g2o_kummerle} and GTSAM\cite{factor_graphs_for_robot_perception}. The framework also supports 3D reconstruction via different voxel grid models, TSDF\cite{dong2022ash} and emerging techniques like Gaussian Splatting\cite{bao20253d}, and integrates deep learning models for depth prediction and semantic scene understanding.

Beyond SLAM, pySLAM also provides a unified \textit{ScenefromViews} architecture for modern feed-forward inference of 3D scenes from multiple views. The module supports both pairwise models (DUSt3R\cite{wang2023dust3r}, MASt3R\cite{leroy2024grounding}) that process image pairs and perform global alignment, as well as multi-view models (Depth Anything V3\cite{depthanything3}, MV-DUSt3R\cite{tang2025mv}, VGGT\cite{wang2025vggt}, robust VGGT\cite{vggt_robust}, and Fast3R\cite{yang2025fast3r}) that process multiple views simultaneously in a single forward pass, and the multi-view solver SOLV3R\cite{kwak2026solv3r}. Sessions extend the same interface to views added over time, in an incremental or a streaming mode. This modular design enables flexible integration of diverse 3D reconstruction techniques while maintaining a consistent API across different model architectures.

Together, these capabilities position pySLAM as a powerful research toolkit -- aiming to be a ``Swiss army knife`` for SLAM development, evaluation, and education. By streamlining reproducibility, supporting modular experimentation, and lowering the technical threshold for entry, pySLAM addresses some of the most persistent challenges in contemporary SLAM research.

The remainder of this document is organized as follows. Section~\ref{main-features} summarizes the main features of the framework, and Section~\ref{related-work} relates pySLAM to prior SLAM systems and to the methods it integrates. Section~\ref{sec:architecture} describes the system architecture: the indirect (feature-based) pipeline and its main components, the direct and semi-direct pipeline, the C++ core, the camera models, and the Scene from Views module. Section~\ref{supported-components-and-models} lists the supported components and models. Section~\ref{evaluating-slam} presents the evaluation tools, the results obtained on TUM, EuRoC, and KITTI, the automated test suite, and an initial validation of the camera models on TUM-VI. Section~\ref{conclusions} concludes and outlines future work. Appendix~\ref{usage} describes how to run the main tools, Appendix~\ref{configuration} how to configure datasets and cameras, and Appendix~\ref{visual-models-appendix} how to install the visual models and set their runtime options.

\section{Main Features}\label{main-features}

\textbf{pySLAM} provides a python implementation of a \emph{Visual SLAM} pipeline that supports \textbf{monocular}, \textbf{stereo} and
\textbf{RGBD} cameras. It offers the following \textbf{features} in a single python environment: 

\begin{itemize}
    \item A wide range of classical and modern
    \textbf{\protect\hyperlink{supported-local-features}{local features}}
    with a convenient interface for their integration. 
    
    \item Various loop closing methods, including
    \textbf{\protect\hyperlink{supported-global-descriptors-and-local-descriptor-aggregation-methods}{descriptor
    aggregators}} such as visual Bag of Words (BoW\cite{galvez2012bags}, iBow\cite{garcia2018ibow}), Vector of Locally
    Aggregated Descriptors (VLAD\cite{jegou2010aggregating}), and modern
    \textbf{\protect\hyperlink{supported-global-descriptors-and-local-descriptor-aggregation-methods}{global
    descriptors}} (image-wise descriptors).
    
    \item A
    \textbf{\protect\hyperlink{volumetric-reconstruction}{volumetric
    reconstruction pipeline}} that processes available depth and color
    images with volumetric integration and provides an output dense
    reconstruction. This can use \textbf{TSDF}\cite{dong2022ash} with voxel hashing or
    incremental \textbf{Gaussian Splatting}\cite{matsuki2023gaussian,kerbl20233d}.
    
    \item Integration of
    \textbf{\protect\hyperlink{depth-prediction}{depth prediction models}}
    within the SLAM pipeline. These include DepthPro\cite{depthpro2023}, Depth Pro 2\cite{depthpro2_2026}, DepthAnythingV2\cite{depthanythingv2_2024}, DepthAnythingV3\cite{depthanything3},
    RAFT-Stereo\cite{teed2021raft}, CREStereo\cite{li2022cres}, MASt3R\cite{master}, MVDUSt3R\cite{mvduster}, etc.

    \item A suite of segmentation models for
    \textbf{\protect\hyperlink{semantic-mapping-and-image-segmentation}{semantic understanding}} of
    the scene, such as DeepLabv3\cite{chen2017rethinking}, Segformer\cite{xie2021segformer}, CLIP\cite{lin2023clip}, EOV-Seg\cite{eov_seg}, DETIC\cite{detic}, ODISE\cite{odise}, RFDETR\cite{rfdetr}, and YOLO\cite{redmon2016yolo,redmon2017yolo9000,redmon2018yolov3,bochkovskiy2020yolov4,jocher2020yolov5, wang2022yolov7, jocher2023yolov8, sapkota2025ultralytics}.
    
    \item A \textbf{\protect\hyperlink{scene-from-views}{Scene from Views}} module providing a unified interface for feed-forward inference of 3D scenes from multiple views, supporting models such as DUSt3R\cite{wang2023dust3r}, MASt3R\cite{leroy2024grounding}, Depth Anything V3\cite{depthanything3}, MV-DUSt3R\cite{tang2025mv}, VGGT\cite{wang2025vggt}, Fast3R\cite{yang2025fast3r}, and SOLV3R\cite{kwak2026solv3r}, with incremental and streaming sessions for views added over time.
    
    \item A modular sparse SLAM \textbf{\protect\hyperlink{cpp-core}{Core}}, implemented both in Python and C++ (with custom pybind11 bindings), allowing users to switch between high-performance/speed and high-flexibility modes.

    \item A C++ \textbf{\protect\hyperlink{direct-slam}{direct and semi-direct SLAM}} pipeline with three tracking modes: KLT tracks\cite{lucas1981iterative}, a hybrid mode with sparse photometric alignment, and an SVO-style\cite{forster2014svo} semi-direct mode with a Bayesian inverse-depth filter\cite{vogiatzis2011video}. It supports stereo and RGB-D input, and monocular input in the semi-direct mode, with local bundle adjustment and the same loop closing as the sparse pipeline (with Sim(3) corrections for monocular input).

    \item Several \textbf{\protect\hyperlink{camera-models}{camera models}} in the sparse pipeline: pinhole, Kannala--Brandt\cite{kannala2006generic}, rational distortion, and double sphere\cite{usenko2018double}, with native and rectified fisheye stereo on TUM-VI\cite{schubert2018tumvi}.
    
    \item Additional tools for VO (Visual Odometry) and SLAM, with built-in
    support for both \textbf{g2o}\cite{g2o_kummerle} and \textbf{GTSAM}\cite{factor_graphs_for_robot_perception}, along with custom
    Python bindings for features not available in the original libraries

    \item Built-in support for more than \textbf{\protect\hyperlink{datasets}{15 dataset
    types}}. The list is currently growing.

    \item Tools for \textbf{\protect\hyperlink{evaluating-slam}{reproducible evaluation}}: automated multi-dataset evaluation, a regression test suite, and an end-to-end replay harness with performance and accuracy gates.
\end{itemize}

pySLAM serves as a flexible baseline framework to experiment with VO/SLAM
techniques, \emph{\protect\hyperlink{supported-local-features}{local
features}},
\emph{\protect\hyperlink{supported-global-descriptors-and-local-descriptor-aggregation-methods}{descriptor
aggregators}},
\emph{\protect\hyperlink{supported-global-descriptors-and-local-descriptor-aggregation-methods}{global
descriptors}},
\emph{\protect\hyperlink{volumetric-reconstruction}{volumetric
integration}}, \emph{\protect\hyperlink{depth-prediction}{depth
prediction}}, \emph{\protect\hyperlink{semantic-mapping-and-image-segmentation}{semantic
mapping}}, \emph{\protect\hyperlink{scene-from-views}{scene from views}}, \emph{\protect\hyperlink{cpp-core}{C++ core}}, and \emph{\protect\hyperlink{direct-slam}{direct SLAM}}. It allows to explore, prototype and develop VO/SLAM
pipelines. pySLAM is a research framework and a work in progress. It is
not optimized for real-time performance.

\hypertarget{related-work}{%
\section{Related Work}\label{related-work}}

\paragraph{Feature-based visual SLAM.}
PTAM~\cite{PTAM} introduced the separation of camera tracking and keyframe-based mapping into parallel threads, with bundle adjustment running in the background. ORB-SLAM2~\cite{ORB_SLAM2} built on this design and added loop closing, relocalization, and support for monocular, stereo, and RGB-D cameras, using ORB features~\cite{rublee2011orb} and a bag of binary words~\cite{galvez2012bags} for place recognition. ORB-SLAM3~\cite{campos2021orbslam3} extended the system to visual-inertial and multi-map SLAM and abstracted the camera model, so that pinhole and fisheye cameras share the same pipeline. pySLAM adopts the same tracking, local mapping, and loop closing decomposition (Sect.~\ref{sec:architecture}), but implements it as a modular framework in which each component, from the local features to the optimization backend, can be replaced through a factory interface. A broader overview of the SLAM problem and its open challenges is given in~\cite{cadena2017past}.

\paragraph{Direct and semi-direct methods.}
Direct methods estimate motion from image intensities instead of matched descriptors. LSD-SLAM~\cite{engel2014lsd} tracks semi-dense depth maps with photometric alignment, and DSO~\cite{engel2018dso} optimizes poses and sparse inverse depths jointly in a windowed photometric bundle adjustment. SVO~\cite{forster2014svo,forster2017svo} is semi-direct: sparse image alignment of small patches gives an initial pose, each feature is then aligned individually, and the pose and structure are refined on reprojection error. New points are estimated by a Bayesian inverse-depth filter~\cite{vogiatzis2011video}. KLT-based front ends such as VINS-Mono~\cite{qin2018vins} track Lucas--Kanade features~\cite{lucas1981iterative} and estimate the pose from reprojection error, so they are indirect. pySLAM offers both kinds of tracking in its direct-SLAM pipeline (Sect.~\ref{direct-slam}), and keeps a geometric local bundle adjustment.

\paragraph{Place recognition and loop closing.}
Loop detection has moved from aggregations of local descriptors, such as bags of words~\cite{galvez2012bags} and VLAD~\cite{jegou2010aggregating}, to global image descriptors learned end to end, such as NetVLAD~\cite{arandjelovic2016netvlad}, CosPlace~\cite{berton2023cosplace}, EigenPlaces~\cite{berton2023eigenplaces}, and MegaLoc~\cite{berton2025megaloc}. Schubert et al.~\cite{schubert2023visual} provide a tutorial on visual place recognition. pySLAM exposes both families behind the same loop detector interface, so they can be compared within a complete SLAM pipeline rather than in isolation.

\paragraph{Dense, semantic, and learned reconstruction.}
Dense reconstruction from RGB-D data has been surveyed in~\cite{zollhofer2018state}. pySLAM integrates TSDF fusion with voxel hashing~\cite{dong2022ash}, incremental Gaussian Splatting~\cite{matsuki2023gaussian,kerbl20233d}, and semantic volumetric mapping, following earlier work on combining sparse SLAM with dense and semantic mapping~\cite{freda2023plvs,pyslamSemantic2025}. More recently, feed-forward models such as DUSt3R~\cite{wang2023dust3r}, MASt3R~\cite{leroy2024grounding}, and VGGT~\cite{wang2025vggt} infer camera poses and 3D structure directly from a set of images. pySLAM offers them through a single \textit{Scene from Views} interface (Sect.~\ref{scene-from-views}).

\paragraph{Camera models.}
Wide-angle and fisheye lenses are commonly described by the Kannala--Brandt model~\cite{kannala2006generic} or by the double sphere model~\cite{usenko2018double}, and the TUM-VI benchmark~\cite{schubert2018tumvi} provides fisheye stereo sequences with motion-capture ground truth. Sect.~\ref{camera-models} describes how these models are integrated into the sparse pipeline of pySLAM.

\hypertarget{system_overview}{%
\section{System Architecture}\label{sec:architecture}}

\subsection{Main Scripts}

A convenient entry point is the following collection of \textbf{main scripts}:
\begin{itemize}
\item 
\texttt{main\_vo.py} combines the simplest VO
ingredients without performing any image point triangulation or windowed
bundle adjustment. At each step \(k\), \texttt{main\_vo.py} estimates
the current camera pose \(C_k\) with respect to the previous one
\(C_{k-1}\). The inter-frame pose estimation returns
\([R_{k-1,k},t_{k-1,k}]\) with \(\Vert t_{k-1,k} \Vert=1\). With this
very basic approach, you need to use a ground truth in order to recover
a correct inter-frame scale \(s\) and estimate a valid trajectory by
composing \(C_k = C_{k-1} [R_{k-1,k}, s t_{k-1,k}]\). This script is a
first start to understand the basics of inter-frame feature tracking and
camera pose estimation.
\item
  \texttt{main\_slam.py} adds feature tracking along multiple frames,
  point triangulation, keyframe management, bundle adjustment, loop
  closing, dense mapping and depth inference in order to estimate the
  camera trajectory and build both a sparse and dense map. It's a full
  SLAM pipeline and includes all the basic and advanced blocks which are
  necessary to develop a real visual SLAM pipeline.
\item
  \texttt{main\_slam\_direct.py} runs the C++ direct-SLAM pipeline
  (Sect.~\ref{direct-slam}): KLT, hybrid, or semi-direct tracking, local
  bundle adjustment, and the same loop closing thread as
  \texttt{main\_slam.py}, with Sim(3) corrections for monocular input.
\item
  \texttt{main\_feature\_matching.py} shows how to use the basic feature
  tracker capabilities (\emph{feature detector} + \emph{feature
  descriptor} + \emph{feature matcher}) and allows to test the different
  available local features.
\item
  \texttt{main\_depth\_prediction.py} shows how to use the available
  depth inference models to get depth estimations from input color
  images.
\item
  \texttt{main\_semantic\_image\_segmentation.py} runs the configured
  semantic segmentation model on each frame of the selected dataset and
  visualizes the extracted semantic information.
\item
  \texttt{main\_map\_viewer.py} reloads a saved map and visualizes it.
  Further details on how to save a map
  \protect\hyperlink{reload-a-saved-map-and-relocalize-in-it}{here}.
\item
  \texttt{main\_map\_dense\_reconstruction.py} reloads a saved map and
  uses a configured volumetric integrator to obtain a dense
  reconstruction (see
  \protect\hyperlink{volumetric-reconstruction}{here}).
  \item
  \texttt{main\_slam\_evaluation.py} enables automated SLAM evaluation
  by executing \texttt{main\_slam.py} across a collection of datasets
  and configuration presets (see
  \protect\hyperlink{evaluating-slam}{here}).
  \item
  \texttt{main\_scene\_from\_views.py} infers 3D scenes from multiple
  images using models like Mast3r, VGGT, DepthFromAnythingV3, Fast3R, and SOLV3R (see
  \protect\hyperlink{scene-from-views}{here}).
\end{itemize}

\subsection{System Overview}\label{system_overview}

pySLAM contains two SLAM pipelines and a multi-view reconstruction module, summarized in Table~\ref{tab:system_parts}. The three parts share the configuration, the dataset interface, the camera models, and the viewers. The \textit{indirect} pipeline is described in Sect.~\ref{indirect-components} and the \textit{direct and semi-direct} pipeline in Sect.~\ref{direct-slam}. Sect.~\ref{cpp-core} describes the C++ implementation of the sparse core, Sect.~\ref{camera-models} the camera models shared by both pipelines, Sect.~\ref{feature-management} how classic and learned local features are managed, Sect.~\ref{visual-model-catalog} the visual model catalog, and Sect.~\ref{scene-from-views} the \textit{Scene from Views} module. The diagrams in these sections are sketches of the workflows, components, and class relationships; some minor components and arrows are omitted for readability. Their Mermaid sources are kept in \texttt{docs/images}, in one subfolder per part.

\begin{table}[!ht]
\centering
\caption{The three main parts of pySLAM.}
\label{tab:system_parts}
\fontsize{8pt}{9.6pt}\selectfont
\begin{tabular}{*{5}{>{\raggedright\arraybackslash}p{0.165\textwidth}}}
\toprule
Part & Entry point & Front end & Back end & Maps \\
\midrule
Indirect (feature-based) SLAM & \texttt{main\_slam.py} & Local features, descriptor or learned matching, local-map tracking & Local mapping thread, loop closing, global BA, volumetric and semantic mapping & Sparse map, optional dense and semantic maps \\
Direct and semi-direct SLAM & \texttt{main\_slam\_\allowbreak direct.py} & Descriptor-free \texttt{klt}, \texttt{hybrid}, or \texttt{semi\_direct} tracking in C++ & Synchronous local BA, optional loop closing on descriptor keyframes & Direct landmark map, descriptor map for loop closing \\
Scene from Views & \texttt{main\_scene\_\allowbreak from\_views.py} & Learned multi-view models (DUSt3R, MASt3R, VGGT, \ldots) & Model-specific global alignment & Point clouds, meshes, camera poses \\
\bottomrule
\end{tabular}
\end{table}

\hypertarget{indirect-slam}{%
\subsection{Indirect (Feature-Based) SLAM}\label{indirect-components}}

\begin{figure}[!t]
\begin{center}
    \includegraphics[width=\textwidth]{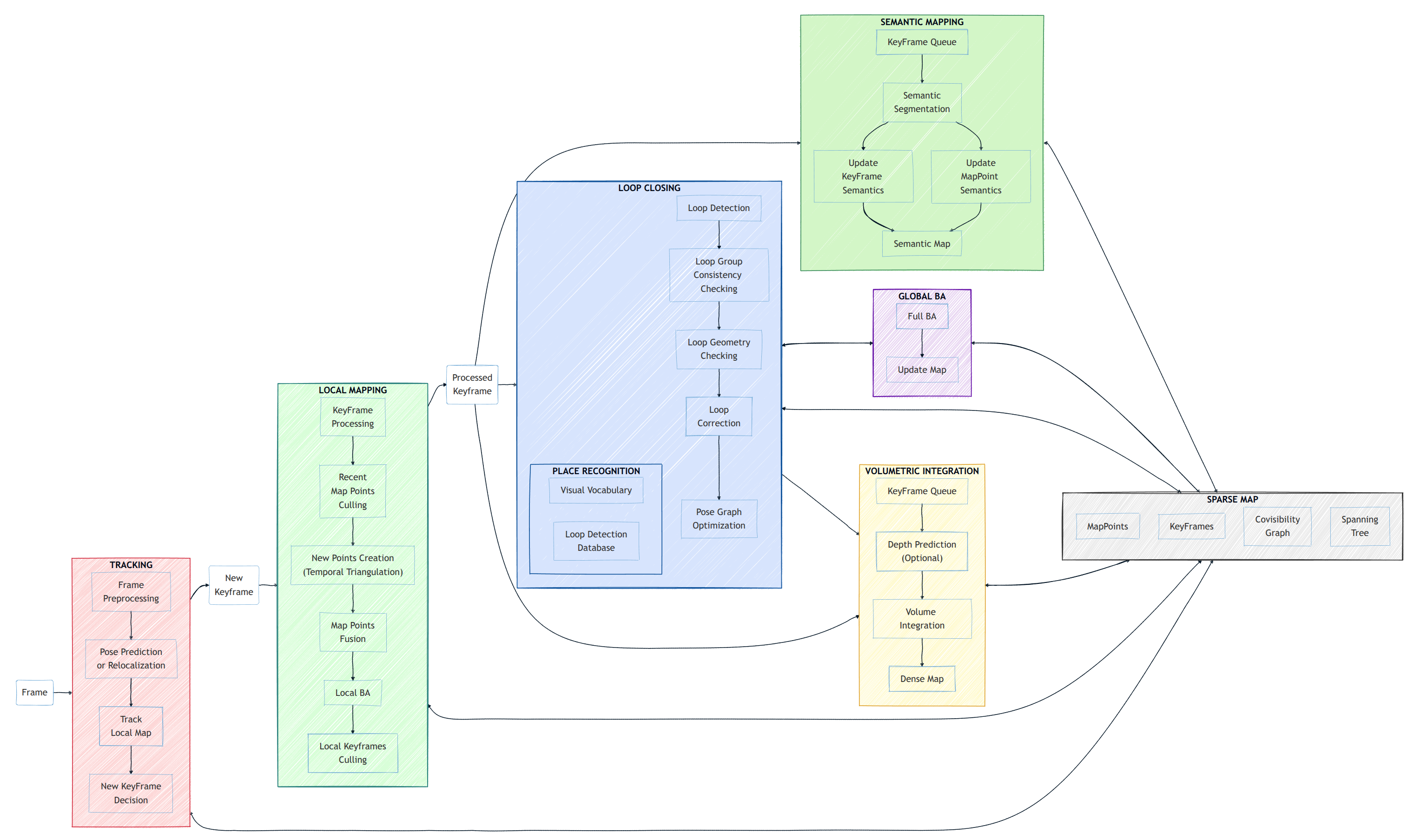}
\end{center}
\caption{Indirect (feature-based) SLAM workflow.}
\label{Fig:SLAMWorfklow}
\end{figure}

\begin{figure}[!t]
\begin{center}
    \includegraphics[width=\textwidth]{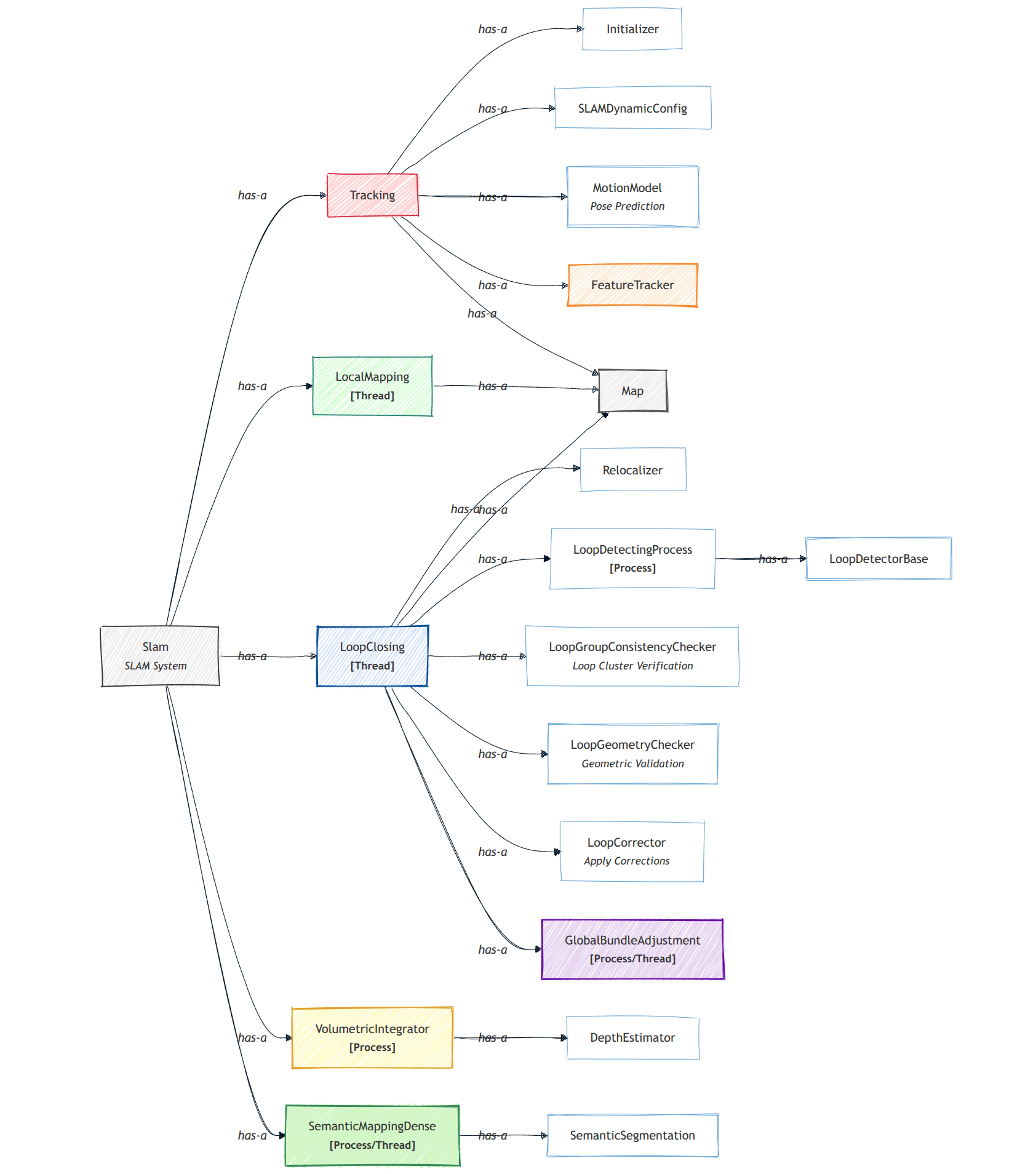}
\end{center}
\caption{Indirect (feature-based) SLAM components.}
\label{Fig:SLAMComponents}
\end{figure}

\subsubsection{Workflow and Components}

The main pipeline, run by \texttt{main\_slam.py}, follows the PTAM~\cite{PTAM} and ORB-SLAM~\cite{ORB_SLAM2} paradigm: tracking, local mapping, and loop closing all operate on local features and their descriptors. Fig.~\ref{Fig:SLAMWorfklow} illustrates its workflow, which is composed of six main parallel processing modules:

\begin{itemize}
\item \textit{Tracking}: estimates the camera pose for each incoming frame by extracting and matching local features to the local map, followed by minimizing the reprojection error through motion-only Bundle Adjustment (BA). It includes components such as pose prediction (or relocalization), feature tracking, local map tracking, and keyframe decision-making.
\item \textit{Local Mapping}: updates and refines the local map by processing new keyframes. This involves culling redundant map points, creating new points via temporal triangulation, fusing nearby map points, performing Local BA, and pruning redundant local keyframes.
\item \textit{Loop Closing}: detects and validates loop closures to correct drift accumulated over time. Upon loop detection, it performs loop group consistency checks and geometric verification, applies corrections, and then launches Pose Graph Optimization (PGO) followed by a full Global Bundle Adjustment (GBA). Loop detection itself is delegated to a parallel process, the \textit{Loop Detector}, which operates independently for better responsiveness and concurrency.
\item \textit{Global Bundle Adjustment}: triggered by the Loop Closing module after PGO, this step globally optimizes the trajectory and the sparse structure of the map to ensure consistency across the entire sequence.
\item \textit{Volumetric Integration}: uses the keyframes, with their estimated poses and back-projected point clouds, to reconstruct a dense 3D map of the environment. This module optionally integrates predicted depth maps and maintains a volumetric representation such as a TSDF~\cite{dong2022ash} or incremental Gaussian Splatting-based volume~\cite{matsuki2023gaussian,kerbl20233d}.
\item \textit{Semantic Mapping}: enriches the SLAM map with dense semantic information by applying pixel-wise segmentation to selected keyframes. Semantic predictions are fused across views to assign semantic labels or descriptors to keyframes and map points. The module operates in parallel, consuming keyframes and associated image data from a queue, applying a configured semantic segmentation model, and updating the map with fused semantic features. This enables advanced downstream tasks such as semantic navigation, scene understanding, and category-level mapping.
\end{itemize}

The first four modules reproduce the PTAM and ORB-SLAM decomposition. The \textit{Tracking} module serves as the front-end, while the remaining modules operate as part of the back-end.

In parallel, the system constructs two types of maps:
\begin{itemize}
\item a \emph{sparse map} ${\cal M}_s = ({\cal K}, {\cal P})$, composed of a set of keyframes $\cal K$ and 3D points ${\cal P}$ derived from matched features;
\item a \emph{volumetric map} (or dense map) ${\cal M}_v$, constructed by the Volumetric Integration module, which fuses back-projected point clouds from the keyframes $\cal K$ into a dense 3D model.
\end{itemize}

To ensure consistency between the sparse and volumetric representations, the volumetric map is updated or re-integrated whenever global pose adjustments occur (e.g., after loop closures).

Fig.~\ref{Fig:SLAMComponents} details the internal components and interactions of the above modules. In certain cases, \textbf{processes} are employed instead of \textbf{threads}. This is due to Python's Global Interpreter Lock (GIL), which prevents concurrent execution of multiple threads in a single process. The use of multiprocessing circumvents this limitation, enabling true parallelism at the cost of some inter-process communication overhead (e.g., via pickling). For an insightful discussion, see this related \href{https://www.theserverside.com/blog/Coffee-Talk-Java-News-Stories-and-Opinions/Is-Pythons-GIL-the-software-worlds-biggest-blunder}{post}.

\begin{figure}[!ht]
\begin{center}
    \includegraphics[width=0.8\textwidth]{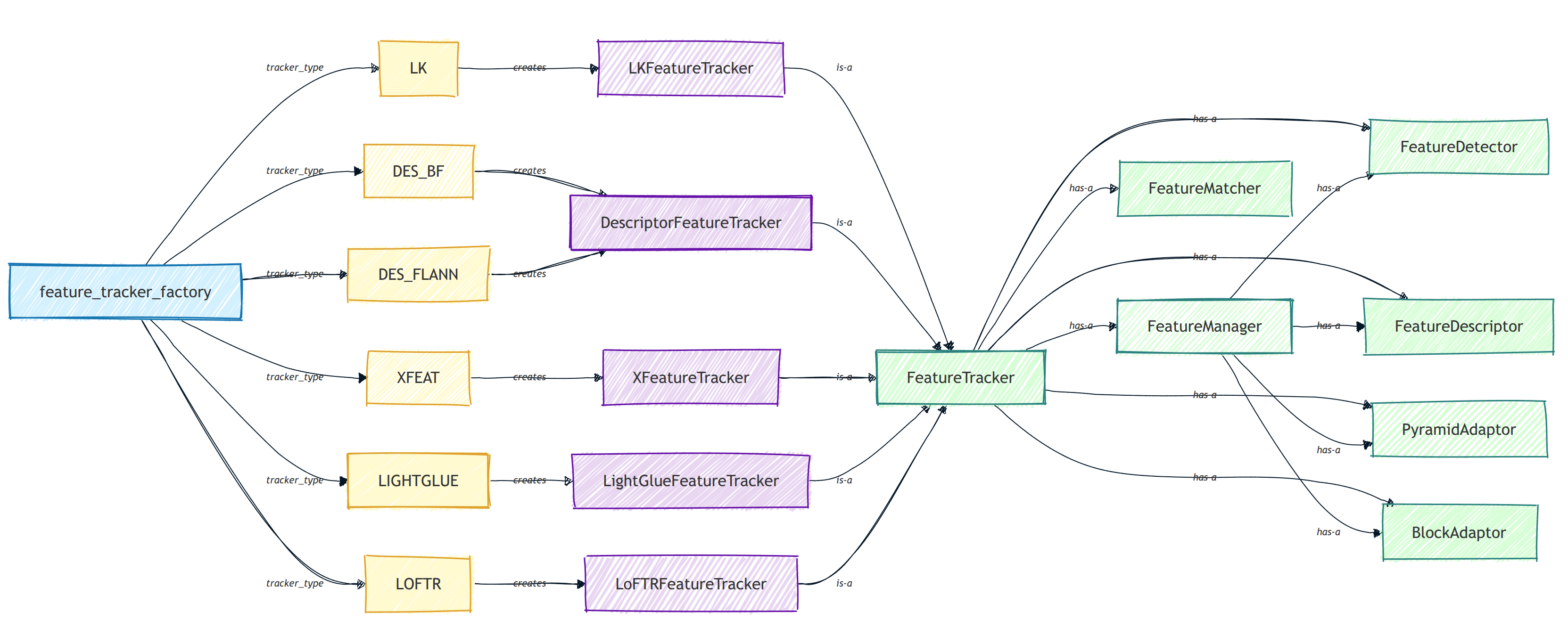}
\end{center}
\caption{Feature tracker.}
\label{Fig:FeatureTracker}
\end{figure}

\subsubsection{Feature Tracker}

The \textit{Feature Tracker} consists of the following key sub-components:

\begin{itemize} 
\item \textit{Feature Manager}: Manages local feature operations. It includes the \texttt{FeatureDetector}, \texttt{FeatureDescriptor}, and adaptors for pyramid management and image tiling. 
\begin{itemize}
\item \textit{Feature Detector}: Identifies salient and repeatable keypoints in the image, such as corners or blobs, which are likely to be robust under viewpoint and illumination changes. 

\item \textit{Feature Descriptor}: Computes a distinctive descriptor for each detected keypoint, encoding its local appearance to enable robust matching across frames. Examples include ORB~\cite{rublee2011orb}, SIFT~\cite{lowe1999object}, or SuperPoint~\cite{detone18superpoint} descriptors.

\end{itemize}

\item \textit{Feature Matcher}: Establishes correspondences between features in successive frames (or stereo pairs) by comparing their descriptors or directly inferring matches from image content. Matching can be performed using brute-force, k-NN with ratio test, or learned matching strategies. Refer to Sect.~\ref{Sect:FeatureMatcher} for further details.
\end{itemize}

Sect.~\ref{supported-local-features} lists the supported local feature extractors and detectors.

The diagram in Fig.~\ref{Fig:FeatureTracker} presents the architecture of the \textit{Feature Tracker} system. It is structured around a \texttt{feature\_tracker\_factory}, which instantiates specific tracker types such as \texttt{LK}, \texttt{DES\_BF}, \texttt{DES\_FLANN}, \texttt{XFEAT}\cite{barroso2024xfeat}, \texttt{LIGHTGLUE}\cite{lindenberger2023lightglue}, and \texttt{LOFTR}\cite{sun2021loftr}. Each tracker type creates a corresponding implementation (e.g., \texttt{LKFeatureTracker}, \texttt{DescriptorFeatureTracker}, etc.), all of which inherit from a common \texttt{FeatureTracker} interface.

The \texttt{FeatureTracker} class is composed of several key sub-components, including a \texttt{FeatureManager}, \texttt{FeatureDetector}, \texttt{FeatureDescriptor}, \texttt{PyramidAdaptor}, \texttt{BlockAdaptor}, and \texttt{FeatureMatcher}. The \texttt{FeatureManager} also encapsulates instances of the detector, descriptor, and adaptors, highlighting the modular and reusable design of the tracking pipeline.

\paragraph{Stereo and RGB-D initialization.} During initialization, both stereo image extractors receive the startup feature boost. After repeated failures, the required point support is progressively relaxed, down to twice the minimum of the depth-based PnP. An initialization is accepted only if the optimization is finite and, for stereo and RGB-D, the left-image reprojection error, measured in the same way before and after BA, improves; monocular initialization keeps its original solver $\chi^2$ comparison. A reset restores the original checks.

\begin{figure}[!t]
\begin{center}
    \includegraphics[width=0.8\textwidth]{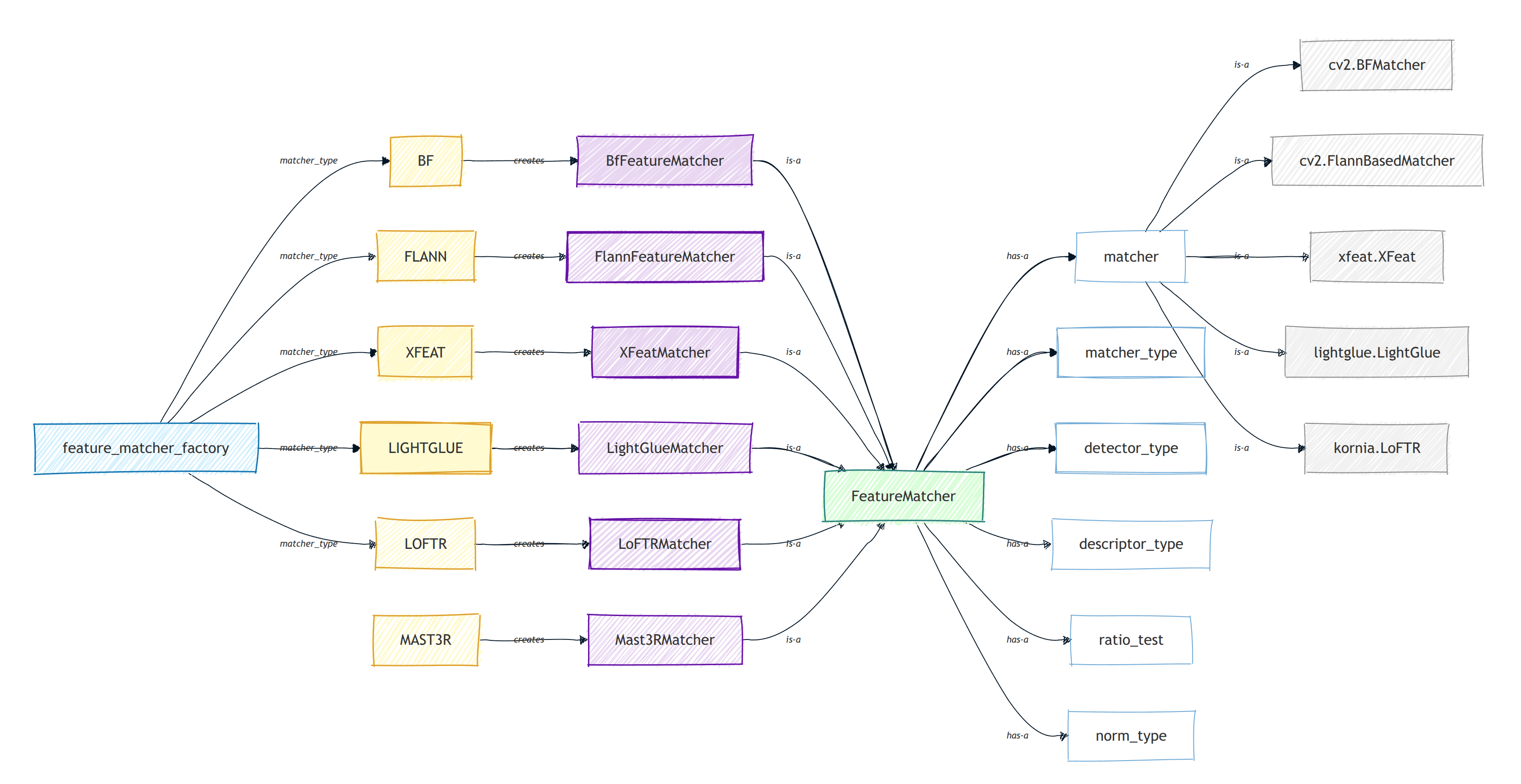}
\end{center}
\caption{Feature matcher.}
\label{Fig:FeatureMatcher}
\end{figure}

\begin{figure}[!t]
\begin{center}
    \includegraphics[width=0.7\textwidth]{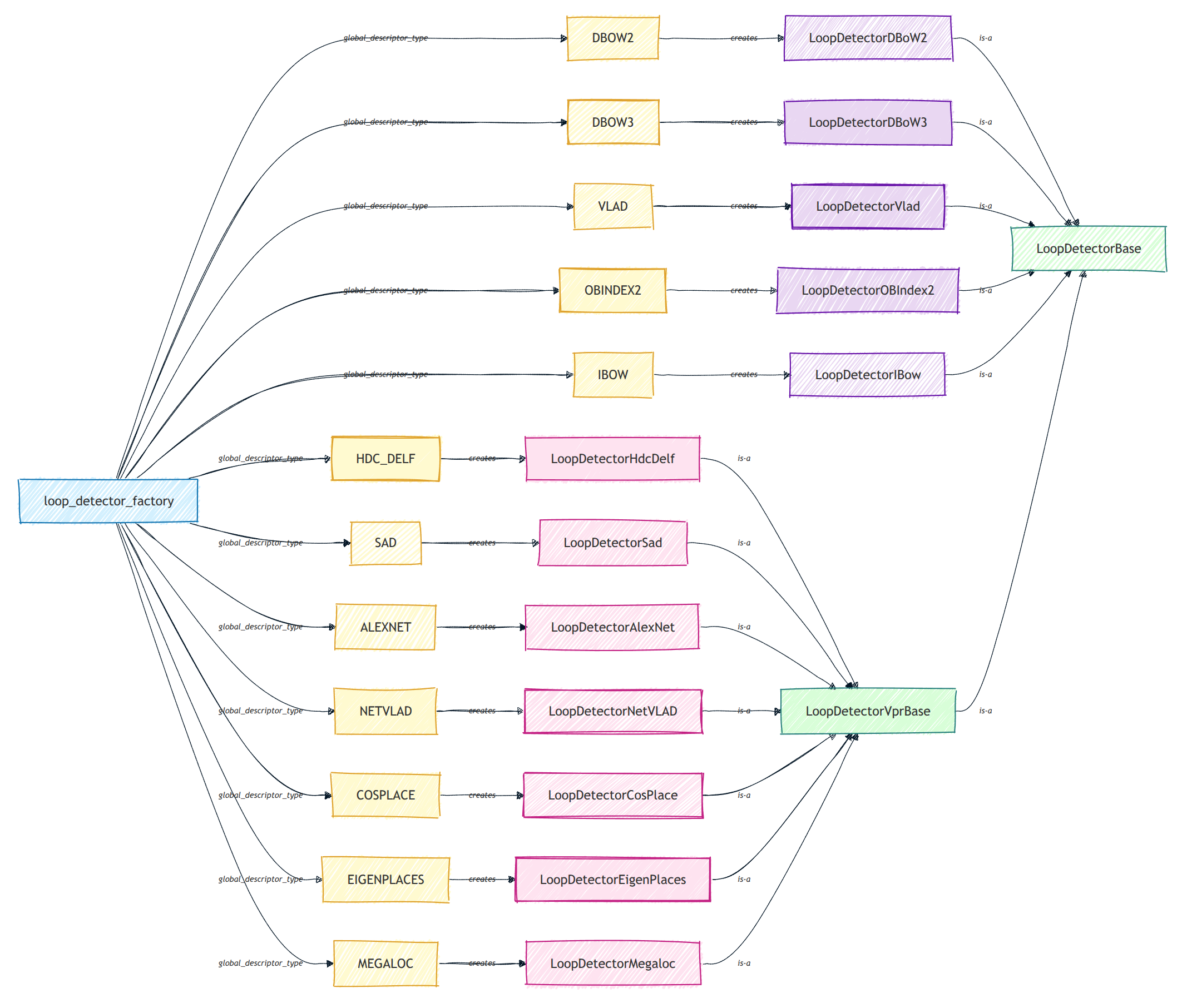}
\end{center}
\caption{Loop detector.}
\label{Fig:LoopDetector}
\end{figure}

\subsubsection{Feature Matcher}\label{Sect:FeatureMatcher}

The diagram in Fig.~\ref{Fig:FeatureMatcher} illustrates the architecture of the \textit{Feature Matcher} module. At its core is the \texttt{feature\_matcher\_factory}, which instantiates matchers based on a specified \texttt{matcher\_type}, such as \texttt{BF}, \texttt{FLANN}, \texttt{XFEAT}\cite{barroso2024xfeat}, \texttt{LIGHTGLUE}\cite{lindenberger2023lightglue}, and \texttt{LOFTR}\cite{sun2021loftr}. Each of these creates a corresponding matcher implementation (e.g., \texttt{BfFeatureMatcher}, \texttt{FlannFeatureMatcher}, etc.), all inheriting from a common \texttt{FeatureMatcher} interface.

The \texttt{FeatureMatcher} class encapsulates several configuration parameters and components, including the matcher engine (\texttt{cv2.BFMatcher}, \texttt{FlannBasedMatcher}, \texttt{xfeat.XFeat}, etc.), as well as the \texttt{matcher\_type}, \texttt{detector\_type}, \texttt{descriptor\_type}, \texttt{norm\_type}, and \texttt{ratio\_test} fields. This modular structure supports extensibility and facilitates switching between traditional and learning-based feature matching backends.

The Section~\ref{supported-matchers} reports a list of supported feature matchers.

\begin{figure}[!t]
\begin{center}
    \includegraphics[width=\textwidth]{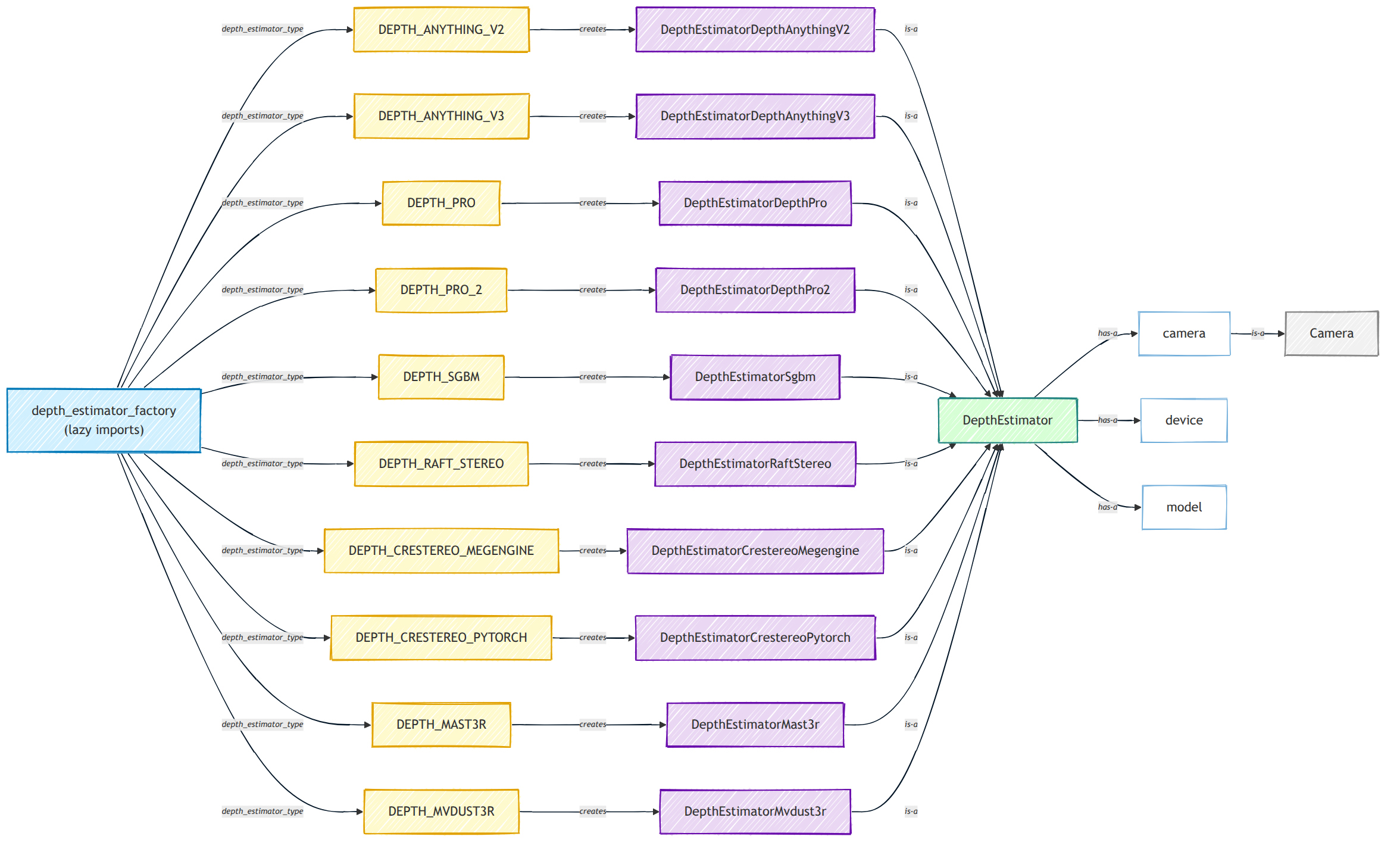}
\end{center}
\caption{Depth estimator.}
\label{Fig:DepthEstimator}
\end{figure}

\subsubsection{Loop Detector}

The diagram in Fig.~\ref{Fig:LoopDetector} shows the architecture of the \textit{Loop Detector} component. A central \texttt{loop\_detector\_factory} instantiates loop detectors based on the selected \texttt{global\_descriptor\_type}, which may include traditional descriptors (e.g., \texttt{DBOW2}\cite{galvez2012bags}, \texttt{VLAD}\cite{jegou2010aggregating}, \texttt{IBOW}\cite{garcia2018ibow}) or deep learning-based embeddings (e.g., \texttt{NetVLAD}\cite{arandjelovic2016netvlad}, \texttt{CosPlace}\cite{berton2023cosplace}, \texttt{EigenPlaces}\cite{berton2023eigenplaces}).

Each descriptor type creates a corresponding loop detector implementation (e.g., \texttt{LoopDetectorDBoW2}, \texttt{LoopDetectorNetVLAD}), all of which inherit from a base class hierarchy. Traditional methods inherit directly from \texttt{LoopDetectorBase}, while deep learning-based approaches inherit from \texttt{LoopDetectorVprBase}, which itself extends \texttt{LoopDetectorBase}. This design supports modular integration of diverse place recognition techniques within a unified loop closure framework.

The Section~\ref{supported-global-descriptors-and-local-descriptor-aggregation-methods} reports a list of supported loop closure methods with the adopted global descriptors and local descriptor aggregation methods.

\subsubsection{Depth Estimator}

The diagram in Fig.~\ref{Fig:DepthEstimator} illustrates the architecture of the \textit{Depth Estimator} module. A central \texttt{depth\_estimator\_factory} creates instances of various depth estimation backends based on the selected \texttt{depth\_estimator\_type}, including both traditional and learning-based methods such as \texttt{DEPTH\_SGBM}\cite{hirschmuller2007stereo}, \texttt{DEPTH\_RAFT\_STEREO}\cite{teed2021raft}, \texttt{DEPTH\_ANYTHING\_V2}\cite{depthanythingv2_2024}, \texttt{DEPTH\_ANYTHING\_V3}\cite{depthanything3}, \texttt{DEPTH\_PRO}\cite{depthpro2023}, \texttt{DEPTH\_PRO\_2}\cite{depthpro2_2026}, \texttt{DEPTH\_MAST3R}\cite{master}, and \texttt{DEPTH\_MVDUST3R}\cite{mvduster}. The factory imports the selected implementation lazily, when it is constructed: importing the factory loads no estimator or thirdparty package.

Each estimator type instantiates a corresponding implementation (e.g., \texttt{DepthEstimatorSgbm}, \texttt{DepthEstimator\-Crestereo\-Megengine}, etc.), all inheriting from a common \texttt{DepthEstimator} interface. This base class encapsulates shared dependencies such as the \texttt{camera}, \texttt{device}, and \texttt{model} components, allowing for modular integration of heterogeneous depth estimation techniques across stereo, monocular, and multi-view pipelines.

The Section~\ref{supported-depth-prediction-models} reports a list of supported depth estimation/prediction models.

\begin{figure}[!t]
\begin{center}
    \includegraphics[width=\textwidth]{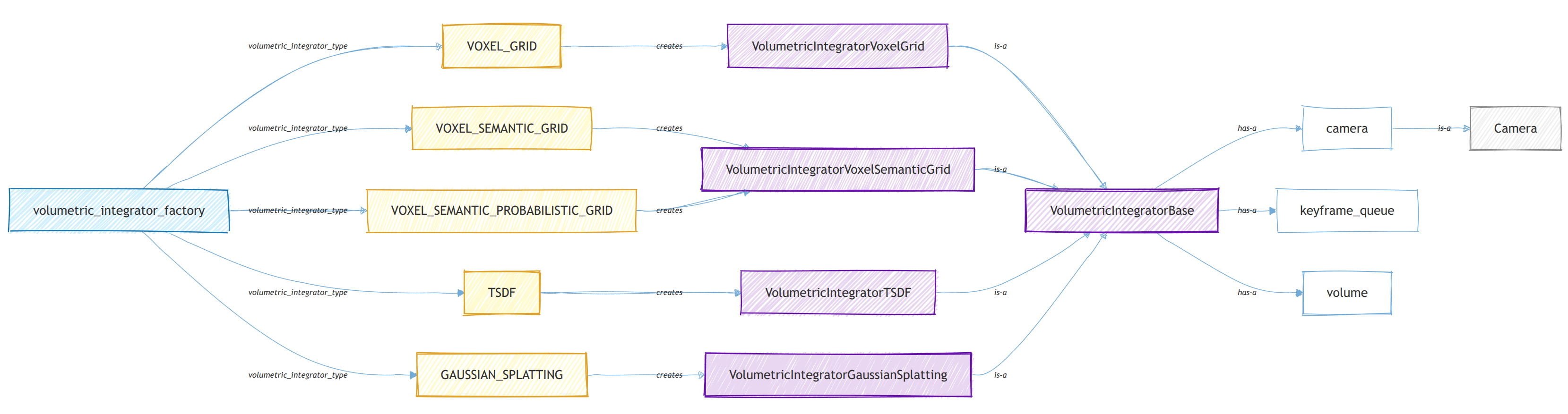}
\end{center}
\caption{Volumetric integrator.}
\label{Fig:VolumetricIntegrator}
\end{figure}

\subsubsection{Volumetric Integrator}

The diagram in Fig.~\ref{Fig:VolumetricIntegrator} illustrates the structure of the \textit{Volumetric Integrator} module. At its core, the \texttt{volumetric\_integrator\_factory} generates specific volumetric integrator instances based on the selected \texttt{volumetric\_integrator\_type}, such as \texttt{VOXEL\_GRID}, \texttt{VOXEL\_SEMANTIC\_GRID}, \texttt{VOXEL\_SEMANTIC\_PROBABILISTIC\_GRID}, \texttt{TSDF}\cite{dong2022ash}, and \texttt{GAUSSIAN\_SPLATTING}\cite{matsuki2023gaussian,kerbl20233d}.

Each type instantiates a dedicated implementation (e.g., \texttt{VolumetricIntegratorVoxelGrid}, \texttt{VolumetricIntegrator\-Voxel\-Semantic\-Grid}, \texttt{VolumetricIntegratorTSDF}, \texttt{VolumetricIntegrator\-Gaussian\-Splatting}), which inherits from a common \texttt{VolumetricIntegratorBase}. This base class encapsulates key components including the \texttt{camera}, a \texttt{keyframe\_queue}, and the \texttt{volume}, enabling flexible integration of various 3D reconstruction methods within a unified pipeline.

The Section~\ref{supported-volumetric-mapping-methods} reports a list of supported volume integration methods.

\begin{figure}[!t]
\begin{center}
    \includegraphics[width=\textwidth]{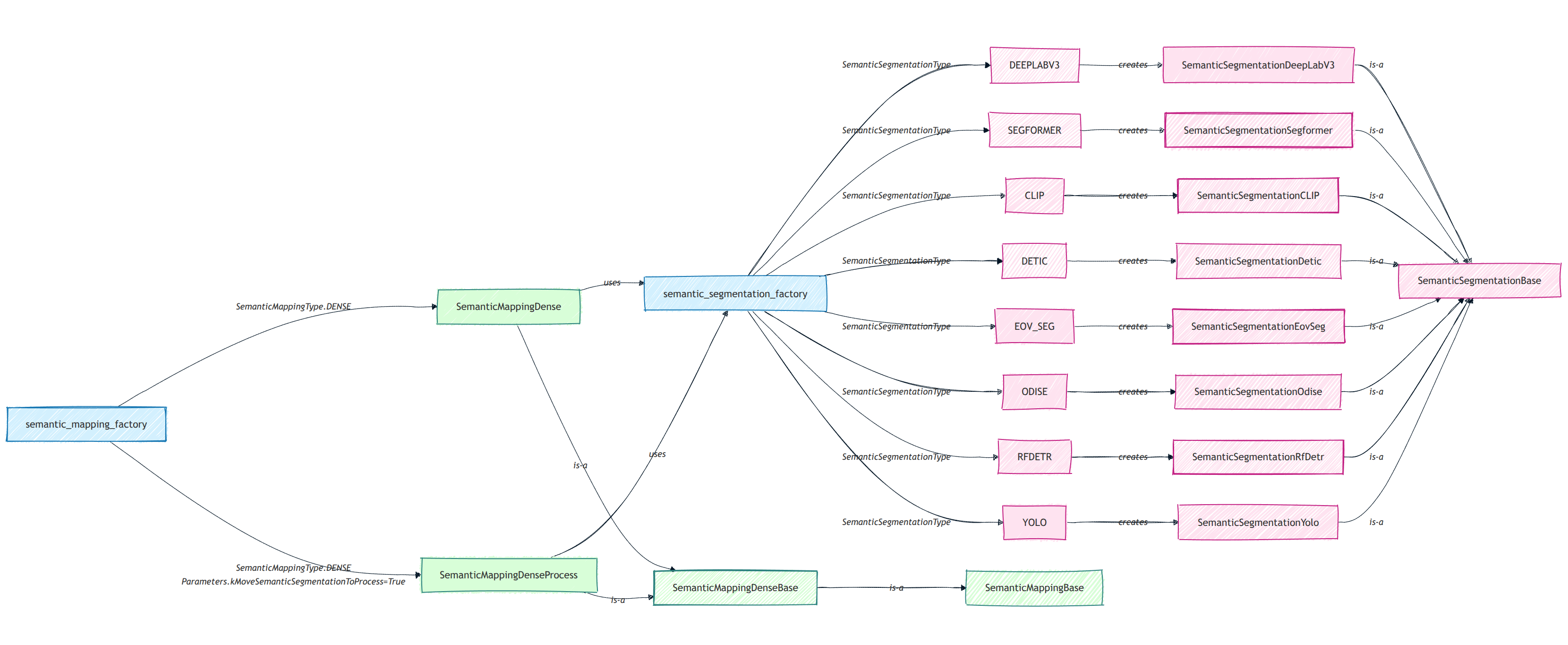}
\end{center}
\caption{Semantic Mapping.}
\label{Fig:SemanticMapping}
\end{figure}

\subsubsection{Semantic Mapping}

The diagram in Fig.~\ref{Fig:SemanticMapping} outlines the architecture of the \textit{Semantic Mapping} module. At its core is the \texttt{semantic\_mapping\_factory}, which creates semantic mapping instances according to the selected \texttt{semantic\_mapping\_type}. Currently, the supported type is \texttt{DENSE}, which instantiates either the \texttt{SemanticMappingDense} class or the \texttt{SemanticMappingDenseProcess} class if
\begin{scriptsize}
\begin{Shaded}
\begin{Highlighting}[]
\NormalTok{Parameters.kSemanticMappingMoveSemanticSegmentationToSeparateProcess=True}.
\end{Highlighting}
\end{Shaded}
\end{scriptsize}
Both these classes extend \texttt{SemanticMappingDenseBase} and run asynchronously in a dedicated thread to process keyframes as they become available. In particular, the \texttt{SemanticMappingDenseProcess} class offloads semantic segmentation inference to a separate process, thereby making better use of CPU core parallelism.

Both \texttt{SemanticMappingDense} and \texttt{SemanticMappingDenseProcess} integrate semantic information into the SLAM map by leveraging per-frame predictions from a semantic segmentation model. The segmentation model is instantiated via the \texttt{semantic\_segmentation\_factory}, based on the selected \texttt{semantic\_segmentation\_type}. Supported segmentation backends include \texttt{DEEPLABV3}\cite{chen2017rethinking}, \texttt{SEGFORMER}\cite{xie2021segformer}, \texttt{CLIP}\cite{lin2023clip}, \texttt{EOV\_SEG}\cite{eov_seg}, \texttt{DETIC}\cite{detic}, \texttt{ODISE}\cite{odise}, \texttt{YOLO}\cite{redmon2016yolo,redmon2017yolo9000,redmon2018yolov3,bochkovskiy2020yolov4,jocher2020yolov5, wang2022yolov7, jocher2023yolov8, sapkota2025ultralytics}, and \texttt{RFDETR}\cite{rfdetr}, each of which corresponds to a dedicated class (\texttt{SemanticSegmentationDeepLabV3}, \texttt{SemanticSegmentationSegformer}, \texttt{SemanticSegmentationCLIP}, \texttt{SemanticSegmentationEovSeg}, \texttt{Semantic\-Segmentation\-Detic}, \texttt{Semantic\-Segmentation\-Odise}, \texttt{Semantic\-Segmentation\-Yolo}, \texttt{Semantic\-Segmentation\-RfDetr}), all inheriting from the shared \texttt{SemanticSegmentationBase}.

The system supports multiple semantic feature representations~--~such as categorical labels, probability vectors, and high-dimensional feature embeddings~--~and fuses them into the map using configurable methods like count-based fusion, Bayesian fusion, or feature averaging.

This modular design decouples semantic segmentation from mapping logic, enabling flexible combinations of segmentation models, datasets (e.g., \texttt{NYU40}\cite{Silberman:ECCV12}, \texttt{Cityscapes}\cite{Cityscapes}), and fusion strategies. It also supports customization via configuration files or programmatic APIs for dataset-specific tuning or deployment.

The Section~\ref{semantic-mapping-and-image-segmentation} provides a list of supported semantic segmentation methods.
A more in-depth presentation of the semantic mapping module is provided in~\cite{pyslamSemantic2025}.

\hypertarget{direct-slam}{%
\subsection{Direct and Semi-Direct SLAM}\label{direct-slam}}

\begin{figure}[!t]
\begin{center}
    \includegraphics[width=\textwidth]{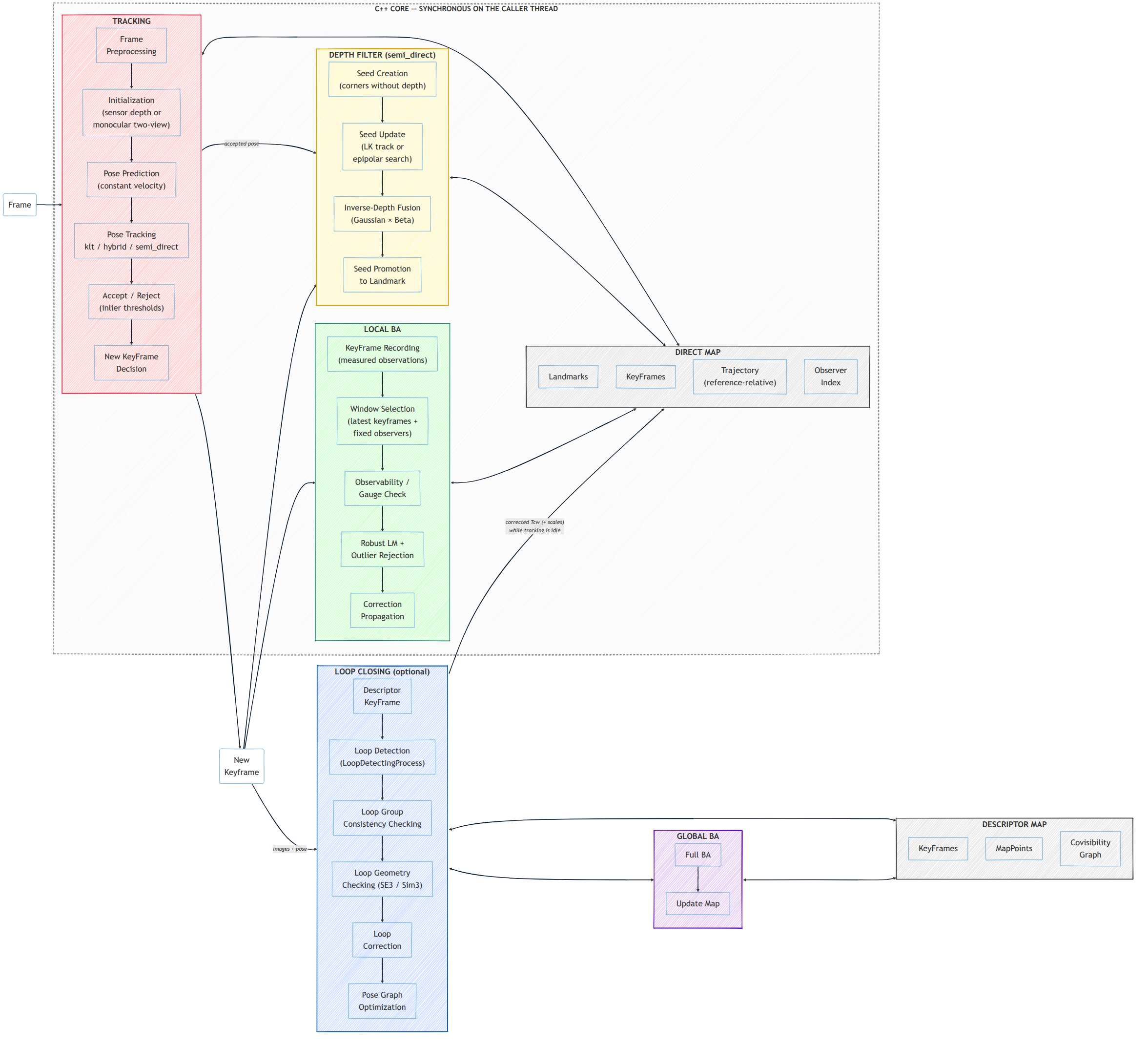}
\end{center}
\caption{Direct and semi-direct SLAM workflow. Tracking, the depth filter, and local BA run synchronously inside one \texttt{DirectSlam.track()} call; loop closing runs on its own thread over a separate descriptor map.}
\label{Fig:DirectSLAMWorkflow}
\end{figure}

\begin{figure}[!t]
\begin{center}
    \includegraphics[width=\textwidth]{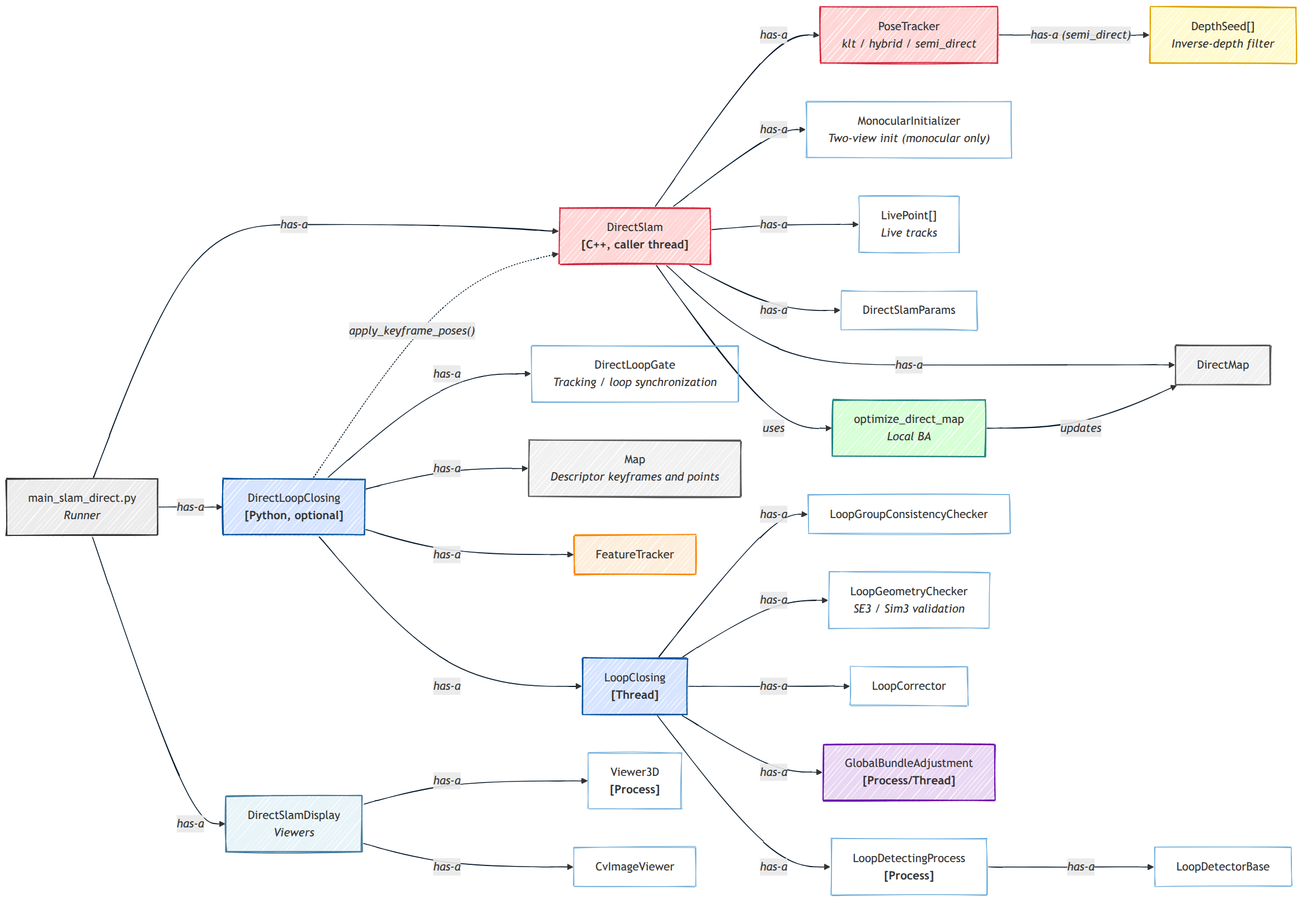}
\end{center}
\caption{Direct and semi-direct SLAM components.}
\label{Fig:DirectSLAMComponents}
\end{figure}

\texttt{main\_slam\_direct.py} runs a second visual SLAM pipeline, which exists only in C++ (\texttt{cpp/core/slam/direct}) and is compiled into the same \texttt{cpp\_core} module. Tracking and local mapping are descriptor-free: their correspondences come from image intensities on raw pixels (grid FAST or GFTT corners, pyramidal Lucas--Kanade, photometric patch alignment), and descriptors are extracted only on separate keyframes for the optional loop closing. The pipeline reuses the configuration, datasets, camera factory, viewers, and feature-trail renderer of the indirect pipeline. It does not provide dense or semantic mapping.

\subsubsection{Workflow and Components}

Fig.~\ref{Fig:DirectSLAMWorkflow} shows the workflow, which has four modules. Unlike in the indirect pipeline, the first three run synchronously on the caller's thread, inside one \texttt{DirectSlam.track()} call, and the Python binding releases the GIL during that call.
\begin{itemize}
\item \textit{Tracking}: initializes the map from sensor depth (stereo or RGB-D) or, for monocular input, from a two-view geometry. It predicts each pose with a constant-velocity model, estimates it with the configured mode, and accepts or rejects it from its inlier count. A rejected frame keeps the last accepted reference for recovery. A low live-track count, or the maximum keyframe interval, triggers a new keyframe.
\item \textit{Depth Filter} (\texttt{semi\_direct} only): corners without a depth measurement become inverse-depth seeds, which each accepted frame updates and which are promoted to landmarks when they converge.
\item \textit{Local BA}: records the measured keyframe observations and jointly optimizes the latest keyframes and their landmarks. The validated corrections are propagated to non-keyframe poses and live tracks.
\item \textit{Loop Closing} (optional): reuses the \texttt{LoopClosing} thread of the indirect pipeline on separately extracted descriptor keyframes, and writes the corrected poses back into the direct map while tracking is idle.
\end{itemize}

The system therefore keeps two maps: the \emph{direct map}, which archives keyframes, persistent landmarks, reference-relative trajectory poses, and an observer index; and the \emph{descriptor map}, a standard pySLAM \texttt{Map} of descriptor keyframes and points that is used only by loop closing and whose keyframes are aligned to the direct keyframes.

Fig.~\ref{Fig:DirectSLAMComponents} shows the main classes. \texttt{DirectSlam} is the C++ system object. It owns the mode-specific \texttt{PoseTracker}, the \texttt{MonocularInitializer}, the \texttt{DirectMap}, the live tracks, and the parameters, and calls \texttt{optimize\_direct\_map} for local BA. On the Python side, \texttt{DirectLoopClosing} adapts the direct map to the standard \texttt{LoopClosing} thread, a \texttt{DirectLoopGate} lets tracking and loop closing take turns on the map, and \texttt{DirectSlamDisplay} copies map snapshots after each \texttt{track()} call and feeds the shared \texttt{Viewer3D} and \texttt{CvImageViewer}.

\begin{figure}[!t]
\begin{center}
    \includegraphics[width=\textwidth]{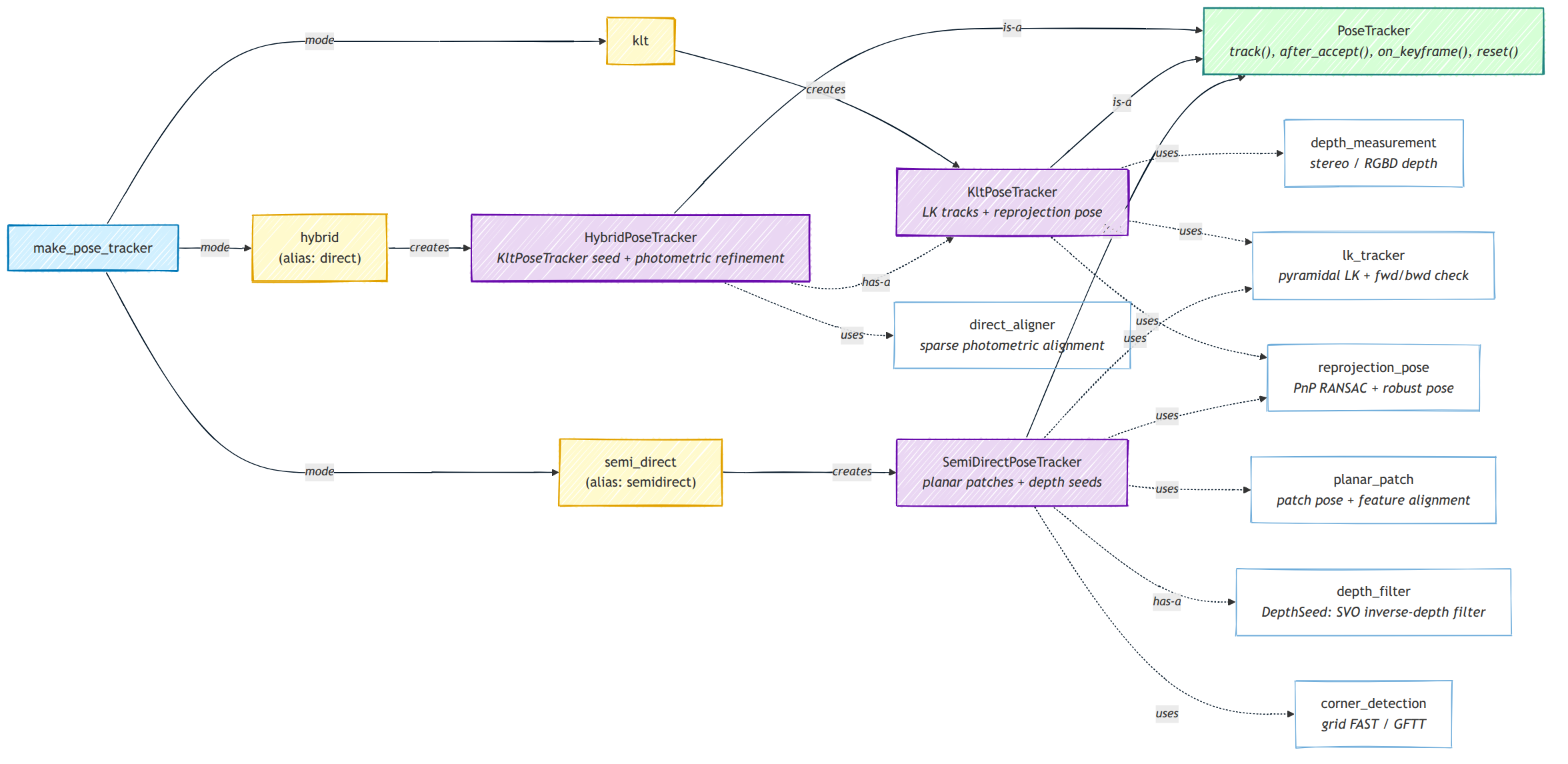}
\end{center}
\caption{Direct SLAM pose tracker.}
\label{Fig:DirectPoseTracker}
\end{figure}

\subsubsection{Pose Tracker}

The tracking mode is selected by \texttt{DIRECT\_SLAM.mode}, and \texttt{make\_pose\_tracker} creates one implementation of the \texttt{PoseTracker} interface per mode (Fig.~\ref{Fig:DirectPoseTracker}):
\begin{itemize}
\item \texttt{klt} (\texttt{KltPoseTracker}): grid FAST or GFTT corners are tracked by pyramidal Lucas--Kanade~\cite{lucas1981iterative} with a forward--backward check. The pose comes from PnP RANSAC followed by robust Gauss--Newton on the reprojection error. In the usual taxonomy, which classifies a method by the error that pose and structure are optimized on, this mode is indirect.
\item \texttt{hybrid} (\texttt{HybridPoseTracker}): the \texttt{klt} pose is refined by sparse photometric alignment of patches with known depth, and the refinement is kept only when it agrees with the geometric tracks. If corner flow fails, photometric-only tracking can recover when enough reference landmarks support it.
\item \texttt{semi\_direct} (\texttt{SemiDirectPoseTracker}): the SVO-style mode~\cite{forster2014svo,forster2017svo}. Each frame is aligned to the reference keyframe in three steps. First, a photometric solve aligns fronto-parallel 3D patches, obtained by unprojecting every patch pixel at the feature depth. Then each projected patch is aligned on its own in 2D, with the pose fixed. Finally, the aligned pixels feed the same reprojection pose solver. This mode owns the depth seeds and is the only one that accepts monocular input; it does not fall back to Lucas--Kanade tracking.
\end{itemize}
\texttt{DirectSlam} keeps what is common to all modes (input checks, initialization, the accept/reject decision, live-track bookkeeping, keyframes, and local BA), and the trackers plug in through the \texttt{track()}, \texttt{after\_accept()}, \texttt{on\_keyframe()}, and \texttt{reset()} hooks. The building blocks are free functions grouped by folder: Lucas--Kanade code under \texttt{klt/}, photometric code under \texttt{direct/}, and shared geometry, corner detection, depth measurement, and image utilities under \texttt{geometry/} and \texttt{features/}. Undistorted pinhole cameras and the raw-pixel models of Sect.~\ref{camera-models} are supported in all modes without remapping the images; with a raw-pixel model, the photometric warp uses the projection Jacobian of the camera.

\paragraph{Depth filter.} In \texttt{semi\_direct} mode, a corner without sensor depth becomes an inverse-depth seed, updated by the SVO depth filter~\cite{forster2014svo} with the parametric model of Vogiatzis and Hern\'andez~\cite{vogiatzis2011video}. The inverse depth is a Gaussian inlier hypothesis mixed with a uniform outlier hypothesis over $[1/d_{\max}, 1/d_{\min}]$, where $d_{\min}$ and $d_{\max}$ are the configured depth bounds (\texttt{seed\_min\_depth} and \texttt{seed\_max\_depth}), and a Beta distribution models the inlier ratio. The prior variance is chosen so that $\pm 3\sigma$ covers the depth range, and the mean starts at the median scene depth of the keyframe. A seed is tracked from frame to frame with pyramidal Lucas--Kanade, starting at its keyframe pixel; this keeps it alive under forward motion, where a fronto-parallel patch matched against an older keyframe image no longer fits. Each tracked position is triangulated against the keyframe ray and yields one inverse-depth measurement. Its variance $\tau^2$ is the inverse-depth change caused by a one-pixel angular error on the current ray, as in SVO. When the Lucas--Kanade track is lost, the seed receives one epipolar search along its $\pm 3\sigma$ segment, using zero-mean SSD on the same planar patch with a parabolic subpixel refinement. A match re-acquires the seed; otherwise the seed is dropped, so that stale seeds do not take capacity from new keyframes. A seed is promoted to a landmark when its inverse-depth variance falls below a threshold and its match has enough parallax, and it is removed after repeated failures or when its expected inlier ratio becomes too low.

\paragraph{Monocular initialization.} A monocular map starts from a two-view initialization, in the style of SVO and ORB-SLAM. Corners detected on a reference frame are tracked with pyramidal Lucas--Kanade until their median displacement exceeds a threshold. A five-point essential matrix~\cite{nister2004efficient} estimated with RANSAC on the bearings then gives the relative pose, and the inliers are triangulated subject to parallax and reprojection gates. The scale is chosen so that the median scene depth equals $\sqrt{d_{\min} d_{\max}}$. The reference frame becomes the first keyframe at the identity, each triangulated landmark receives the same inverse-depth prior as a promoted seed, and local BA runs. The reference is replaced when too few tracks survive or the geometry is degenerate (e.g., under pure rotation). The tracking state stays \texttt{INIT} until initialization succeeds. With stereo or RGB-D input, the first frame with enough depth-backed corners defines the world origin; a measured corner becomes a landmark immediately, and only corners without a valid depth become seeds.

\begin{figure}[!t]
\begin{center}
    \includegraphics[width=\textwidth]{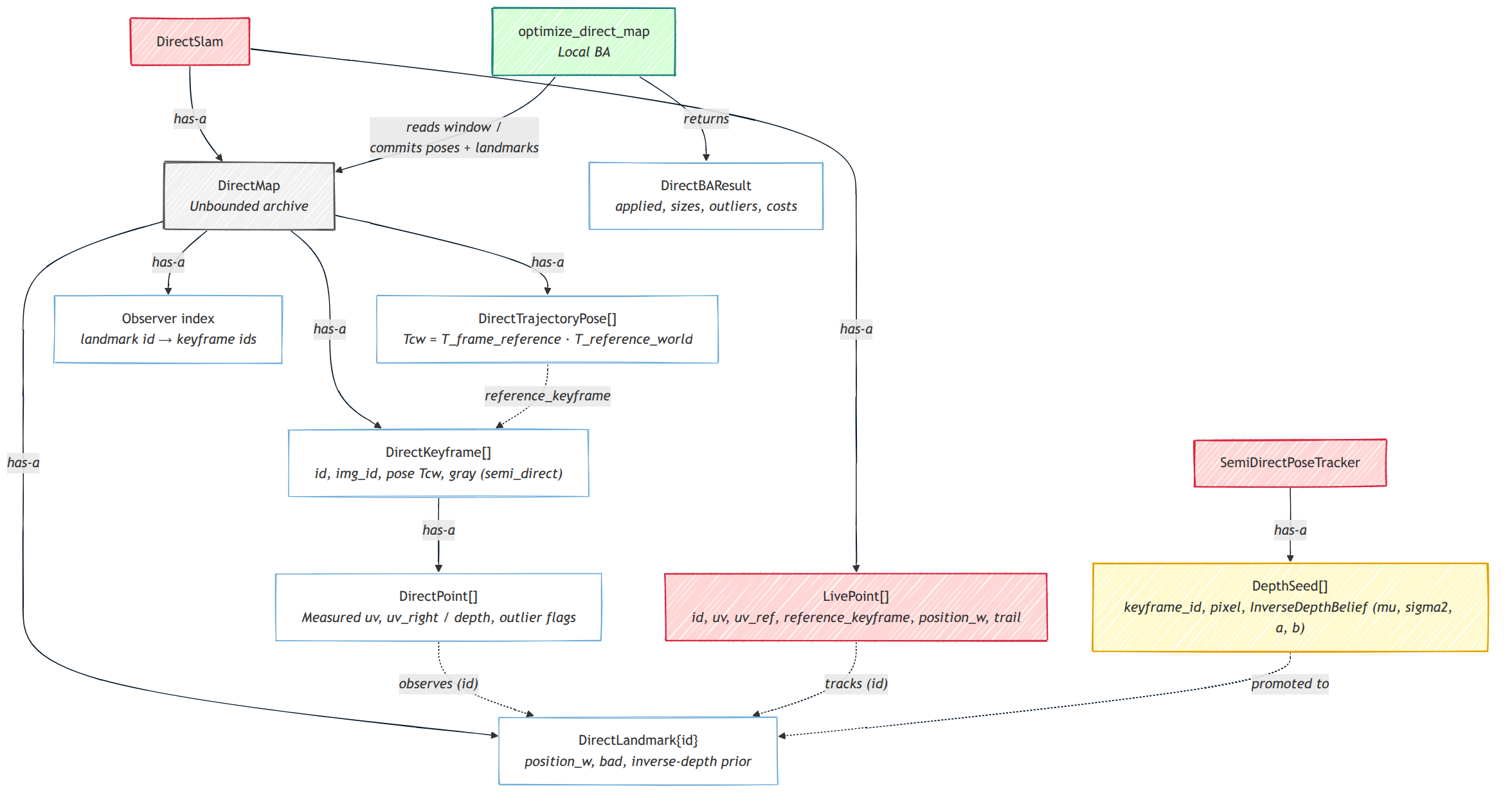}
\end{center}
\caption{Direct map data model and local BA.}
\label{Fig:DirectMap}
\end{figure}

\subsubsection{Direct Map and Local Bundle Adjustment}

Fig.~\ref{Fig:DirectMap} shows the data model of the direct map. \texttt{DirectMap} is an unbounded archive; only the optimization window is bounded. A \texttt{DirectKeyframe} stores its pose and the immutable measurements of its observations (\texttt{DirectPoint}: left pixel, right pixel or depth, and per-channel outlier flags); in \texttt{semi\_direct} mode it also keeps its gray image while it is the tracking reference or still owns seeds. A \texttt{DirectLandmark} stores the world position shared by all observations of the same identifier and, for monocular input, the inverse-depth prior of the seed it was promoted from. A \texttt{DirectTrajectoryPose} stores each accepted frame relative to its reference keyframe, $T_{fw} = T_{fr} T_{rw}$, so that non-keyframe poses follow every BA and loop correction. A \texttt{LivePoint} is a track that is currently followed, linked to a landmark and to a reference keyframe.

The backend follows the joint pose--landmark formulation of \texttt{slamplay}'s \texttt{vslam} and runs synchronous local bundle adjustment with g2o~\cite{g2o_kummerle} on the latest keyframes, in chronological order. The oldest active pose is fixed, and older keyframes that observe active landmarks contribute fixed measurement constraints, so that the sliding window cannot move shared points without accounting for their archived observations. The residuals are the left-camera reprojection error, the right-camera reprojection error for stereo (with the full calibrated transform for native rigs), and the measured depth for RGB-D, with standard deviation $\sigma_z = \sigma_0 \max(1, z^2)$, where $\sigma_0$ is the configured \texttt{ba\_depth\_sigma}. Stereo and RGB-D measurements fix the metric scale. A monocular landmark, whether promoted from a seed or triangulated at initialization, adds an inverse-depth prior as a binary edge between the landmark and the keyframe in which the prior was measured; these priors are the scale gauge. Disconnected or scale-unobservable windows are skipped. Robust Levenberg--Marquardt with Huber kernels is followed by outlier exclusion and a second refinement, and the solution is committed only if it is finite and its full robust objective, including excluded edges, is acceptable. There is no photometric bundle adjustment.

\subsubsection{Loop Closing on Descriptor Keyframes}

Loop closing reuses the indirect components of Fig.~\ref{Fig:SLAMComponents}: \texttt{LoopClosing}, \texttt{LoopDetectingProcess} with any loop detector of Sect.~\ref{supported-global-descriptors-and-local-descriptor-aggregation-methods}, the group-consistency and geometry checkers, the loop corrector, pose-graph optimization, and global BA. For each new direct keyframe, the tracking thread registers an empty descriptor keyframe and queues its images; \texttt{LoopDetectingProcess} extracts the configured local features and returns them to the \texttt{LoopClosing} thread, which creates the descriptor map points from the latest direct pose of the keyframe. Feature extraction is therefore removed from the tracking thread. Corrected poses are written back into the direct map only while tracking is idle.

Stereo and RGB-D keyframes obtain the depth of their descriptor keypoints as in the indirect pipeline, and loops are verified and corrected in SE(3). Monocular keyframes have no such depth. A descriptor keypoint within 3\,px of a direct landmark observation takes the 3D position of that landmark, so that one descriptor point per landmark is shared by every keyframe that observes it, and each new keyframe triangulates its remaining descriptor matches with its three predecessors at the direct poses. Geometric verification is then carried out in Sim(3). After a loop correction or global BA, the scale change $s_k$ of each keyframe $k$ is recovered as the median ratio between the camera distances of its landmark-associated points after and before the correction (triangulated matches are excluded), and keyframes with too few such points take the nearest estimate. \texttt{DirectSlam.apply\_keyframe\_poses} then moves each landmark through its anchor keyframe as a similarity transformation and rescales the inverse-depth priors, the reference-relative frame offsets, and the depth seeds accordingly.

\paragraph{Comparison with the indirect pipeline.} The direct pipeline trades the full local mapper of \texttt{main\_slam.py} (with its triangulation, point fusion, and keyframe culling) for a lighter, synchronous design: persistent landmarks, inverse-depth seeds, and a bounded local BA. Recovery after tracking loss retries the last accepted reference rather than performing place-recognition relocalization, and the trajectory is exported as the final accepted poses after all local and loop corrections.

\hypertarget{cpp-core}{%
\subsection{C++ Core for Sparse SLAM}\label{cpp-core}}

The system provides a modular \textbf{sparse-SLAM core}, implemented in \textbf{both Python and C++} (with custom pybind11 bindings), allowing users to switch between high-performance/speed and high-flexibility modes. The C++ core reimplements the sparse SLAM initially implemented in Python, exposing core SLAM classes (frames, keyframes, map points, maps, cameras, optimizers, tracking, and local mapping) to Python via pybind11.

The C++ implementation follows a streamlined design where all core data resides in C++, with Python serving as an interface layer. C++ classes mirror their Python counterparts, maintaining identical interfaces and data field names. The bindings support zero-copy data exchange (e.g., descriptors) and safe memory ownership across the Python/C++ boundary, leveraging automatic zero-copy sharing of NumPy array memory with C++ when possible.

The module is currently \textbf{under active development}.

The C++ core is selected by default (\texttt{USE\_CPP\_CORE = True} in \texttt{pyslam/config\_parameters.py}); setting the flag to \texttt{False}, or launching with the environment variable \texttt{PYSLAM\_USE\_CPP=false}, selects the Python implementation. The two implementations are interoperable: a map saved by one can be loaded by the other. Two practical consequences follow from this design. When the C++ core is active, the Python implementation of the sparse core is bypassed, so changes to it have no effect at runtime. Conversely, changes to the C++ sources take effect only after the core is rebuilt with \texttt{./build\_cpp\_core.sh}.

The codebase is organized into the following categories:
\begin{itemize}
\item \textbf{Core SLAM Classes}: Main C++ implementations (\texttt{frame.cpp/h}, \texttt{keyframe.cpp/h}, \texttt{map\_point.cpp/h}, \texttt{map.cpp/h}, \texttt{optimizer\_g2o.cpp/h}, \texttt{tracking\_core.cpp/h}, \texttt{local\_mapping\_core.cpp/h}).
\item \textbf{SLAM pipeline}: the \texttt{Slam}, \texttt{Tracking}, and \texttt{LocalMapping} classes (\texttt{slam.cpp/h}, \texttt{tracking.cpp/h}, \texttt{local\_mapping.cpp/h}), which follow the control flow of their Python counterparts. Components that have not been ported, such as the feature tracker and loop closing, remain Python objects held by these classes.
\item \textbf{Serialization}: JSON and NumPy serialization modules for data persistence and Python interop.
\item \textbf{Python Bindings}: pybind11 module definitions that expose C++ classes to Python.
\item \textbf{Cameras} (\texttt{cpp/core/camera}): the camera classes (\texttt{camera.cpp/h}, \texttt{camera\_models.cpp/h}), the shared projection math (\texttt{camera\_model\_math.h}), \texttt{camera\_pose.cpp/h}, rig geometry, and the camera g2o edges and GTSAM factors with their bindings.
\item \textbf{Shared headers} (\texttt{cpp/core/common}): utilities shared by the camera, sparse-SLAM, direct-SLAM, and semantic components, such as Eigen aliases, smart pointers, settings dictionaries, JSON and NumPy serialization, descriptor distances, and feature-trail drawing.
\item \textbf{Type Casters}: pybind11 type casters for custom type conversions. The OpenCV, JSON, and dictionary casters are shared in \texttt{cpp/casters}.
\item \textbf{Utilities}: SLAM helper functions for geometry, features, image processing, and map access.
\item \textbf{Tests}: C++ unit tests and Python integration tests (see Sect.~\ref{testing}).
\end{itemize}

The C++ core adopts a streamlined design philosophy:
\begin{itemize}
\item \textbf{All core data resides in C++}: Python serves purely as an interface layer.
\item \textbf{Direct pybind11 exposure}: Python objects are lightweight views of underlying C++ objects.
\item \textbf{Automatic zero-copy}: pybind11 automatically shares NumPy array memory with C++ when possible.
\item \textbf{RAII-based ownership}: C++ smart pointers manage object lifetimes safely and efficiently.
\end{itemize}

\hypertarget{camera-models}{%
\subsection{Camera Models}\label{camera-models}}

The sparse pipeline accesses the camera only through its projection $\pi(\mathbf{X}_c)$, its unprojection $\pi^{-1}(\mathbf{u})$, and the analytic Jacobian $\partial \pi / \partial \mathbf{X}_c$ of the projection with respect to a point $\mathbf{X}_c = (x, y, z)^\top$ in the camera frame. Four models are available and are selected by the \texttt{Camera.type} field of the settings file. A settings file without that field keeps the linear pinhole model, so existing calibration files are unaffected. The direct pipeline (Sect.~\ref{direct-slam}) uses the same camera classes; the measurement conventions and the stereo matching described below refer to the sparse pipeline.

\paragraph{Pinhole.} The default model is
\begin{equation}
u = f_x \frac{x}{z} + c_x, \qquad v = f_y \frac{y}{z} + c_y .
\end{equation}
The distortion coefficients are used only to undistort the detected keypoints, and bundle adjustment compares the linear projection with the undistorted keypoints.

\paragraph{Kannala--Brandt.} The fisheye model of~\cite{kannala2006generic} with four distortion coefficients (\texttt{kannala\_brandt4}) maps the angle $\theta$ between the ray and the optical axis to a distorted angle $\theta_d$:
\begin{equation}
r = \sqrt{x^2 + y^2}, \quad \theta = \operatorname{atan2}(r, z), \quad
\theta_d = \theta \left(1 + k_1 \theta^2 + k_2 \theta^4 + k_3 \theta^6 + k_4 \theta^8\right),
\end{equation}
\begin{equation}
u = f_x \, \theta_d \, \frac{x}{r} + c_x, \qquad v = f_y \, \theta_d \, \frac{y}{r} + c_y .
\end{equation}
This is the polynomial used by OpenCV's fisheye module. Because the field of view can exceed $180^\circ$, a point with $z < 0$ can still have a valid projection.

\paragraph{Rational pinhole.} This model (\texttt{pinhole\_radtan8}) distorts the normalized point $x' = x/z$, $y' = y/z$ with the OpenCV rational model, whose eight coefficients are $k_1, k_2, p_1, p_2, k_3, k_4, k_5, k_6$. With $r^2 = x'^2 + y'^2$, the radial factor is a ratio of two polynomials in $r^2$ and the tangential terms are those of the Brown--Conrady model:
\begin{equation}
c(r) = \frac{1 + k_1 r^2 + k_2 r^4 + k_3 r^6}{1 + k_4 r^2 + k_5 r^4 + k_6 r^6},
\end{equation}
\begin{equation}
x'' = x' \, c(r) + 2 p_1 x' y' + p_2 \left(r^2 + 2 x'^2\right), \qquad
y'' = y' \, c(r) + p_1 \left(r^2 + 2 y'^2\right) + 2 p_2 x' y',
\end{equation}
\begin{equation}
u = f_x \, x'' + c_x, \qquad v = f_y \, y'' + c_y .
\end{equation}
The projection is defined for $z > 0$. With $k_4 = k_5 = k_6 = 0$ it reduces to the five-coefficient OpenCV model.

\paragraph{Double sphere.} The model of~\cite{usenko2018double} (\texttt{double\_sphere}) projects a point through two unit spheres whose centers are offset by $\xi$, followed by a pinhole whose image plane is shifted by $\alpha$:
\begin{equation}
d_1 = \Vert \mathbf{X}_c \Vert = \sqrt{x^2 + y^2 + z^2}, \qquad
d_2 = \sqrt{x^2 + y^2 + (\xi d_1 + z)^2},
\end{equation}
\begin{equation}
u = f_x \frac{x}{\alpha d_2 + (1 - \alpha)(\xi d_1 + z)} + c_x, \qquad
v = f_y \frac{y}{\alpha d_2 + (1 - \alpha)(\xi d_1 + z)} + c_y .
\end{equation}
A point has a valid projection only if $z > -w_2 d_1$, where
\begin{equation}
w_1 = \begin{cases} \alpha / (1 - \alpha), & \alpha \le 0.5, \\ (1 - \alpha) / \alpha, & \alpha > 0.5, \end{cases} \qquad
w_2 = \frac{w_1 + \xi}{\sqrt{2 w_1 \xi + \xi^2 + 1}} .
\end{equation}

\paragraph{Measurements.} For the pinhole model, tracking and optimization use undistorted keypoints, as in ORB-SLAM2~\cite{ORB_SLAM2}. For the three non-linear models, the projection predicts the detected pixel directly, so projection search and bundle adjustment measure the raw keypoints. For every model the framework also keeps the pinhole image of each keypoint ray, which is used by the fundamental-matrix test and by rectified stereo matching. A stereo observation is the triple $(u_l, v_l, u_r)$. With a physical stereo rig, $u_r$ is predicted by projecting the point through the calibrated right camera and its extrinsics. With an RGB-D sensor it is the virtual coordinate $u_r = f_x x/z + c_x - b f_x / z$, where $b f_x$ is the stereo baseline times the focal length.

\paragraph{Native fisheye stereo.} Rectified-row stereo matching assumes a pinhole image. For the non-linear models, stereo correspondences are instead searched on unit bearings: each left bearing defines an epipolar plane in the right camera, and a right keypoint is a candidate when its bearing lies on that plane, between the image of the point at infinity and the right camera center, within a $\chi^2$ bound (one degree of freedom) of the right keypoint's octave. The best descriptor must pass a ratio test, matches are made one-to-one, and each pair is triangulated from both rays and accepted only if it reprojects within a two-degree-of-freedom $\chi^2$ bound in both images. Triangulation search between keyframes uses the same bearing-to-plane distance.

\paragraph{Implementation.} The projection formulas and their Jacobians are written once, in a single C++ header that depends only on Eigen and is shared by the Python and C++ cores. Optimization factors for g2o~\cite{g2o_kummerle} and GTSAM~\cite{factor_graphs_for_robot_perception} are implemented in C++ with analytic Jacobians, so no per-observation Python callbacks enter the optimization. Pinhole observations keep their original optimizer factors.

\paragraph{TUM-VI.} The TUM-VI loader~\cite{schubert2018tumvi} supports two stereo modes. The \textit{native} mode keeps the original fisheye images and builds two Kannala--Brandt cameras with the calibrated right-to-left transform. The \textit{rectified} mode remaps the fisheye images once to a pinhole stereo pair, which then runs through the standard pinhole pipeline; this mode should be used with stereo depth estimators and dense reconstruction. Monocular fisheye tracking is also supported. Motion-capture poses are converted to the tracking camera frame and are never interpolated across gaps longer than a configurable threshold. Sect.~\ref{camera-model-validation} reports an initial comparison of the two modes.

\paragraph{Limitations.} Stereo depth initialization and the legacy initialization and Sim(3) paths have not yet been validated for fields of view beyond $180^\circ$. Supporting a projection model is therefore not equivalent to supporting complete omnidirectional SLAM.

\hypertarget{feature-management}{%
\subsection{Classic and Learned Local Features}\label{feature-management}}

\begin{figure}[!ht]
\begin{center}
    \includegraphics[width=0.75\textwidth]{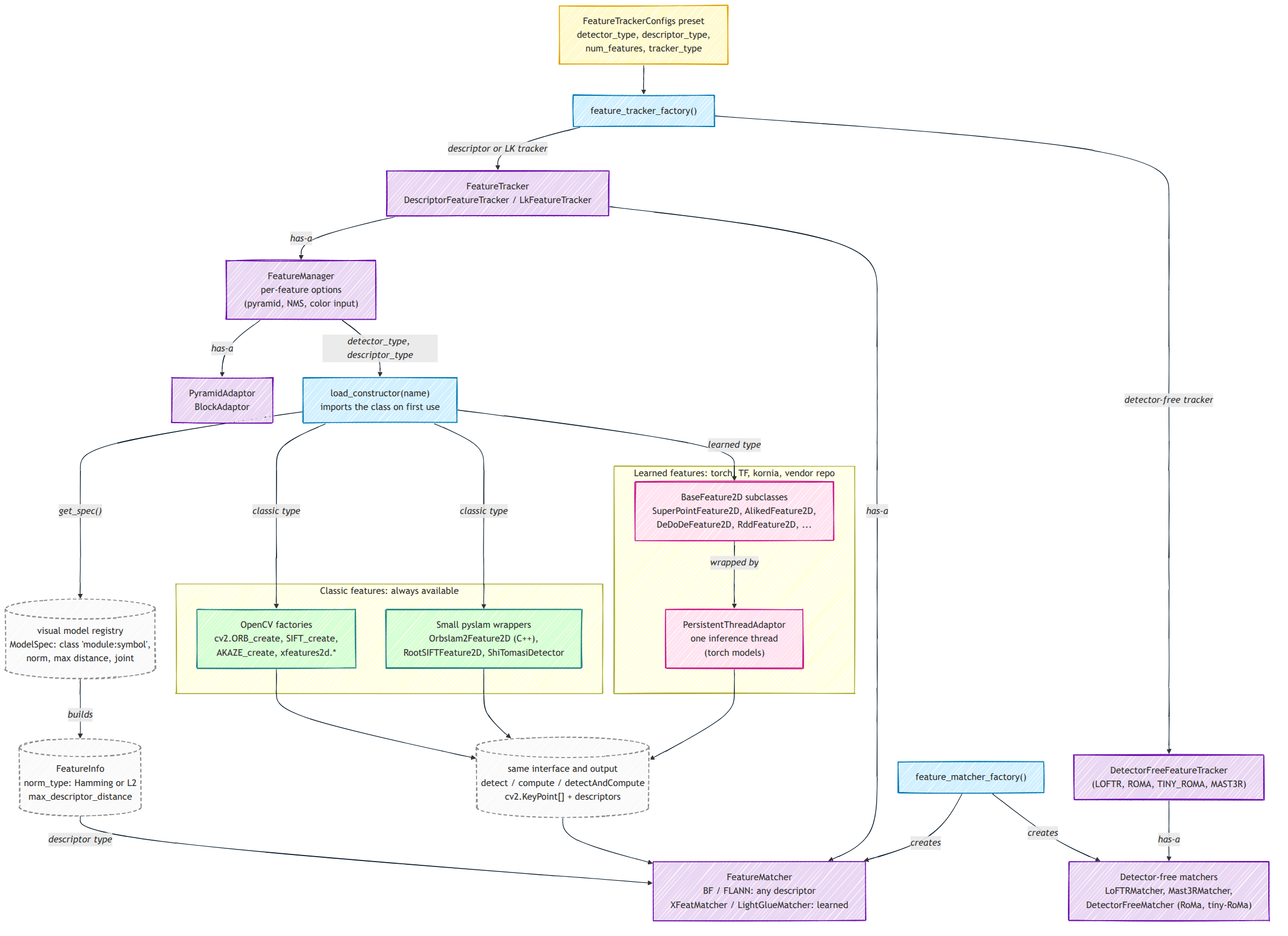}
\end{center}
\caption{How classic and learned local features are created, run, and matched.}
\label{Fig:FeatureManagement}
\end{figure}

pySLAM supports two families of local features. \textit{Classic} features (ORB, ORB2, SIFT, AKAZE, BRISK, FAST, \dots) are hand-designed algorithms provided by OpenCV or by the compiled ORB-SLAM2 extractor: they need no model weights and are always available. \textit{Learned} features (SuperPoint, ALIKED, DISK, XFeat, DeDoDe~v2, RDD, \dots) are neural networks: each one has a pySLAM wrapper that loads the network and its weights. Fig.~\ref{Fig:FeatureManagement} shows how both families go through the same path.

\begin{itemize}
\item A \texttt{FeatureTrackerConfigs} preset names the detector, the descriptor, and the tracker type. \texttt{feature\_tracker\_factory} creates the \texttt{FeatureTracker}, which owns a \texttt{FeatureManager} (extraction) and a \texttt{FeatureMatcher} (matching).
\item The \texttt{FeatureManager} dispatches on the detector and descriptor types and sets the options of each feature (pyramid, non-maximum suppression, color input). It takes the class of the feature from the visual model registry and imports it only on first use. For a classic feature, the class is an OpenCV factory function (e.g., \texttt{cv2.ORB\_create}) or one of a few small pySLAM wrappers (\texttt{Orbslam2Feature2D}, \texttt{RootSIFTFeature2D}, \texttt{ShiTomasiDetector}). For a learned feature, it is a pySLAM wrapper, so that PyTorch, TensorFlow, or a vendor repository is loaded only when a learned feature is selected. Learned wrappers derive from \texttt{BaseFeature2D}, and torch-based ones run on a dedicated inference thread (\texttt{PersistentThreadAdaptor}).
\item Both families expose the same \texttt{detect}, \texttt{compute}, and \texttt{detectAndCompute} methods and return the same output: a list of \texttt{cv2.KeyPoint} and an array of descriptors. From this point on, the pipeline treats them in the same way.
\item \texttt{FeatureInfo} tells the matcher how to compare descriptors: Hamming distance for binary descriptors (e.g., ORB), L2 distance for float descriptors (e.g., SIFT and the learned ones), and a calibrated maximum descriptor distance; its tables are built from the registry. Brute-force and FLANN matchers work with any descriptor; XFeat and LightGlue are learned matchers for their own features.
\item Detector-free trackers (LoFTR, MASt3R, RoMa, tiny-RoMa) share one class, \texttt{DetectorFreeFeatureTracker}: its matcher computes matched keypoints directly from the two images. Its \texttt{FeatureManager} is built with the \texttt{NONE} detector and descriptor, so it loads no network and no keypoints are extracted on single images.
\end{itemize}

\hypertarget{visual-model-catalog}{%
\subsection{Visual Model Catalog}\label{visual-model-catalog}}

\begin{figure}[!ht]
\begin{center}
    \includegraphics[width=\textwidth]{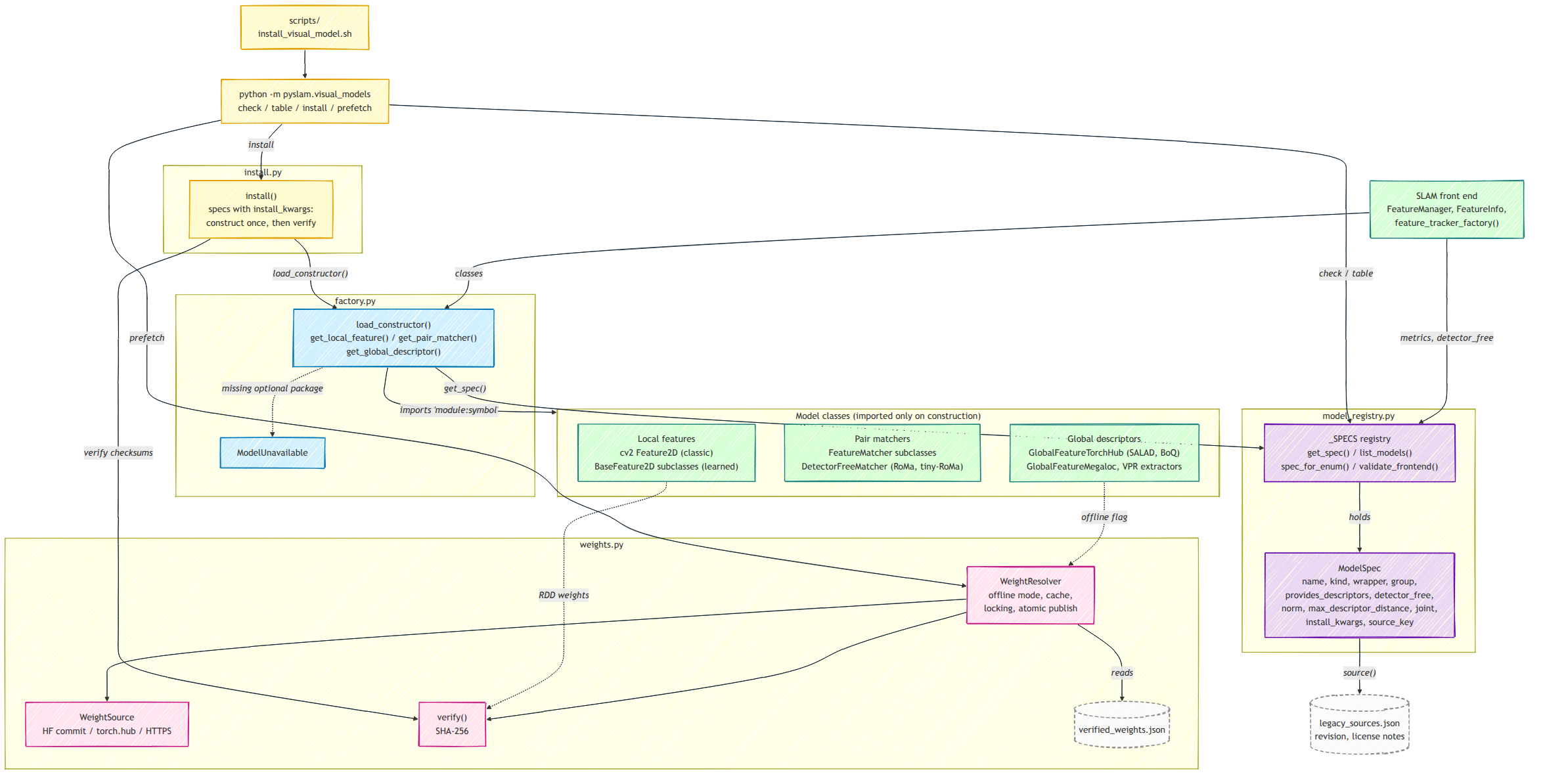}
\end{center}
\caption{Classes of the \texttt{pyslam.visual\_models} catalog.}
\label{Fig:VisualModelCatalog}
\end{figure}

The \texttt{pyslam.visual\_models} catalog lists every visual model integrated in pySLAM: local features, pair matchers, and global descriptors for loop closure. It holds only metadata, so reading it never loads a neural network or an inference framework. It is the single place where a model is declared: the \texttt{FeatureManager} of Sect.~\ref{feature-management} takes its classes from it, \texttt{FeatureInfo} is built from its descriptor metrics, and \texttt{feature\_tracker\_factory} uses its detector-free flag. The same records are used to inspect models, install their weights, and test the integrations. Fig.~\ref{Fig:VisualModelCatalog} shows its classes:

\begin{itemize}
\item \texttt{ModelSpec} is the record of one model: its name, its kind (\texttt{local\_feature}, \texttt{pair\_matcher}, or \texttt{global\_descriptor}), the \texttt{module:symbol} of its class, its dependency group, its capabilities (e.g., whether it produces per-keypoint descriptors or is detector-free), its descriptor metric (norm and maximum descriptor distance), and, for self-installing models, the arguments used to prefetch it. The registry holds all records and answers queries such as \texttt{get\_spec()} and \texttt{list\_models()}. A record also reports the provenance and license of the model.
\item The factory (\texttt{get\_local\_feature()}, \texttt{get\_pair\_matcher()}, \texttt{get\_global\_descriptor()}) builds a model from its name: it looks up the record, imports the class only at that moment (\texttt{load\_constructor()}, also used by the \texttt{FeatureManager} and the installer), and constructs it. If an optional package is missing, it raises \texttt{ModelUnavailable} with the dependency group to install.
\item The weight resolver (\texttt{WeightResolver}, \texttt{WeightSource}, \texttt{verify()}) downloads weights from a pinned source (a Hugging Face commit, torch.hub, or HTTPS), stores them in a cache, and checks their registered SHA-256 checksums. It also supports an offline mode.
\item The installer prefetches the models integrated through pinned sources (the records with prefetch arguments): it constructs each model once through the factory, which downloads its code and weights, and then verifies the checksums.
\item The command-line tool \texttt{python -m pyslam.visual\_models}, which the script \texttt{install\_visual\_model.sh} in \texttt{scripts/} also calls, ties these together: \texttt{check} and \texttt{table} read the registry, \texttt{install} runs the installer, and \texttt{prefetch} fetches one registered weight file through the resolver.
\end{itemize}

Appendix~\ref{visual-models-appendix} describes how to install the models, their runtime behavior and optional accelerations, and their licenses.

\hypertarget{scene-from-views}{%
\subsection{3D Scene from Views}\label{scene-from-views}}

\begin{figure}[!ht]
\begin{center}
    \includegraphics[width=\textwidth]{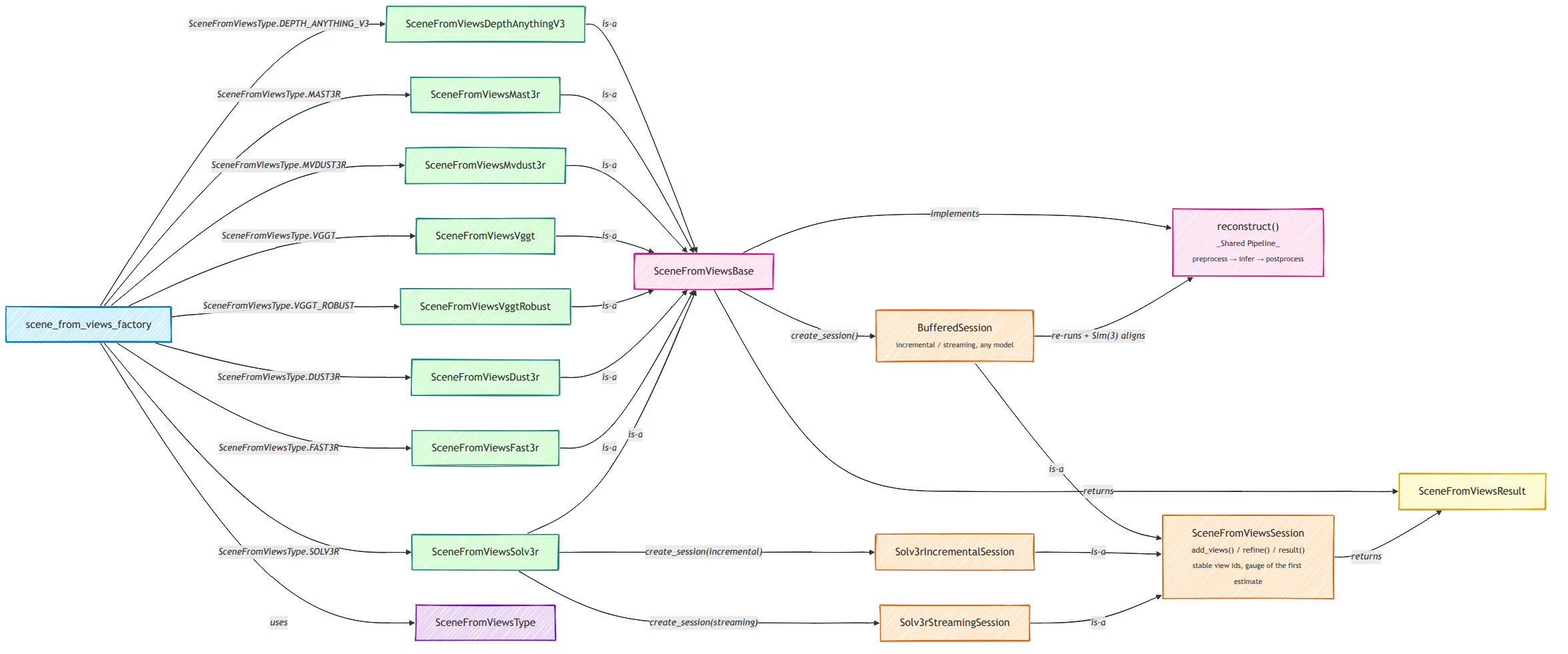}
\end{center}
\caption{\textit{SceneFromViews} architecture.}
\label{Fig:SceneFromViews}
\end{figure}

The diagram in Fig.~\ref{Fig:SceneFromViews} illustrates the architecture of the \textit{Scene from Views} module, which provides a unified interface for feed-forward 3D scene reconstruction from multiple views. At its core, the \texttt{scene\_from\_views\_factory} instantiates specific reconstruction models based on the selected \texttt{SceneFromViewsType}, such as \texttt{DUST3R}\cite{wang2023dust3r}, \texttt{MAST3R}\cite{leroy2024grounding}, \texttt{DEPTH\_ANYTHING\_V3}\cite{depthanything3}, \texttt{MVDUST3R}\cite{tang2025mv}, \texttt{VGGT}\cite{wang2025vggt}, \texttt{VGGT\_ROBUST}\cite{vggt_robust}, \texttt{FAST3R}\cite{yang2025fast3r}, and \texttt{SOLV3R}\cite{kwak2026solv3r}.

Each type creates a corresponding implementation (e.g., \texttt{SceneFromViewsDust3r}, \texttt{SceneFromViewsMast3r}, \texttt{SceneFromViewsDepthAnythingV3}, \texttt{SceneFromViewsMvdust3r}, \texttt{SceneFromViewsVggt}, \texttt{SceneFromViewsVggtRobust}, \texttt{SceneFromViewsFast3r}, \texttt{SceneFromViewsSolv3r}), all inheriting from a common \texttt{SceneFromViewsBase}. This base class implements a unified three-step reconstruction pipeline: \texttt{preprocess\_images()} prepares input images for the specific model, \texttt{infer()} runs model inference, and \texttt{postprocess\_results()} converts raw model output to a standardized \texttt{SceneFromViewsResult} format containing merged point clouds, meshes, camera poses, and optional depth maps or intrinsics.

The module supports both pairwise models (\texttt{DUSt3R}\cite{wang2023dust3r}, \texttt{MASt3R}\cite{leroy2024grounding}) that are feed-forward per pair and then run global alignment, as well as multi-view models (\texttt{DEPTH\_ANYTHING\_V3}\cite{depthanything3}, \texttt{MV-DUSt3R}\cite{tang2025mv}, \texttt{VGGT}\cite{wang2025vggt}, \texttt{Robust VGGT}\cite{vggt_robust}, \texttt{FAST3R}\cite{yang2025fast3r}) that process multiple views simultaneously in a single forward pass. \texttt{SOLV3R}\cite{kwak2026solv3r} is a multi-view solver: it keeps one state per view and refines all states with the same looped Transformer block at every call, so a solve can be continued, extended with new views, or partly held fixed. This modular design enables flexible integration of diverse 3D reconstruction techniques while maintaining a consistent API across different model architectures.

\paragraph{Sessions.} \texttt{reconstruct()} is stateless: it reconstructs a fixed set of images. For views that arrive over time, such as the keyframes of a SLAM run, \texttt{create\_session()} returns a \texttt{SceneFromViewsSession} that keeps one reconstruction up to date. Views are added with \texttt{add\_views()} and keep stable identifiers; \texttt{result()} returns the current estimate of every view, the views changed by the last update, and the pose of each view when it was first estimated. In the \textit{incremental} mode every view is kept and every estimate can change; in the \textit{streaming} mode the first views anchor the solve, the latest ones are refined, and the views that leave a bounded window keep their last estimate, so memory and time per update stay bounded. All estimates of a session are expressed in the frame and scale of its first estimate: feed-forward models answer each run in their own gauge, so each update is aligned to the previous estimates with a similarity transform (rotation from the chordal mean over the overlapping views, scale from the median depth ratio). Every model gets a generic session, which re-runs \texttt{reconstruct()} on the buffered views or on a sliding window; \texttt{SceneFromViewsSolv3r} extends its solve natively, appending new views to the ongoing solve (incremental) or using SOLV3R's streaming solver with anchors, a strided memory of held frames, and a refined recent window (streaming).

The package and the factory import the model implementations lazily: importing \texttt{pyslam.scene\_from\_views} or the factory loads no model package, and only the selected model is imported when it is constructed.

\FloatBarrier

\hypertarget{supported-components-and-models}{%
\section{Supported Components and
Models}\label{supported-components-and-models}}

\hypertarget{supported-local-features}{%
\subsection{Supported Local
Features}\label{supported-local-features}}

\begin{figure}[!t]
\begin{center}
    \includegraphics[width=\textwidth]{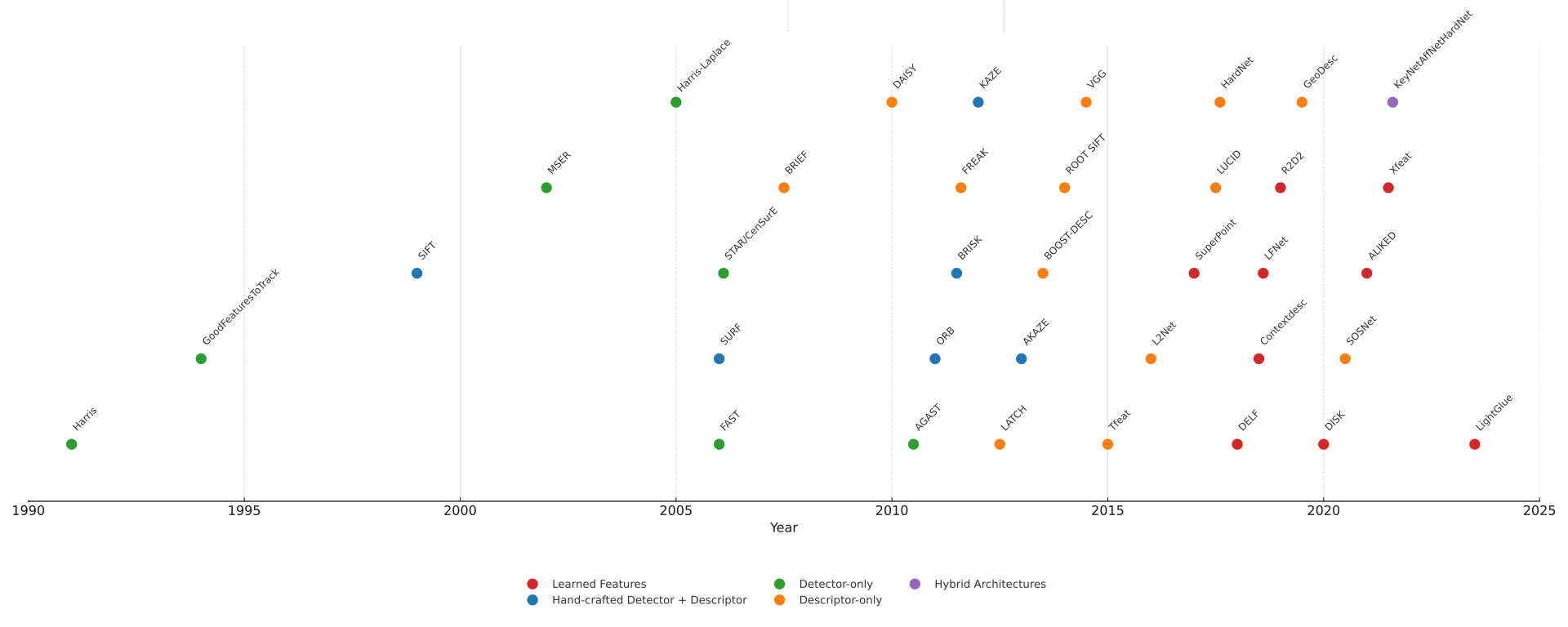}
\end{center}
\caption{Timeline of some of the most famous local features for image matching, place recognition and SLAM.}
\label{Fig_TimelineLocalFeatures}
\end{figure}

Fig.~\ref{Fig_TimelineLocalFeatures} shows a timeline with some of the most famous local features for image matching, place recognition and SLAM. At present time, pySLAM supports the following feature \textbf{detectors}:

\begin{itemize}
    \item \href{https://www.edwardrosten.com/work/fast.html}{FAST} \cite{rosten2006machine}
    \item \href{https://ieeexplore.ieee.org/document/323794}{Good features to track} \cite{shi1994good}
    \item \href{http://www.willowgarage.com/sites/default/files/orb_final.pdf}{ORB} \cite{rublee2011orb}
    \item \href{https://github.com/raulmur/ORB_SLAM2}{ORB2} (improvements of ORB-SLAM2\cite{ORB_SLAM2} to ORB detector)
    \item \href{https://www.cs.ubc.ca/~lowe/papers/iccv99.pdf}{SIFT} \cite{lowe1999object}
    \item \href{http://people.ee.ethz.ch/~surf/eccv06.pdf}{SURF} \cite{bay2006surf}
    \item \href{https://www.doc.ic.ac.uk/~ajd/Publications/alcantarilla_etal_eccv2012.pdf}{KAZE} \cite{alcantarilla2012kaze}
    \item \href{http://www.bmva.org/bmvc/2013/Papers/paper0013/paper0013.pdf}{AKAZE} \cite{alcantarilla2013fast}
    \item \href{http://www.margaritachli.com/papers/ICCV2011paper.pdf}{BRISK} \cite{leutenegger2011brisk}
    \item \href{chrome-extension://efaidnbmnnnibpcajpcglclefindmkaj/https://archive.air.in.tum.de/Main/Publications/Mair2010c.pdf}{AGAST} \cite{mair2010adaptive}
    \item \href{http://cmp.felk.cvut.cz/~matas/papers/matas-bmvc02.pdf}{MSER} \cite{matas2002robust}
    \item \href{https://link.springer.com/chapter/10.1007/978-3-540-88693-8_8}{StarDector/CenSurE} \cite{censure}
    \item \href{https://www.robots.ox.ac.uk/~vgg/research/affine/det_eval_files/mikolajczyk_ijcv2004.pdf}{Harris-Laplace} \cite{mikolajczyk_inria00548554}
    \item \href{https://github.com/MagicLeapResearch/SuperPointPretrainedNetwork}{SuperPoint}\cite{detone18superpoint}
    \item \href{https://github.com/mihaidusmanu/d2-net}{D2-Net} \cite{dusmanu2019d2}
    \item \href{https://github.com/tensorflow/models/tree/master/research/delf}{DELF} \cite{noh2017large}
    \item \href{https://github.com/lzx551402/contextdesc}{Contextdesc} \cite{luo2020contextdesc}
    \item \href{https://github.com/vcg-uvic/lf-net-release}{LFNet} \cite{ono2018lf}
    \item \href{https://github.com/naver/r2d2}{R2D2} \cite{revaud2019r2d2}
    \item \href{https://github.com/axelBarroso/Key.Net}{Key.Net} \cite{barroso2020key}
    \item \href{https://arxiv.org/abs/2006.13566}{DISK} \cite{tyszkiewicz2020disk}
    \item \href{https://arxiv.org/abs/2304.03608}{ALIKED} \cite{barroso2023alike}
    \item \href{https://arxiv.org/abs/2404.19174}{Xfeat} \cite{barroso2024xfeat}
    \item \href{https://github.com/axelBarroso/Key.Net}{KeyNetAffNetHardNet} (KeyNet detector \cite{barroso2020key} + AffNet + HardNet descriptor \cite{mishchuk2017working})
    \item \href{https://github.com/Parskatt/DeDoDe}{DeDoDe v2} \cite{edstedt2024dedode,edstedt2024dedodev2} (via kornia)
    \item \href{https://github.com/xtcpete/rdd}{RDD} \cite{chen2025rdd}
\end{itemize}

The following feature \textbf{descriptors} are supported: 

\begin{itemize}
    \item \href{http://www.willowgarage.com/sites/default/files/orb_final.pdf}{ORB} \cite{rublee2011orb}
    \item \href{https://www.cs.ubc.ca/~lowe/papers/iccv99.pdf}{SIFT} \cite{lowe1999object}
    \item \href{https://www.robots.ox.ac.uk/~vgg/publications/2012/Arandjelovic12/arandjelovic12.pdf}{ROOT SIFT} \cite{arandjelovic2012three}
    \item \href{http://people.ee.ethz.ch/~surf/eccv06.pdf}{SURF} \cite{bay2006surf}
    \item \href{http://www.bmva.org/bmvc/2013/Papers/paper0013/paper0013.pdf}{AKAZE} \cite{alcantarilla2013fast}
    \item \href{http://www.margaritachli.com/papers/ICCV2011paper.pdf}{BRISK} \cite{leutenegger2011brisk}
    \item \href{https://www.researchgate.net/publication/258848394_FREAK_Fast_retina_keypoint}{FREAK} \cite{alahi2012freak}
    \item \href{https://github.com/MagicLeapResearch/SuperPointPretrainedNetwork}{SuperPoint} \cite{detone18superpoint}
    \item \href{https://github.com/vbalnt/tfeat}{Tfeat} \cite{balntas2016learning}
    \item \href{https://www.labri.fr/perso/vlepetit/pubs/trzcinski_pami15.pdf}{BOOST-DESC} \cite{trzcinski2013boosting}
    \item \href{https://ieeexplore.ieee.org/document/4815264}{DAISY} \cite{tola2010daisy}
    \item \href{https://arxiv.org/abs/1501.03719}{LATCH} \cite{levi2006latch}
    \item \href{https://pdfs.semanticscholar.org/85bd/560cdcbd4f3c24a43678284f485eb2d712d7.pdf}{LUCID} \cite{lucid2015}
    \item \href{https://www.robots.ox.ac.uk/~vedaldi/assets/pubs/simonyan14learning.pdf}{VGG} \cite{simonyan2014learning}
    \item \href{https://github.com/DagnyT/hardnet.git}{Hardnet} \cite{mishchuk2017working}
    \item \href{https://github.com/lzx551402/geodesc.git}{GeoDesc} \cite{verdie2018tilde}
    \item \href{https://github.com/yuruntian/SOSNet.git}{SOSNet} \cite{SOSNet}
    \item \href{https://github.com/yuruntian/L2-Net}{L2Net} \cite{L2Net}
    \item \href{https://github.com/cvlab-epfl/log-polar-descriptors}{Log-polar descriptor} \cite{LogPolarDescriptor}
    \item \href{https://github.com/mihaidusmanu/d2-net}{D2-Net} \cite{dusmanu2019d2}
    \item \href{https://github.com/tensorflow/models/tree/master/research/delf}{DELF} \cite{noh2017large}
    \item \href{https://github.com/lzx551402/contextdesc}{Contextdesc} \cite{luo2020contextdesc}
    \item \href{https://github.com/vcg-uvic/lf-net-release}{LFNet} \cite{ono2018lf}
    \item \href{https://github.com/naver/r2d2}{R2D2} \cite{revaud2019r2d2}
    \item \href{https://raw.githubusercontent.com/iago-suarez/BEBLID/master/BEBLID_Boosted_Efficient_Binary_Local_Image_Descriptor.pdf}{BEBLID} \cite{BEBLID}
    \item \href{https://arxiv.org/abs/2006.13566}{DISK} \cite{tyszkiewicz2020disk}
    \item \href{https://arxiv.org/abs/2304.03608}{ALIKED} \cite{barroso2023alike}
    \item \href{https://arxiv.org/abs/2404.19174}{Xfeat} \cite{barroso2024xfeat}
    \item \href{https://github.com/axelBarroso/Key.Net}{KeyNetAffNetHardNet} (KeyNet detector \cite{barroso2020key} + AffNet + HardNet descriptor \cite{mishchuk2017working})
    \item \href{https://github.com/Parskatt/DeDoDe}{DeDoDe v2} \cite{edstedt2024dedode,edstedt2024dedodev2} (via kornia)
    \item \href{https://github.com/xtcpete/rdd}{RDD} \cite{chen2025rdd}
\end{itemize}

For more information, refer to \href{https://github.com/luigifreda/pyslam/blob/master/pyslam/local_features/feature_types.py}{pyslam/local\_features/feature\_types.py}
file. Some of the local features consist of a \emph{joint
detector-descriptor}.

\textbf{N.B.}: You just need a \emph{single} python environment to be
able to work with all the
\protect\hyperlink{supported-local-features}{supported local features}!

\hypertarget{supported-matchers}{%
\subsection{Supported Matchers}\label{supported-matchers}}

\begin{itemize}
    \item \texttt{BF}: Brute force matcher on descriptors (with KNN).
    \item \href{https://www.semanticscholar.org/paper/Fast-Approximate-Nearest-Neighbors-with-Automatic-Muja-Lowe/35d81066cb1369acf4b6c5117fcbb862be2af350}{FLANN} \cite{muja2009fast}
    \item \href{https://arxiv.org/abs/2404.19174}{XFeat} \cite{barroso2024xfeat}
    \item \href{https://arxiv.org/abs/2306.13643}{LightGlue} \cite{lindenberger2023lightglue}
    \item \href{https://arxiv.org/abs/2104.00680}{LoFTR} \cite{sun2021loftr}
    \item \href{https://arxiv.org/abs/2406.09756}{MASt3R} \cite{master}
    \item \href{https://github.com/Parskatt/RoMa}{RoMa} and tiny-RoMa \cite{edstedt2024roma} (dense, detector-free)
\end{itemize}

See the file \texttt{pyslam/local\_features/feature\_matcher.py} for further
details.

\hypertarget{supported-global-descriptors-and-local-descriptor-aggregation-methods}{%
\subsection{Supported Global Descriptors and Local Descriptor
Aggregation
Methods}\label{supported-global-descriptors-and-local-descriptor-aggregation-methods}}

\hypertarget{local-descriptor-aggregation-methods}{%
\subparagraph{Local descriptor aggregation
methods}\label{local-descriptor-aggregation-methods}}

\begin{itemize}
    \item Bag of Words (BoW): \href{https://github.com/dorian3d/DBoW2}{DBoW2} \cite{galvez2012bags}, \href{https://github.com/rmsalinas/DBow3}{DBoW3} \cite{galvez2012bags}.  \href{https://doi.org/10.1109/TRO.2012.2197158}{[paper]}
    \item Vector of Locally Aggregated Descriptors: \href{https://www.vlfeat.org/api/vlad.html}{VLAD} \cite{jegou2010aggregating}.  \href{https://doi.org/10.1109/CVPR.2010.5540039}{[paper]}
    \item Incremental Bags of Binary Words (iBoW) via Online Binary Image Index: \href{https://github.com/emiliofidalgo/ibow-lcd}{iBoW} \cite{garcia2018ibow}, \href{https://github.com/emiliofidalgo/obindex2}{OBIndex2} \cite{garcia2018ibow}.  \href{https://doi.org/10.1109/LRA.2018.2849609}{[paper]}
    \item Hyperdimensional Computing: \href{https://www.tu-chemnitz.de/etit/proaut/hdc_desc}{HDC} \cite{neubert2021hyperdimensional}.  \href{https://openaccess.thecvf.com/content/CVPR2021/html/Neubert_Hyperdimensional_Computing_as_a_Framework_for_Systematic_Aggregation_of_Image_CVPR_2021_paper.html}{[paper]}
\end{itemize}

\textbf{NOTE}: \emph{iBoW} and \emph{OBIndex2} incrementally build a
binary image index and do not need a prebuilt vocabulary. In the
implemented classes, when needed, the input non-binary local descriptors
are transparently transformed into binary descriptors.

\hypertarget{global-descriptors}{%
\subparagraph{Global descriptors}\label{global-descriptors}}

Also referred to as \emph{holistic descriptors}:

\begin{itemize}
\tightlist
\item
  \href{https://ieeexplore.ieee.org/document/6224623}{SAD} \cite{milford2012seqslam}
\item
  \href{https://github.com/BVLC/caffe/tree/master/models/bvlc_alexnet}{AlexNet} \cite{krizhevsky2012imagenet}
\item
  \href{https://www.di.ens.fr/willow/research/netvlad/}{NetVLAD} \cite{arandjelovic2016netvlad}
\item
  \href{https://www.tu-chemnitz.de/etit/proaut/hdc_desc}{HDC-DELF} \cite{schlegel2022comparison}
\item
  \href{https://github.com/gmberton/CosPlace}{CosPlace} \cite{berton2023cosplace}
\item
  \href{https://github.com/gmberton/EigenPlaces}{EigenPlaces} \cite{berton2023eigenplaces}
\item
  \href{https://github.com/gmberton/MegaLoc}{MegaLoc} \cite{berton2025megaloc}
\item
  \href{https://github.com/serizba/salad}{SALAD} \cite{izquierdo2024salad}
\item
  \href{https://github.com/amaralibey/Bag-of-Queries}{BoQ} \cite{alibey2024boq}
\end{itemize}

Different \protect\hyperlink{loop-closing}{loop closing methods} are
available. These combines the above aggregation methods and global
descriptors. See the file \href{https://github.com/luigifreda/pyslam/blob/master/pyslam/loop_closing/loop_detector_configs.py}{pyslam/loop\_closing/loop\_detector\_configs.py}
for further details.

\hypertarget{supported-depth-prediction-models}{%
\subsection{Supported Depth Prediction
Models}\label{supported-depth-prediction-models}}

Both monocular and stereo depth prediction models are available. SGBM
algorithm has been included as a classic reference approach.

\begin{itemize}
\tightlist
\item
  \href{https://ieeexplore.ieee.org/document/4359315}{SGBM}\cite{hirschmuller2007stereo}: Depth SGBM
  from OpenCV (Stereo, classic approach) 
\item
  \href{https://arxiv.org/abs/2410.02073}{Depth-Pro} (Monocular) \cite{depthpro2023}
\item
  \href{https://arxiv.org/abs/2611.26603}{Depth Pro 2} (Monocular, optionally temporally consistent over video) \cite{depthpro2_2026}. Its weights are licensed for non-commercial research only.
\item
  \href{https://arxiv.org/abs/2406.09414}{DepthAnythingV2} (Monocular) \cite{depthanythingv2_2024}
\item
  \href{https://arxiv.org/abs/2511.10647}{DepthAnythingV3} (Monocular) \cite{depthanything3}
\item
  \href{https://arxiv.org/abs/2109.07547}{RAFT-Stereo} (Stereo) \cite{teed2021raft}
\item
  \href{https://arxiv.org/abs/2203.11483}{CREStereo} (Stereo) \cite{li2022cres}
\item
  \href{https://arxiv.org/abs/2406.09756}{MASt3R} (Monocular/Stereo) \cite{master}
\item
  \href{https://arxiv.org/abs/2412.06974}{MV-DUSt3R} (Monocular/Stereo) \cite{mvduster}    
\end{itemize}

\hypertarget{supported-volumetric-mapping-methods}{%
\subsection{Supported Volumetric Mapping
Methods}\label{supported-volumetric-mapping-methods}}

\begin{itemize}
\tightlist
\item
  A C++ template-based voxel grid implementation leverages parallel
  spatial hashing, supporting both direct voxel hashing and indirect
  voxel-block hashing strategies. Parallel execution is managed using
  TBB, and the design accommodates both simple and semantic voxels.
  Further information about the volumetric grid models is available
  \href{https://github.com/luigifreda/pyslam/blob/master/cpp/modules/volumetric/README.md}{here}.
\item    
  \href{https://arxiv.org/pdf/2110.00511}{TSDF} with voxel block grid
  (parallel spatial hashing)~\cite{dong2022ash}
\item
  Incremental 3D Gaussian Splatting. See
  \href{https://repo-sam.inria.fr/fungraph/3d-gaussian-splatting/}{here}
  and \href{https://arxiv.org/abs/2312.06741}{MonoGS} for a description
  of its backend \cite{matsuki2023gaussian, kerbl20233d}.
\end{itemize}

\hypertarget{supported-semantic-segmentation-methods}{%
\subsection{Supported Semantic Segmentation
Methods}\label{supported-semantic-segmentation-methods}}

\begin{itemize}
\tightlist
\item
  \href{https://arxiv.org/abs/1706.05587}{DeepLabv3}~\cite{chen2017rethinking}: from \texttt{torchvision}, pre-trained on COCO/VOC.
\item
  \href{https://arxiv.org/abs/2105.15203}{Segformer}~\cite{xie2021segformer}: from \texttt{transformers}, pre-trained on Cityscapes or ADE20k.
\item
  \href{https://arxiv.org/abs/2212.09506}{CLIP}~\cite{lin2023clip}: from \texttt{f3rm} package for open-vocabulary support.
\item
  \href{https://arxiv.org/abs/2412.08628v1}{EOV\_SEG}~\cite{eov_seg}: EOV-Seg: Efficient Open-Vocabulary Panoptic Segmentation.
\item
  \href{https://arxiv.org/abs/2201.02605}{DETIC}~\cite{detic}: Detecting Twenty-thousand Classes using Image-level Supervision.
\item
  \href{https://arxiv.org/abs/2303.04803}{ODISE}~\cite{odise}: Open-Vocabulary Panoptic Segmentation with Text-to-Image Diffusion Models.
\item
  \href{https://arxiv.org/abs/2511.09554}{RFDETR}~\cite{rfdetr}: A real-time transformer architecture for object detection and instance segmentation, built on a DINOv2 vision transformer backbone.
\item
  \href{https://arxiv.org/abs/2510.09653}{YOLO}~\cite{sapkota2025ultralytics}: One of the most influential and widely adopted series of models for real-time object detection, that attains an excellent balance
between high accuracy and inference speed.
\end{itemize}
\hypertarget{supported-models-for-feed-forward-inference-of-3d-scenes-from-multiple-images}{%
\subsection{Supported Models for Feed-Forward Inference of 3D Scenes
from Multiple
Images}\label{supported-models-for-feed-forward-inference-of-3d-scenes-from-multiple-images}}

\begin{itemize}
\tightlist
\item
  \href{https://arxiv.org/abs/2312.14132}{DUSt3R}\cite{wang2023dust3r}: Geometric 3D Vision
  Made Easy
\item
  \href{https://arxiv.org/abs/2406.09756}{MASt3R}\cite{master}: Grounding Image
  Matching in 3D with MASt3R 
\item
  \href{https://arxiv.org/abs/2511.10647}{Depth Anything V3}\cite{depthanything3}: Recovering
  the Visual Space from Any Views
\item
  \href{https://github.com/naver/mvdust3r}{MVDust3r}\cite{mvduster}: Multi-view DUSt3R
  variant
\item
  \href{https://github.com/facebookresearch/vggt}{VGGT}\cite{wang2025vggt}: Visual Geometry
  Grounded Transformer
\item
  \href{https://github.com/cvlab-kaist/RobustVGGT.git}{Robust VGGT}\cite{vggt_robust}:
  Emergent Outlier View Rejection in Visual Geometry Grounded
  Transformers
\item
  \href{https://arxiv.org/abs/2501.13928}{Fast3R}\cite{yang2025fast3r}: Fast3R: Towards 3D Reconstruction of 1000+ Images in One Forward Pass
\item
  \href{https://arxiv.org/abs/2611.02857}{SOLV3R}\cite{kwak2026solv3r}: 3D Looped Transformer as a Multi-view Solver
\end{itemize}

See \href{https://github.com/luigifreda/pyslam/blob/master/pyslam/scene_from_views/README.md}{here} for further details about
\texttt{SceneFromViews} architecture.

\hypertarget{model-licenses}{%
\subsection{Model Licenses}\label{model-licenses}}

The visual model catalog (Sect.~\ref{visual-model-catalog}) records the license of the upstream code and weights of each model. Licenses should be checked before any commercial use or redistribution: some of the models above, including SuperPoint, R2D2, LF-Net, MASt3R, and Depth Pro~2, are restricted to non-commercial or research use. Appendix~\ref{visual-model-licenses} gives the details.


\hypertarget{evaluating-slam}{%
\section{Experimental Evaluation}\label{evaluating-slam}}

This section describes the tools pySLAM provides for evaluating SLAM configurations, reports results obtained with them on the TUM, EuRoC, and KITTI benchmarks, and presents the automated test suite and an initial validation of the camera models on TUM-VI.

\hypertarget{run-a-slam-evaluation}{%
\subsection{Running a SLAM Evaluation}\label{run-a-slam-evaluation}}

The \texttt{main\_slam\_evaluation.py} script enables automated SLAM
evaluation by executing \texttt{main\_slam.py} across a collection of
\textit{datasets} and configuration \textit{presets}. The main input to the script is an evaluation configuration file (e.g., \texttt{evaluation/configs/evaluation.json}) that specifies which datasets and presets to be used. For convenience, sample configurations for the datasets \texttt{TUM}, \texttt{EUROC} and \texttt{KITTI} datasets are already provided in the \texttt{evaluation/configs/} directory.

For each evaluation run, results are
stored in a dedicated subfolder within the \texttt{results} directory,
containing all the computed metrics. These metrics are then processed
and compared. The final output is a report, available in PDF,
LaTeX, and HTML formats, that includes comparison tables summarizing the
\emph{Absolute Trajectory Error} (ATE), the maximum deviation from the
ground truth trajectory and other metrics.

Evaluation results are presented in the following subsections. For convenience, they are also available at this \href{https://github.com/luigifreda/pyslam/blob/master/docs/evaluations/evaluations.md}{link}. The evaluation uses commit \textit{f7c9a13} with the following presets: \texttt{baseline}, \texttt{root\_sift}, and \texttt{superpoint}, which correspond to local feature extractors \texttt{ORB2}, \texttt{root\_SIFT}, and \texttt{SuperPoint}, respectively.

pySLAM saves both the \textit{online} and the \textit{final} trajectory estimates (Appendix~\ref{trajectory-saving}), whereas ORB-SLAM3 saves only the final ones, recorded after the whole dataset has been processed. This \href{https://github.com/anathonic/Trajectory-Comparison-ORB-SLAM3-pySLAM/blob/main/trajectories_comparison.ipynb}{notebook} compares the online trajectory of pySLAM with the final trajectory of ORB-SLAM3. In the GUI, the \texttt{Draw\ Ground\ Truth} button (Appendix~\ref{slam-gui}) plots the history of the Absolute Trajectory Error, so that both online and final errors can be followed up to the current time.

\hypertarget{tum-evaluation}{%
\subsection{TUM Evaluation}\label{tum-evaluation}}

\begin{minipage}{\textwidth}
\noindent
\captionsetup{type=table}
\captionof{table}{TUM: Table RMSE}\label{tab:tum_table_rmse}

\fontsize{7pt}{8pt}\selectfont

\begin{tabularx}{\linewidth}{ >{\RaggedRight\arraybackslash}p{ 8.0cm } >{\RaggedRight\arraybackslash}p{ 1.0cm } >{\RaggedRight\arraybackslash}p{ 1.25cm } >{\RaggedRight\arraybackslash}p{ 1.5cm }  }\toprule
Dataset & baseline & root\_sift & superpoint \\
\midrule
rgbd\_dataset\_freiburg1\_desk & 0.06634 & 0.05289 & 0.04815 \\
rgbd\_dataset\_freiburg1\_desk2 & 0.04645 & 0.06255 & 0.03868 \\
rgbd\_dataset\_freiburg1\_room & 0.05885 & 0.07968 & 0.07199 \\
rgbd\_dataset\_freiburg1\_xyz & 0.01196 & 0.01758 & 0.01726 \\
rgbd\_dataset\_freiburg3\_long\_office\_household & 0.00921 & 0.00982 & 0.00952 \\
rgbd\_dataset\_freiburg3\_nostructure\_texture\_far & 0.08397 & 0.11342 & 0.12118 \\
rgbd\_dataset\_freiburg3\_nostructure\_texture\_near\_withloop & 0.02245 & 0.05022 & 0.05037 \\
Average & 0.04275 & 0.05517 & 0.05102 \\
Std Dev & 0.0362 & 0.04531 & 0.04503 \\
Best (Average) Preset & baseline &  &  \\
Best (Average) Metric & 0.04275 &  &  \\

\bottomrule
\end{tabularx}
\end{minipage}

\bigskip

\begin{minipage}{\textwidth}
\noindent
\captionsetup{type=table}
\captionof{table}{TUM: Table Max }\label{tab:tum_table_max}

\fontsize{7pt}{8pt}\selectfont

\begin{tabularx}{\linewidth}{ >{\RaggedRight\arraybackslash}p{ 8.0cm } >{\RaggedRight\arraybackslash}p{ 1.0cm } >{\RaggedRight\arraybackslash}p{ 1.25cm } >{\RaggedRight\arraybackslash}p{ 1.5cm }  }\toprule
Dataset & baseline & root\_sift & superpoint \\
\midrule
rgbd\_dataset\_freiburg1\_desk & 0.326 & 0.19361 & 0.22686 \\
rgbd\_dataset\_freiburg1\_desk2 & 0.44543 & 0.28865 & 0.38157 \\
rgbd\_dataset\_freiburg1\_room & 0.14959 & 0.2838 & 0.17632 \\
rgbd\_dataset\_freiburg1\_xyz & 0.06059 & 0.0677 & 0.0635 \\
rgbd\_dataset\_freiburg3\_long\_office\_household & 0.03609 & 0.03543 & 0.04005 \\
rgbd\_dataset\_freiburg3\_nostructure\_texture\_far & 0.29727 & 0.4361 & 0.70826 \\
rgbd\_dataset\_freiburg3\_nostructure\_texture\_near\_withloop & 0.09011 & 0.13774 & 0.11602 \\
Average & 0.20073 & 0.20615 & 0.24465 \\
Std Dev & 0.18661 & 0.18182 & 0.26334 \\
Best (Average) Preset & baseline &  &  \\
Best (Average) Metric & 0.20073 &  &  \\

\bottomrule
\end{tabularx}
\end{minipage}

\bigskip

\begin{minipage}{\textwidth}
\noindent
\captionsetup{type=table}
\captionof{table}{TUM: Table Percent Lost }\label{tab:tum_table_percent_lost}

\fontsize{7pt}{8pt}\selectfont

\begin{tabularx}{\linewidth}{ >{\RaggedRight\arraybackslash}p{ 8.0cm } >{\RaggedRight\arraybackslash}p{ 1.0cm } >{\RaggedRight\arraybackslash}p{ 1.25cm } >{\RaggedRight\arraybackslash}p{ 1.5cm }  }\toprule
Dataset & baseline & root\_sift & superpoint \\
\midrule
rgbd\_dataset\_freiburg1\_desk & 1.364 & 1.502 & 1.674 \\
rgbd\_dataset\_freiburg1\_desk2 & 0.548 & 1.55 & 0.612 \\
rgbd\_dataset\_freiburg1\_room & 0.074 & 0.148 & 0.088 \\
rgbd\_dataset\_freiburg1\_xyz & 0.05 & 0.076 & 0.078 \\
rgbd\_dataset\_freiburg3\_long\_office\_household & 0.0 & 0.0 & 0.0 \\
rgbd\_dataset\_freiburg3\_nostructure\_texture\_far & 0.622 & 0.622 & 0.666 \\
rgbd\_dataset\_freiburg3\_nostructure\_texture\_near\_withloop & 0.024 & 0.036 & 0.036 \\
Average & 0.38314 & 0.562 & 0.45057 \\
Std Dev & 0.50438 & 0.82947 & 0.59847 \\
Best (Average) Preset & baseline &  &  \\
Best (Average) Metric & 0.38314 &  &  \\

\bottomrule
\end{tabularx}
\end{minipage}

\bigskip

\hypertarget{euroc-evaluation}{%
\subsection{EUROC Evaluation}\label{euroc-evaluation}}

\begin{minipage}{\textwidth}
\noindent
\captionsetup{type=table}
\captionof{table}{EUROC: Table RMSE}\label{tab:euroc_table_rmse}

\fontsize{7pt}{8pt}\selectfont

\begin{tabularx}{\linewidth}{ >{\RaggedRight\arraybackslash}p{ 4.5cm } >{\RaggedRight\arraybackslash}p{ 1.25cm } >{\RaggedRight\arraybackslash}p{ 1.25cm } >{\RaggedRight\arraybackslash}p{ 1.5cm }  }\toprule
Dataset & baseline & root\_sift & superpoint \\
\midrule
V101 & 0.08819 & 0.08816 & 0.08834 \\
V102 & 0.06852 & 0.06609 & 0.06694 \\
V201 & 0.07062 & 0.06409 & 0.09735 \\
V202 & 0.62539 & 0.56925 & 0.53736 \\
MH01 & 0.03775 & 0.03781 & 0.04172 \\
MH02 & 0.04217 & 0.0456 & 0.04177 \\
MH03 & 0.0512 & 0.04905 & 0.04882 \\
MH04 & 0.0702 & 0.05907 & 0.0646 \\
MH05 & 0.08177 & 0.06607 & 0.07131 \\
Average & 0.1262 & 0.11613 & 0.11758 \\
Std Dev & 0.19831 & 0.17517 & 0.15435 \\
Best (Average) Preset & root\_sift &  &  \\
Best (Average) Metric & 0.11613 &  &  \\

\bottomrule
\end{tabularx}
\end{minipage}

\bigskip

\begin{minipage}{\textwidth}
\noindent
\captionsetup{type=table}
\captionof{table}{EUROC: Table Max }\label{tab:euroc_table_max}

\fontsize{7pt}{8pt}\selectfont

\begin{tabularx}{\linewidth}{ >{\RaggedRight\arraybackslash}p{ 4.5cm } >{\RaggedRight\arraybackslash}p{ 1.5cm } >{\RaggedRight\arraybackslash}p{ 1.25cm } >{\RaggedRight\arraybackslash}p{ 1.5cm }  }\toprule
Dataset & baseline & root\_sift & superpoint \\
\midrule
V101 & 0.16335 & 0.163 & 0.16208 \\
V102 & 0.14545 & 0.12379 & 0.13136 \\
V201 & 0.15864 & 0.18921 & 0.22163 \\
V202 & 2.06607 & 2.52628 & 1.97847 \\
MH01 & 0.09141 & 0.10259 & 0.21303 \\
MH02 & 0.11922 & 0.1318 & 0.11196 \\
MH03 & 0.19995 & 0.17533 & 0.14082 \\
MH04 & 0.32021 & 0.27938 & 0.26722 \\
MH05 & 0.24828 & 0.17474 & 0.18991 \\
Average & 0.39029 & 0.42957 & 0.37961 \\
Std Dev & 0.67128 & 0.75543 & 0.6093 \\
Best (Average) Preset & superpoint &  &  \\
Best (Average) Metric & 0.37961 &  &  \\

\bottomrule
\end{tabularx}
\end{minipage}

\bigskip

\begin{minipage}{\textwidth}
\noindent
\captionsetup{type=table}
\captionof{table}{EUROC: Table Percent Lost }\label{tab:euroc_table_percent_lost}

\fontsize{7pt}{8pt}\selectfont

\begin{tabularx}{\linewidth}{ >{\RaggedRight\arraybackslash}p{ 4.5cm } >{\RaggedRight\arraybackslash}p{ 1.5cm } >{\RaggedRight\arraybackslash}p{ 1.25cm } >{\RaggedRight\arraybackslash}p{ 1.5cm }  }\toprule
Dataset & baseline & root\_sift & superpoint \\
\midrule
V101 & 0.0 & 0.0 & 0.0 \\
V102 & 0.0 & 0.0 & 0.0 \\
V201 & 0.04 & 0.04 & 0.04 \\
V202 & 0.146 & 0.138 & 0.112 \\
MH01 & 0.05 & 0.074 & 0.08 \\
MH02 & 0.018 & 0.026 & 0.012 \\
MH03 & 0.046 & 0.03 & 0.032 \\
MH04 & 0.0 & 0.01 & 0.0 \\
MH05 & 0.07 & 0.016 & 0.034 \\
Average & 0.04111 & 0.03711 & 0.03444 \\
Std Dev & 0.05351 & 0.04829 & 0.04435 \\
Best (Average) Preset & superpoint &  &  \\
Best (Average) Metric & 0.03444 &  &  \\

\bottomrule
\end{tabularx}
\end{minipage}

\bigskip

\hypertarget{kitti-evaluation}{%
\subsection{KITTI Evaluation}\label{kitti-evaluation}}

\begin{minipage}{\textwidth}
\noindent
\captionsetup{type=table}
\captionof{table}{KITTI: Table RMSE}\label{tab:kitt_table_rmse}

\fontsize{7pt}{8pt}\selectfont

\begin{tabularx}{\linewidth}{ >{\RaggedRight\arraybackslash}p{ 4.5cm } >{\RaggedRight\arraybackslash}p{ 1.25cm } >{\RaggedRight\arraybackslash}p{ 1.25cm } >{\RaggedRight\arraybackslash}p{ 1.5cm }  }\toprule
Dataset & baseline & root\_sift & superpoint \\
\midrule
00 & 3.947 & 3.24849 & 3.93432 \\
01 & 21.60014 & 21.86468 & 23.37292 \\
02 & 6.1194 & 5.6707 & 10.14455 \\
03 & 5.96451 & 5.92601 & 5.88895 \\
04 & 1.74556 & 1.76121 & 1.65506 \\
05 & 2.13915 & 2.11047 & 2.52263 \\
06 & 2.74011 & 2.73981 & 2.6784 \\
07 & 1.12358 & 1.12823 & 1.09496 \\
08 & 3.99078 & 4.13702 & 4.14543 \\
09 & 4.26068 & 4.35988 & 4.06244 \\
10 & 2.22097 & 2.25127 & 2.27694 \\
Average & 5.07744 & 5.01798 & 5.61605 \\
Std Dev & 5.47265 & 5.63609 & 6.77 \\
Best (Average) Preset & root\_sift &  &  \\
Best (Average) Metric & 5.01798 &  &  \\

\bottomrule
\end{tabularx}
\end{minipage}

\bigskip

\begin{minipage}{\textwidth}
\noindent
\captionsetup{type=table}
\captionof{table}{KITTI: Table Max }\label{tab:kitt_table_max}

\fontsize{7pt}{8pt}\selectfont

\begin{tabularx}{\linewidth}{ >{\RaggedRight\arraybackslash}p{ 4.5cm } >{\RaggedRight\arraybackslash}p{ 1.25cm } >{\RaggedRight\arraybackslash}p{ 1.25cm } >{\RaggedRight\arraybackslash}p{ 1.5cm }  }\toprule
Dataset & baseline & root\_sift & superpoint \\
\midrule
00 & 6.82983 & 5.8663 & 7.04067 \\
01 & 33.32903 & 33.85867 & 36.62253 \\
02 & 13.97623 & 9.47974 & 21.05188 \\
03 & 10.16642 & 10.16978 & 10.06014 \\
04 & 3.09135 & 3.0576 & 2.87271 \\
05 & 3.81484 & 3.77687 & 4.9547 \\
06 & 6.53758 & 5.14239 & 5.11673 \\
07 & 1.84418 & 1.82355 & 1.8655 \\
08 & 10.809 & 11.23635 & 10.75959 \\
09 & 8.53888 & 8.70237 & 8.31166 \\
10 & 4.13901 & 4.37849 & 4.27905 \\
Average & 9.37058 & 8.86292 & 10.26683 \\
Std Dev & 8.62548 & 8.62145 & 12.23529 \\
Best (Average) Preset & root\_sift &  &  \\
Best (Average) Metric & 8.86292 &  &  \\

\bottomrule
\end{tabularx}
\end{minipage}

\bigskip

\begin{minipage}{\textwidth}
\noindent
\captionsetup{type=table}
\captionof{table}{KITTI: Table Percent Lost }\label{tab:kitt_table_percent_lost}

\fontsize{7pt}{8pt}\selectfont

\begin{tabularx}{\linewidth}{ >{\RaggedRight\arraybackslash}p{ 4.5cm } >{\RaggedRight\arraybackslash}p{ 1.5cm } >{\RaggedRight\arraybackslash}p{ 1.25cm } >{\RaggedRight\arraybackslash}p{ 1.5cm }  }\toprule
Dataset & baseline & root\_sift & superpoint \\
\midrule
00 & 0.0 & 0.038 & 0.012 \\
01 & 0.0 & 0.0 & 0.0 \\
02 & 0.012 & 0.004 & 0.004 \\
03 & 0.0 & 0.0 & 0.0 \\
04 & 0.0 & 0.0 & 0.0 \\
05 & 0.0 & 0.0 & 0.028 \\
06 & 0.108 & 0.018 & 0.0 \\
07 & 0.0 & 0.0 & 0.0 \\
08 & 0.0 & 0.0 & 0.0 \\
09 & 0.0 & 0.0 & 0.0 \\
10 & 0.0 & 0.0 & 0.0 \\
Average & 0.01091 & 0.00545 & 0.004 \\
Std Dev & 0.07244 & 0.02373 & 0.0196 \\
Best (Average) Preset & superpoint &  &  \\
Best (Average) Metric & 0.004 &  &  \\

\bottomrule
\end{tabularx}
\end{minipage}

\bigskip

\hypertarget{testing}{%
\subsection{Automated Testing}\label{testing}}

Reproducibility also depends on detecting regressions as the framework evolves. The repository therefore separates the automated test suite (folder \texttt{tests}) from the usage examples (folder \texttt{examples}). The suite comprises Python unit tests run with \texttt{pytest}, C++ unit tests of the sparse core, Python tests of the C++ core and its bindings, and headless checks of the C++ extension modules. A single launcher, \texttt{tests/run\_all.sh}, runs all of them in sequence and reports the failing steps. Several tests reproduce past defects, such as crashes in local mapping and relocalization or inconsistent loop-detector database indices, so that they cannot reappear unnoticed. The direct-SLAM pipeline has its own synthetic C++ tests: one executable covers photometric and geometric motion, stereo and RGB-D scale, native-rig BA, outlier rejection, fixed gauges, bounded windows, failure and recovery, the depth-filter update against a transcription of SVO's seed update, and monocular two-view initialization, while a second one unit-tests the shared building blocks (image samplers and pyramids, occupancy masks, live-point measurement, map bookkeeping, and the tracker hooks). Python tests check the binding, the monocular semi-direct path, and the loop-closing bridge.

Changes that affect the camera models are additionally checked end to end. A replay harness processes a dataset sequence deterministically (ORB2 features, synchronous local mapping, and no loop closure unless requested) and records tracking states, per-frame timings, keyframe and map-point counts, and the absolute trajectory error, with Sim(3) alignment for monocular runs and SE(3) alignment for metric sensors. A comparison script replays the same sequences with a baseline checkout and with the current one, alternating their order, and fails if the median paired frame-time ratio exceeds $1.20$, if any pair loses more than two percentage points of tracking or ground-truth coverage, or if the trajectory error grows by more than the larger of $20\%$ and $0.02$\,m.

The visual models (Sect.~\ref{visual-model-catalog}) have their own checks. Inventory and benchmark scripts capture source and patch provenance, enum and \texttt{FeatureInfo} compatibility, median/p95 pair timings, and numerical output fixtures, and a comparison script acts as a regression gate on time, memory, and match counts. Missing measurements remain explicit: synthetic pair inliers do not replace sequence ATE, and a CPU baseline does not certify GPU performance. Map benchmarks isolate map operations from feature extraction; cross-backend speed ratios are advisory by default, with optional same-backend baseline regression checks. New methods and changes of runtime defaults require output equivalence, a license review, and completed performance gates.

\hypertarget{camera-model-validation}{%
\subsection{Camera Models on TUM-VI}\label{camera-model-validation}}

As an initial check of the camera models of Sect.~\ref{camera-models}, the first 300 frames of the TUM-VI \texttt{room1} sequence (512$\times$512) were replayed in the rectified and native stereo modes, using the C++ core, g2o, ORB2 with 1200 features, and synchronous local mapping. Table~\ref{tab:tumvi_modes} summarizes the results. Both modes tracked all frames after initialization. The rectified mode, which runs the established pinhole pipeline, was faster and more accurate on this short segment; the native mode pays for bearing-based stereo matching and model-aware projection and optimization.

\begin{table}[!ht]
\centering
\caption{TUM-VI \texttt{room1}, first 300 frames: rectified versus native fisheye stereo. Timings include tracking and synchronous local mapping.}
\label{tab:tumvi_modes}
\fontsize{8pt}{9.6pt}\selectfont
\begin{tabular}{lrr}
\toprule
Measurement & Rectified & Native \\
\midrule
Tracking states & 1 initializing, 299 OK & 1 initializing, 299 OK \\
Trajectory / mocap associations & 299 / 299 & 299 / 299 \\
SE(3)-aligned translation RMSE & 0.0061\,m & 0.0127\,m \\
Mean tracking + local mapping & 81.5\,ms/frame & 238.1\,ms/frame \\
Median tracking + local mapping & 38.4\,ms/frame & 188.0\,ms/frame \\
Mean image loading and preprocessing & 14.5\,ms/pair & 9.7\,ms/pair \\
Final keyframes / map points & 48 / 2141 & 54 / 2417 \\
\bottomrule
\end{tabular}
\end{table}

These are exploratory measurements on a short segment, not a controlled benchmark: timings include warm-up and logging, and full-sequence drift, loop closing, the remaining TUM-VI sequences, and the Python and GTSAM paths on real data remain to be evaluated.

\hypertarget{conclusions}{%
\section{Conclusions and Future Work}\label{conclusions}}

This document presented pySLAM, an open-source framework for visual SLAM that combines a sparse SLAM core, available in both Python and C++, and a C++ direct and semi-direct SLAM pipeline with interchangeable local features, loop detectors, depth estimators, semantic segmentation models, volumetric integrators, and feed-forward multi-view reconstruction models. A unified dataset interface, automated evaluation scripts, and a regression test suite make it possible to compare these components within a complete pipeline and across more than fifteen dataset types. The framework is intended as a research and teaching platform rather than a real-time system.

Several directions remain open. Inertial and LiDAR measurements are not yet integrated. Object-level mapping is under development. For the camera models, wide fields of view beyond $180^\circ$ still need to be validated in initialization and Sim(3) estimation, and the native fisheye pipeline needs a full-sequence evaluation on TUM-VI and other fisheye datasets.

%




\hypertarget{credits}{%
\section*{Acknowledgments}\label{credits}}
\addcontentsline{toc}{section}{Acknowledgments}

A list of the frameworks that inspired or has been integrated into pySLAM are available \href{https://github.com/luigifreda/pyslam?tab=readme-ov-file#credits}{here}. Many thanks to their Authors for their great work.

\bibliographystyle{plain}
\bibliography{bibliography}

\clearpage
\appendix

\hypertarget{usage}{%
\section{Usage}\label{usage}}

Open a new terminal and start experimenting with the scripts. In each
new terminal you are supposed to start with this command:

\begin{scriptsize}
\begin{Shaded}
\begin{Highlighting}[]
\CommentTok{#pixi shell           # If you use pixi, this activates the pyslam environment. }
\NormalTok{ }\BuiltInTok{.} \ExtensionTok{pyenv-activate.sh}   \CommentTok{#  Activate pyslam python virtual environment. This is only needed once in a new terminal.}
\end{Highlighting}
\end{Shaded}
\end{scriptsize}

If you are using \texttt{pixi} then just run \texttt{pixi\ shell} to
activate the \texttt{pyslam} environment. The file \href{https://github.com/luigifreda/pyslam/blob/master/config.yaml}{config.yaml} can be used as a unique
entry-point to configure the system and its global configuration
parameters contained in
\href{https://github.com/luigifreda/pyslam/blob/master/pyslam/config_parameters.py}{pyslam/config\_parameters.py}. 
\begin{itemize}
\item The \textbf{C++ sparse-SLAM core} is enabled by default: the flag \texttt{USE\_CPP\_CORE} in the file \texttt{pyslam/config\_parameters.py} is \texttt{True}. To use the Python core, set the flag to \texttt{False} or launch with \texttt{PYSLAM\_USE\_CPP=false} (see Sect.~\ref{cpp-core}).
\item Further information on how to configure pySLAM is provided \protect\hyperlink{selecting-a-dataset-and-different-configuration-parameters}{here}.
\end{itemize}

\hypertarget{visual-odometry}{%
\subsection{Visual odometry}\label{visual-odometry}}

The basic \textbf{Visual Odometry} (VO) can be run with the following
commands:

\begin{scriptsize}
\begin{Shaded}
\begin{Highlighting}[]
\CommentTok{#pixi shell           # If you use pixi, this activates the pyslam environment. }
\NormalTok{$ }\BuiltInTok{.} \ExtensionTok{pyenv-activate.sh}   \CommentTok{#  Activate pyslam python virtual environment. This is only needed once in a new terminal.}
\NormalTok{$ }\ExtensionTok{./main_vo.py}
\end{Highlighting}
\end{Shaded}
\end{scriptsize}

By default, the script processes a
\href{http://www.cvlibs.net/datasets/kitti/eval_odometry.php}{KITTI}
video (available in the folder \texttt{data/videos}) by using its
corresponding camera calibration file (available in the folder
\texttt{settings}), and its groundtruth (available in the same
\texttt{data/videos} folder). If matplotlib windows are used, you can
stop \texttt{main\_vo.py} by clicking on one of them and
pressing the key `Q'. As explained above, this very \emph{basic} script
\texttt{main\_vo.py} \textbf{strictly requires a ground truth}. Now,
with RGBD datasets, you can also test the \textbf{RGBD odometry} with
the classes \texttt{VisualOdometryRgbd} or
\texttt{VisualOdometryRgbdTensor} (ground truth is not required here).

\hypertarget{full-slam}{%
\subsection{Full SLAM}\label{full-slam}}

Similarly, you can test the \textbf{full SLAM} by running
\texttt{main\_slam.py}:\\

\begin{scriptsize}
\begin{Shaded}
\begin{Highlighting}[]
\CommentTok{#pixi shell           # If you use pixi, this activates the pyslam environment. }
\NormalTok{$ }\BuiltInTok{.} \ExtensionTok{pyenv-activate.sh}   \CommentTok{#  Activate pyslam python virtual environment. This is only needed once in a new terminal.}
\NormalTok{$ }\ExtensionTok{./main_slam.py}
\end{Highlighting}
\end{Shaded}
\end{scriptsize}

This will process the same default
\href{http://www.cvlibs.net/datasets/kitti/eval_odometry.php}{KITTI}
video (available in the folder \texttt{data/videos}) by using its
corresponding camera calibration file (available in the folder
\texttt{settings}). You can stop it by clicking on one of the
opened windows and pressing the key `Q' or closing the 3D pangolin GUI.

\hypertarget{direct-slam-usage}{%
\subsection{Direct SLAM}\label{direct-slam-usage}}

To run the direct-SLAM pipeline of Sect.~\ref{direct-slam}, select a stereo or RGB-D dataset in \texttt{config.yaml} (or a monocular one with \texttt{semi\_direct}), build \texttt{cpp\_core} with \texttt{./build\_cpp\_core.sh}, and run \texttt{main\_slam\_direct.py}:\\

\begin{scriptsize}
\begin{Shaded}
\begin{Highlighting}[]
\NormalTok{$ }\ExtensionTok{./main_slam_direct.py} \AttributeTok{-c}\NormalTok{ config.yaml}
\NormalTok{$ }\ExtensionTok{./main_slam_direct.py} \AttributeTok{-c}\NormalTok{ config.yaml }\AttributeTok{--headless --max-frames}\NormalTok{ 100}
\NormalTok{$ }\ExtensionTok{./main_slam_direct.py} \AttributeTok{-c}\NormalTok{ config.yaml }\AttributeTok{--mode}\NormalTok{ semi_direct}
\end{Highlighting}
\end{Shaded}
\end{scriptsize}

\texttt{--mode} overrides \texttt{DIRECT\_SLAM.mode} and \texttt{DIRECT\_SLAM.monocular.mode} (\texttt{klt}, \texttt{hybrid}, or \texttt{semi\_direct}). For a monocular camera, the keys of the \texttt{DIRECT\_SLAM.monocular} block override the corresponding \texttt{DIRECT\_SLAM} keys, so that one settings file serves both stereo and monocular runs; the KITTI settings, for instance, select \texttt{hybrid} for stereo and \texttt{semi\_direct} with a larger corner and seed budget for monocular input. \texttt{--trajectory-file} exports the final accepted trajectory, after all local and loop corrections, as KITTI $T_{wc}$ rows. \texttt{--no-local-ba} and \texttt{--no-loop-closing} disable the automatic local bundle adjustment and the loop-closing thread. The full parameter reference is in \texttt{docs/direct\_slam.md}.
\hypertarget{selecting-a-dataset-and-different-configuration-parameters}{%
\subsection{Selecting a dataset and different configuration
parameters}\label{selecting-a-dataset-and-different-configuration-parameters}}

The file \href{https://github.com/luigifreda/pyslam/blob/master/config.yaml}{config.yaml} can be used as a unique
entry-point to configure the system, the target dataset and its global
configuration parameters set in
\href{https://github.com/luigifreda/pyslam/blob/master/pyslam/config_parameters.py}{pyslam/config\_parameters.py}.
To process a different \textbf{dataset} with both VO and SLAM scripts,
you need to update the file \href{https://github.com/luigifreda/pyslam/blob/master/config.yaml}{config.yaml}:
\begin{itemize}
\item Select
your dataset \texttt{type} in the section \texttt{DATASET} (further
details in the section \emph{\protect\hyperlink{datasets}{Datasets}}
below for further details). This identifies a corresponding dataset
section (e.g.~\texttt{KITTI\_DATASET}, \texttt{TUM\_DATASET}, etc). 
\item  Select the \texttt{sensor\_type} (\texttt{mono}, \texttt{stereo},
\texttt{rgbd}) in the chosen dataset section.
\item  Select the camera \texttt{settings} file in the dataset section
(further details in the section
\emph{\protect\hyperlink{camera-settings}{Camera Settings}} below).
\item Set the \texttt{groundtruth\_file} accordingly. Further details in the
section \emph{\protect\hyperlink{datasets}{Datasets}} below (see also
the files \texttt{pyslam/io/ground\_truth.py} and
\texttt{pyslam/io/convert\_groundtruth\_to\_simple.py}).
\end{itemize}

You can use the section \texttt{GLOBAL\_PARAMETERS} of the file
\href{https://github.com/luigifreda/pyslam/blob/master/config.yaml}{config.yaml} to override the global configuration
parameters set in \href{https://github.com/luigifreda/pyslam/blob/master/pyslam/config_parameters.py}{pyslam/config\_parameters.py}.
This is particularly useful when running a
\protect\hyperlink{evaluating-slam}{SLAM evaluation}.

\hypertarget{feature-tracking}{%
\subsection{Feature tracking}\label{feature-tracking}}

If you just want to test the basic feature tracking capabilities
(\emph{feature detector} + \emph{feature descriptor} + \emph{feature
matcher}) and get a taste of the different available local features, run \\

\begin{scriptsize}
\begin{Shaded}
\begin{Highlighting}[]
\CommentTok{#pixi shell           # If you use pixi, this activates the pyslam environment. }
\NormalTok{$ }\BuiltInTok{.} \ExtensionTok{pyenv-activate.sh}   \CommentTok{#  Activate pyslam python virtual environment. This is only needed once in a new terminal.}
\NormalTok{$}\ExtensionTok{./main_feature_matching.py}
\end{Highlighting}
\end{Shaded}
\end{scriptsize}

In any of the above scripts, you can choose any detector/descriptor
among \emph{ORB}, \emph{SIFT}, \emph{SURF}, \emph{BRISK}, \emph{AKAZE},
\emph{SuperPoint}, etc. (see Sect.~\ref{supported-local-features} for the
complete list).

In both the scripts \texttt{main\_vo.py} and \texttt{main\_slam.py}, you
can create your preferred detector-descriptor configuration and feed it
to the function \texttt{feature\_tracker\_factory()}. Some ready-to-use
configurations are already available in the file
\href{https://github.com/luigifreda/pyslam/blob/master/pyslam/local_features/feature_tracker_configs.py}{pyslam/local\_features/feature\_tracker\_configs.py}.

The function \texttt{feature\_tracker\_factory()} can be found in the
file \texttt{pyslam/local\_features/feature\_tracker.py}. Take a look at the
file \texttt{pyslam/local\_features/feature\_manager.py} for further details.

Some basic examples are available in the subfolder
\texttt{examples/cv}. In particular, as for feature
detection/description, you may want to take a look at
\href{https://github.com/luigifreda/pyslam/blob/master/examples/cv/example_feature_manager.py}{examples/cv/example\_feature\_manager.py}
too.

\hypertarget{loop-closing}{%
\subsection{Loop closing}\label{loop-closing}}

Many \protect\hyperlink{loop-closing}{loop closing methods} are
available, combining different
\protect\hyperlink{local-descriptor-aggregation-methods}{aggregation
methods} and \protect\hyperlink{global-descriptors}{global descriptors}.

While running full SLAM, loop closing is enabled by default and can be
disabled by setting \texttt{kUseLoopClosing=False} in
\texttt{pyslam/config\_parameters.py}. Different configuration options
\texttt{LoopDetectorConfigs} can be found in
\href{https://github.com/luigifreda/pyslam/blob/master/pyslam/loop_closing/loop_detector_configs.py}{pyslam/loop\_closing/loop\_detector\_configs.py}: Code comments provide additional useful details.

One can start experimenting with loop closing methods by using the
examples in \texttt{examples/loopclosing}. The example
\href{https://github.com/luigifreda/pyslam/blob/master/examples/loopclosing/example_loop_detector.py}{examples/loopclosing/example\_loop\_detector.py}
is the recommended entry point.

\hypertarget{vocabulary-management}{%
\paragraph{Vocabulary management}\label{vocabulary-management}}

\texttt{DBoW2}, \texttt{DBoW3}, and \texttt{VLAD} require
\textbf{pre-trained vocabularies}. ORB-based vocabularies are
automatically downloaded into the \texttt{data} folder (see
\href{https://github.com/luigifreda/pyslam/blob/master/pyslam/loop_closing/loop_detector_configs.py}{pyslam/loop\_closing/loop\_detector\_configs.py}).

To create a new vocabulary, follow these steps:

\begin{enumerate}
\def\labelenumi{\arabic{enumi}.}
\item
  \textbf{Generate an array of descriptors}: Use the script \\
  \texttt{examples/loopclosing/example\_gen\_des\_array\_from\_imgs.py} to
  generate the array of descriptors that will be used to train the new
  vocabulary. Select your desired descriptor type via the tracker
  configuration.
\item
  \textbf{DBOW vocabulary generation}: Train your target DBOW vocabulary
  by using the script
  \texttt{examples/loopclosing/example\_gen\_dbow\_voc\_from\_des\_array.py}.
\item
  \textbf{VLAD vocabulary generation}: Train your target VLAD
  ``vocabulary'' by using the script
  \texttt{examples/loopclosing/example\_gen\_vlad\_voc\_from\_des\_array.py}.
\end{enumerate}

Once you have trained the vocabulary, you can add it in
\href{https://github.com/luigifreda/pyslam/blob/master/pyslam/loop_closing/loop_detector_vocabulary.py}{pyslam/loop\_closing/loop\_detector\_vocabulary.py} and correspondingly
create a new loop detector configuration in
\href{https://github.com/luigifreda/pyslam/blob/master/pyslam/loop_closing/loop_detector_configs.py}{pyslam/loop\_closing/loop\_detector\_configs.py} that uses it.

\hypertarget{vocabulary-free-loop-closing}{%
\paragraph{Vocabulary-free loop
closing}\label{vocabulary-free-loop-closing}}

Most methods do not require pre-trained vocabularies. Specifically:
\begin{itemize}
\item \texttt{iBoW} and \texttt{OBindex2}: These methods incrementally build
bags of binary words and, if needed, convert (front-end) non-binary
descriptors into binary ones. 
\item Others: Methods like \texttt{HDC\_DELF},
\texttt{SAD}, \texttt{AlexNet}, \texttt{NetVLAD}, \texttt{CosPlace}, and
\texttt{EigenPlaces}, \texttt{Megaloc}, \texttt{SALAD}, and \texttt{BoQ} directly extract their specific \textbf{global
descriptors} and process them using dedicated aggregators, independently
from the used front-end descriptors.
\end{itemize}

As mentioned above, only \texttt{DBoW2}, \texttt{DBoW3}, and
\texttt{VLAD} require pre-trained vocabularies.

\hypertarget{double-check-your-loop-detection-configuration-and-verify-vocabulary-compability}{%
\paragraph{Verify your loop detection configuration and verify
vocabulary
compability}\label{double-check-your-loop-detection-configuration-and-verify-vocabulary-compability}}

\hypertarget{loop-detection-method-based-on-a-pre-trained-vocabulary}{%
\subparagraph{Loop detection method based on a pre-trained
vocabulary}\label{loop-detection-method-based-on-a-pre-trained-vocabulary}}

When selecting a \textbf{loop detection method based on a pre-trained
vocabulary} (such as \texttt{DBoW2}, \texttt{DBoW3}, and \texttt{VLAD}),
ensure the following: 
\begin{enumerate}
\item The back-end and the front-end are using the
same descriptor type (this is also automatically checked for
consistency) or their descriptor managers are independent (see further
details in the configuration options \texttt{LoopDetectorConfigs}
available in \href{https://github.com/luigifreda/pyslam/blob/master/pyslam/loop_closing/loop_detector_configs.py}{pyslam/loop\_closing/loop\_detector\_configs.py}). 
\item A corresponding pre-trained vocubulary is available. For more details,
refer to the \protect\hyperlink{vocabulary-management}{vocabulary
management section}.
\end{enumerate}

\hypertarget{missing-vocabulary-for-the-selected-front-end-descriptor-type}{%
\subparagraph{Missing vocabulary for the selected front-end descriptor
type}\label{missing-vocabulary-for-the-selected-front-end-descriptor-type}}

If you lack a compatible vocabulary for the selected front-end
descriptor type, you can follow one of these options:\\
1. Create and load the vocabulary (refer to the
\protect\hyperlink{vocabulary-management}{vocabulary management
section}).\\
2. Choose an \texttt{*\_INDEPENDENT} loop detector method, which works
with an independent local\_feature\_manager.\\
3. Select a vocabulary-free loop closing method.

See the file \href{https://github.com/luigifreda/pyslam/blob/master/pyslam/loop_closing/loop_detector_configs.py}{pyslam/loop\_closing/loop\_detector\_configs.py} for further
details.

\hypertarget{volumetric-reconstruction}{%
\subsection{Volumetric reconstruction}\label{volumetric-reconstruction}}

\hypertarget{dense-reconstruction-while-running-slam}{%
\paragraph{Dense reconstruction while running
SLAM}\label{dense-reconstruction-while-running-slam}}

The SLAM back-end hosts a volumetric reconstruction pipeline. This is
disabled by default. You can enable it by setting
\texttt{kDoVolumetricIntegration=True} and selecting your preferred
method \texttt{kVolumetricIntegrationType} in
\texttt{pyslam/config\_parameters.py}. At present, the following methods
are available: 
\begin{itemize}
\item \texttt{VOXEL\_GRID}, \texttt{VOXEL\_SEMANTIC\_GRID},
\texttt{VOXEL\_SEMANTIC\_PROBABILISTIC\_GRID}: Voxel grid
implementations that leverage parallel spatial hashing, supporting both
direct voxel hashing and indirect voxel-block hashing strategies.
Parallel execution is managed using TBB, and the design accommodates
both simple and semantic voxels. 
\item \texttt{TSDF} (Truncated Signed
Distance Function): It is able to return in output either a pointcloud
or a mesh. 
\item  \texttt{GAUSSIAN\_SPLATTING} (Incremental Gaussian
splatting)\href{https://github.com/luigifreda/pyslam/blob/master/pyslam/dense/volumetric_integrator_factory.py}{pyslam/dense/volumetric\_integrator\_factory.py}). 
\end{itemize}

Note that you need CUDA in order to run \texttt{GAUSSIAN\_SPLATTING} method.

The volumetric reconstruction pipeline works with:
\begin{itemize}
\item RGBD datasets 
\item When a \protect\hyperlink{depth-prediction}{depth estimator}
is used
\begin{itemize}
\item[$\circ$] in the back-end with STEREO datasets (you can't use depth
prediction in the back-end with MONOCULAR datasets, further details
\protect\hyperlink{depth-prediction}{here}) 
\item[$\circ$] in the front-end (to emulate an RGBD sensor) and a depth prediction/estimation gets available
for each processed keyframe.
\end{itemize}
\end{itemize}

To obtain a mesh as output, set
\texttt{kVolumetricIntegrationTsdfExtractMesh=True} in
\texttt{config\_parameters.py}.

\hypertarget{reload-a-saved-sparse-map-and-perform-dense-reconstruction}{%
\paragraph{Reload a saved sparse map and perform dense
reconstruction}\label{reload-a-saved-sparse-map-and-perform-dense-reconstruction}}

Use the script \texttt{main\_map\_dense\_reconstruction.py} to reload a
saved sparse map and perform dense reconstruction by using its posed
keyframes as input. You can select your preferred dense reconstruction
method directly in the script.

\begin{itemize}
\tightlist
\item
  To check what the volumetric integrator is doing, run in another shell
  \texttt{tail\ -f\ logs/volumetric\_integrator.log} (from repository
  root folder).
\item
  To save the obtained dense and sparse maps, press the \texttt{Save}
  button on the GUI.
\end{itemize}

\hypertarget{reload-and-check-your-dense-reconstruction}{%
\paragraph{Reload and check your dense
reconstruction}\label{reload-and-check-your-dense-reconstruction}}

You can check the output pointcloud/mesh by using
\href{https://www.cloudcompare.org/}{CloudCompare}.

In the case of a saved Gaussian splatting model, you can visualize it
by: 
\begin{enumerate}
\item Using the
\href{https://playcanvas.com/supersplat/editor}{superslat editor} (drag
and drop the saved Gaussian splatting \texttt{.ply} pointcloud in the
editor interface). 

\item Getting into the folder
\texttt{examples/gaussian\_splatting} and running:\\
\begin{codeblock}
\codeline{python\ example\_gsm.py\ -\/-load\ \textless{}gs\_checkpoint\_path\textgreater{}}
\end{codeblock}

The directory \texttt{<gs\_checkpoint\_path>} is expected to have the following structure:\\
\begin{scriptsize}
\begin{Shaded}
\begin{Highlighting}[]
\NormalTok{+-- gs_checkpoint_path}
\NormalTok{|   +-- pointcloud         \textit{# folder containing different subfolders, each one with a saved .ply}}
\NormalTok{|                          \textit{# encoding the Gaussian splatting model at a specific iteration/checkpoint}}
\NormalTok{|   +-- last_camera.json}
\NormalTok{|   +-- config.yml}
\end{Highlighting}
\end{Shaded}
\end{scriptsize}

\end{enumerate}

\hypertarget{controlling-the-spatial-distribution-of-keyframe-fov-centers}{%
\paragraph{Controlling the spatial distribution of keyframe FOV
centers}\label{controlling-the-spatial-distribution-of-keyframe-fov-centers}}

If you are targeting volumetric reconstruction while running SLAM, you
can enable a \textbf{keyframe generation policy} designed to manage the
spatial distribution of keyframe field-of-view (FOV) centers. The
\emph{FOV center of a camera} is defined as the backprojection of its
image center, calculated using the median depth of the frame. With this
policy, a new keyframe is generated only if its FOV center lies beyond a
predefined distance from the nearest existing keyframe's FOV center. You
can enable this policy by setting the following parameters in the yaml
setting:\\

\begin{scriptsize}
\begin{Shaded}
\begin{Highlighting}[]
\FunctionTok{KeyFrame.useFovCentersBasedGeneration:}\AttributeTok{ }\DecValTok{1}\AttributeTok{  }\CommentTok{# compute 3D fov centers of camera frames by using median depth }
\FunctionTok{}\AttributeTok{ }\DecValTok{}\AttributeTok{  }\CommentTok{                                       # use their distances to control keyframe generation}
\FunctionTok{KeyFrame.maxFovCentersDistance:}\AttributeTok{ }\FloatTok{0.2}\AttributeTok{       }\CommentTok{# max distance between fov centers in order to generate a keyframe}
\end{Highlighting}
\end{Shaded}
\end{scriptsize}

or by setting the following parameters in
\texttt{config\_parameters.py}:

\begin{scriptsize}
\begin{Shaded}
\begin{Highlighting}[]
\NormalTok{kUseFovCentersBasedKfGeneration }\OperatorTok{=} \VariableTok{False}  \CommentTok{# Use FOV centers based keyframe generation;}
\FunctionTok{}\AttributeTok{ }\DecValTok{}\AttributeTok{  }\CommentTok{                                       # not considered if KeyFrame.useFovCentersBasedGeneration is set in yaml}
\NormalTok{kMaxFovCentersDistanceForKfGeneration }\OperatorTok{=} \FloatTok{0.2}  \CommentTok{# [m] Maximum distance between FOV centers for keyframe generation;}
\FunctionTok{}\AttributeTok{ }\DecValTok{}\AttributeTok{  }\CommentTok{                                       # not considered if not considered if KeyFrame.maxFovCentersDistance is set in yaml}
\end{Highlighting}
\end{Shaded}
\end{scriptsize}


\hypertarget{depth-prediction}{%
\subsubsection{Depth prediction}\label{depth-prediction}}

The available depth prediction models can be utilized both in the SLAM
back-end and front-end. 
\begin{itemize}
\item \textbf{Back-end}: Depth prediction can be
enabled in the \protect\hyperlink{volumetric-reconstruction}{volumetric
reconstruction} pipeline by setting the parameter
\texttt{kVolumetricIntegrationUseDepthEstimator=True} and selecting your
preferred \texttt{kVolumetricIntegrationDepthEstimatorType} in
\texttt{pyslam/config\_parameters.py}. 
\item \textbf{Front-end}: Depth prediction
can be enabled in the front-end by setting the parameter
\texttt{kUseDepthEstimator\allowbreak InFrontEnd} in \texttt{pyslam/config\_parameters.py}.
This feature estimates depth images from input color images to emulate a
RGBD camera. Please, note this functionality is still
\emph{experimental} at present time.
\end{itemize}

\textbf{Notes}: 
\begin{itemize}
\item In the case of a \textbf{monocular SLAM}, do NOT use
depth prediction in the back-end volumetric integration: The SLAM (fake)
scale will conflict with the absolute metric scale of depth predictions.
With monocular datasets, you can enable depth prediction to run in the
front-end (to emulate an RGBD sensor). 
\item Depth inference may be
very slow (for instance, with DepthPro it takes \textasciitilde{}1s per
image on a typical machine). Therefore, the resulting volumetric reconstruction
pipeline may be very slow. Depth Pro 2 is much faster (see below).
\end{itemize}

Refer to the file
\texttt{pyslam/depth\_estimation/depth\_estimator\_factory.py} for further
details. Both stereo and monocular prediction approaches are supported.
You can test depth prediction/estimation by using the script
\texttt{main\_depth\_prediction.py}. The factory imports the wrapper of the selected estimator (and its thirdparty package) only when it is constructed, so a missing optional integration raises a \texttt{DepthEstimatorUnavailable} error only when it is selected.

\paragraph{Depth Pro 2.} \texttt{DEPTH\_PRO\_2} (Depth Pro 2\cite{depthpro2_2026}) predicts metric depth at the full input resolution. Its weights are downloaded on first use and are licensed for non-commercial research only. It is configured in \texttt{pyslam/config\_parameters.py}:
\begin{itemize}
\item \texttt{kDepthPro2ModelName}: \texttt{depthpro2-s} (24M parameters), \texttt{depthpro2-b} (97M, default), \texttt{depthpro2-l} (334M) or \texttt{depthpro2-h} (872M). The S model is fast but noticeably less accurate on single images.
\item \texttt{kDepthPro2Streaming}: when enabled, each frame also attends the previous \texttt{kDepthPro2StreamingWindow}$-1$ frames, which gives temporally consistent depth. Enable it only when the estimator receives every frame in order (front-end, \texttt{main\_depth\_prediction.py}), not in the back-end volumetric integration, which only sees keyframes.
\end{itemize}
When a camera is given, its $f_x$ (scaled to the image width) is used as the focal length. Otherwise the model estimates it per image, and the metric scale varies more from frame to frame. Sky and other infinitely far pixels get depth 0 (invalid). On a TUM fr3 sequence, \texttt{depthpro2-b} takes \textasciitilde{}40\,ms per $640\times480$ frame on a laptop RTX A3000 GPU (\textasciitilde{}65\,ms with streaming; DepthPro: \textasciitilde{}1\,s) and is slightly more accurate than DepthPro. Streaming reduces its frame-to-frame scale flicker about 3.5 times. An example is in \texttt{examples/depth\_estimation/example\_depth\_pro\_2.py}.


\hypertarget{semantic-mapping-and-image-segmentation}{%
\subsubsection{Semantic mapping and Image
Segmentation}\label{semantic-mapping-and-image-segmentation}}

The sparse semantic mapping pipeline can be enabled by setting the
parameter \texttt{kDoSparseSemanticMappingAndSegmentation=True} in
\texttt{pyslam/config\_parameters.py}. You can configure the semantic
mapping module by mofifying the default segmentation models assigned to
each dataset in \texttt{pyslam/semantics/semantic\_mapping\_configs.py}.
Otherwise, you can override the used segmentation model by setting
\texttt{kSemanticSegmentationType} in \texttt{config\_parameters.py}.

\hypertarget{sparse-semantic-mapping}{%
\paragraph{Sparse Semantic Mapping}\label{sparse-semantic-mapping}}

Different semantic mapping methods are available (see
\href{https://github.com/luigifreda/pyslam/blob/master/docs/semantics.md}{here} for further details). Currently, we
support semantic mapping using \textbf{dense semantic segmentation}. 
\begin{itemize}
\item \texttt{DEEPLABV3}: from \texttt{torchvision}, pre-trained on COCO/VOC.
\item \texttt{SEGFORMER}: from \texttt{transformers}, pre-trained on Cityscapes or ADE20k.
\item \texttt{CLIP}: from \texttt{f3rm} package for open-vocabulary support.
\item \texttt{EOV\_SEG}: open-vocabulary panoptic segmentation model.
\item \texttt{DETIC}: open-vocabulary object detection and segmentation model.
\item \texttt{ODISE}: open-vocabulary diffusion-based panoptic segmentation model.
\item \texttt{RFDETR}: real-time transformer architecture for object detection and instance segmentation (built on a DINOv2 vision transformer backbone). 
\item \texttt{YOLO}: built on years of foundational research in computer vision and AI. 
\end{itemize}

\textbf{Panoptic/instance segmentation notes}:
\begin{itemize}
\item \texttt{DETIC}: object detection-based. Each detection yields a distinct instance ID.
\item \texttt{ODISE}: panoptic segmentation model. Instances are derived from panoptic segments via the \texttt{isthing} flag and can merge multiple objects of the same category.
\item \texttt{EOV\_SEG}: similar to \texttt{ODISE}, instances are derived from panoptic segments and may group nearby objects of the same category.
\end{itemize}

\textbf{Semantic features} are assigned to \textbf{keypoints} on the
image and fused into map points. The semantic features can be:
\begin{itemize}
\item \textit{Labels}: categorical labels as numbers.
\item \textit{Probability vectors}: probability vectors for each class.
\item \textit{Feature vectors}: feature vectors
obtained from an encoder. This is generally used for open vocabulary
mapping.
\end{itemize}

\hypertarget{volumetric-semantic-mapping}{%
\paragraph{Volumetric Semantic
mapping}\label{volumetric-semantic-mapping}}

Volumetric semantic mapping can be enabled by setting:

\begin{scriptsize}
\begin{Shaded}
\begin{Highlighting}[]
\NormalTok{kDoSparseSemanticMappingAndSegmentation}\OperatorTok{=}\VariableTok{True} \CommentTok{#  enable sparse mapping and segmentation}
\NormalTok{kDoVolumetricIntegration }\OperatorTok{=} \VariableTok{True} \CommentTok{# enable volumetric integration}
\NormalTok{kVolumetricIntegrationType }\OperatorTok{=} \StringTok{"VOXEL_SEMANTIC_PROBABILISTIC_GRID"} \CommentTok{# use semantic volumetric models }
\FunctionTok{}\AttributeTok{ }\DecValTok{}\AttributeTok{  }\CommentTok{                                                      # like VOXEL_SEMANTIC_PROBABILISTIC_GRID and VOXEL_SEMANTIC_GRID}
\end{Highlighting}
\end{Shaded}
\end{scriptsize}

Two semantic fusion backends are supported:
\begin{itemize}
\item \texttt{VOXEL\_SEMANTIC\_GRID}: confidence-counter fusion (majority voting).
\item \texttt{VOXEL\_SEMANTIC\_PROBABILISTIC\_GRID}: probabilistic fusion, with per-class distributions fused in log-space.
\end{itemize}

Configuration notes:
\begin{itemize}
\item If \texttt{kDoSparseSemanticMappingAndSegmentation=True} and \texttt{kVolumetricIntegrationType="VOXEL\_GRID"}, the factory warns and automatically switches to \texttt{VOXEL\_SEMANTIC\_PROBABILISTIC\_GRID} to enable semantic fusion.
\item If \texttt{kDoSparseSemanticMappingAndSegmentation=True} with a non-semantic volumetric integrator (e.g., \texttt{TSDF}, \texttt{GAUSSIAN\_SPLATTING}), a warning is emitted and semantic integration is skipped.
\end{itemize}

If \texttt{kVolumetricSemanticIntegrationUseInstanceIds=True}, the volumetric integrator uses 2D instance IDs (from panoptic/instance segmentation backends) to build \textbf{3D object segments}. Object IDs are assigned via voting across observations and filtered by \texttt{kVolumetricSemanticIntegrationMinVoteRatio} and \texttt{kVolumetricSemanticIntegrationMinVotes}. This enables:
\begin{itemize}
\item Consistent 3D object grouping across views.
\item Per-object colorization of the volumetric map.
\item Improved object identity stability over time.
\end{itemize}

When volumetric semantic mapping is enabled, use the following GUI buttons:
\begin{itemize}
\item \texttt{Color Semantics}: class/label color maps on both the 3D sparse map and the volumetric map.
\item \texttt{Objects}: per-object color map on the volumetric map (requires instance IDs).
\item \texttt{Draw Object BBs}: bounding boxes for detected 3D object segments.
\end{itemize}

Known limitations in volumetric semantic mapping:
\begin{itemize}
\item \textit{Resolution limits}: a fixed voxel size may blur thin structures and small objects.
\item \textit{Pose drift sensitivity}: accumulated pose errors can smear semantic labels over time.
\item \textit{Instance ID dependence}: inconsistent or noisy 2D instance IDs may lead to fragmented or unstable 3D objects.
\item \textit{Class imbalance bias}: frequently observed classes may dominate voting/probabilistic fusion.
\item \textit{Temporal class inconsistency}: label flickering can fragment voxel labeling over time.
\item \textit{Occlusions and visibility gaps}: poorly observed regions may remain unlabeled or misclassified.
\end{itemize}

Further information about the \textbf{volumetric integration models and
SW architecture} are available \href{https://github.com/luigifreda/pyslam/blob/master/cpp/modules/volumetric/README.md}{here}.


\hypertarget{saving-and-reloading}{%
\subsection{Saving and reloading}\label{saving-and-reloading}}

\hypertarget{save-the-a-map}{%
\subsubsection{Save a map}\label{save-the-a-map}}

When you run the script \texttt{main\_slam.py}
(\texttt{main\_map\_dense\_reconstruction.py}): 
\begin{itemize}
\item You can save the
current map state by pressing the button \texttt{Save} on the GUI. This
saves the current map along with front-end, and backend configurations
into the default folder \texttt{results/slam\_state}
(\texttt{results/slam\_state\_dense\_reconstruction}). 
\item To change the
default saving path, open \texttt{config.yaml} and update target
\texttt{folder\_path} in the section:
\begin{scriptsize}
\begin{Shaded}
\begin{Highlighting}[]
\ExtensionTok{SYSTEM_STATE}\NormalTok{:}
  \ExtensionTok{folder_path}\NormalTok{: results/slam_state   }\CommentTok{# default folder path (relative to repository root) where the}
  \ExtensionTok{           }\NormalTok{                       }\CommentTok{# system state is saved or reloaded}
\end{Highlighting}
\end{Shaded}
\end{scriptsize}
\end{itemize}

\hypertarget{reload-a-saved-map-and-relocalize-in-it}{%
\subsubsection{Reload a saved map and relocalize in
it}\label{reload-a-saved-map-and-relocalize-in-it}}

\begin{itemize}
\item
  A saved map can be loaded and visualized in the GUI by running: \\

\begin{scriptsize}
\begin{Shaded}
\begin{Highlighting}[]
\NormalTok{$ }\BuiltInTok{.} \ExtensionTok{pyenv-activate.sh}   \CommentTok{#  Activate pyslam python virtual environment. This is only needed once in a new terminal.}
\NormalTok{$ }\ExtensionTok{./main_map_viewer.py}  \CommentTok{#  Use the --path options to change the input path}
\end{Highlighting}
\end{Shaded}

\end{scriptsize}

\item
  To enable map reloading and relocalization when running
  \texttt{main\_slam.py}, open \texttt{config.yaml} and set \\

\begin{scriptsize}
\begin{Shaded}
\begin{Highlighting}[]
\ExtensionTok{SYSTEM_STATE}\NormalTok{:}
  \ExtensionTok{load_state}\NormalTok{: True                  }\CommentTok{# flag to enable SLAM state reloading (map state + loop closing state)}
  \ExtensionTok{folder_path}\NormalTok{: results/slam_state   }\CommentTok{# default folder path (relative to repository root) where the}
  \ExtensionTok{           }\NormalTok{                       }\CommentTok{# system state is saved or reloaded}  
\end{Highlighting}
\end{Shaded}
\end{scriptsize}

\end{itemize}

Note that pressing the \texttt{Save} button saves the current map,
front-end, and backend configurations. Reloading a saved map replaces
the current system configurations to ensure descriptor compatibility.

\hypertarget{trajectory-saving}{%
\subsubsection{Trajectory saving}\label{trajectory-saving}}

Estimated trajectories can be saved in three formats:
\emph{TUM} (The Open Mapping format), \emph{KITTI} (KITTI Odometry
format), and \emph{EuRoC} (EuRoC MAV format). pySLAM saves two
\textbf{types} of trajectory estimates:

\begin{itemize}
\tightlist
\item
  \textbf{Online}: In \emph{online} trajectories, each pose estimate
  depends only on past poses. A pose estimate is saved at the end of
  each front-end iteration for the current frame.
\item
  \textbf{Final}: In \emph{final} trajectories, each pose estimate
  depends on both past and future poses. A pose estimate is refined
  multiple times by LBA windows that include it, as well as by PGO and GBA during
  loop closures.
\end{itemize}

To enable trajectory saving, open \texttt{config.yaml} and search for
the \texttt{SAVE\_TRAJECTORY}: set \texttt{save\_trajectory:\ True},
select your \texttt{format\_type} (\texttt{tum}, \texttt{kitti},
\texttt{euroc}), and the output filename. For instance for a
\texttt{tum} format output: \\

\begin{scriptsize}
\begin{Shaded}
\begin{Highlighting}[]
\ExtensionTok{SAVE_TRAJECTORY}\NormalTok{:}
  \ExtensionTok{save_trajectory}\NormalTok{: True}
  \ExtensionTok{format_type}\NormalTok{: kitti             }\CommentTok{# supported formats: `tum`, `kitti`, `euroc`}
  \ExtensionTok{output_folder}\NormalTok{: results/metrics }\CommentTok{# relative to pyslam root folder }
  \ExtensionTok{basename}\NormalTok{: trajectory           }\CommentTok{# basename of the trajectory saving output}
\end{Highlighting}
\end{Shaded}
\end{scriptsize}

\hypertarget{optimization-engines}{%
\subsubsection{Optimization engines}\label{optimization-engines}}

Currently, pySLAM supports both \texttt{g2o} and \texttt{gtsam} for
graph optimization, with \texttt{g2o} set as the default engine. You can enable \texttt{gtsam} by setting to \texttt{True} the
following parameters in \texttt{config\_parameters.py}:

\begin{scriptsize}
  \begin{Shaded}
    \begin{Highlighting}[]
      \CommentTok{# Optimization engine }
    \NormalTok{  kOptimizationFrontEndUseGtsam }\OperatorTok{=} \VariableTok{True}    
    \NormalTok{  kOptimizationBundleAdjustUseGtsam }\OperatorTok{=} \VariableTok{True} 
    \NormalTok{  kOptimizationLoopClosingUseGtsam }\OperatorTok{=} \VariableTok{True} 
    \end{Highlighting}
    \end{Shaded}
\end{scriptsize}

Additionally, the \texttt{gtsam\_factors} package provides custom Python
bindings for features not available in the original gtsam framework. See
\href{https://github.com/luigifreda/pyslam/blob/master/thirdparty/gtsam_factors/README.md}{here} for further details.

\hypertarget{slam-gui}{%
\subsection{SLAM GUI}\label{slam-gui}}

Some quick information about the non-trivial GUI buttons of \texttt{main\_slam.py}:

\begin{itemize}
    \item \texttt{Step}: Enter the \emph{Step by step mode}. Press the button \texttt{Step} a first time to pause. Then, press it again to make the pipeline process a single new frame.

    \item \texttt{Save}: Save the map into the file \texttt{map.json}. You can visualize it back by using the script \texttt{main\_map\_viewer.py} (as explained above).

    \item \texttt{Reset}: Reset SLAM system.

    \item \texttt{Draw Ground Truth}: If a ground truth dataset (e.g., KITTI,
    TUM, EUROC, or REPLICA) is loaded, you can visualize it by pressing this
    button. The ground truth trajectory will be displayed in 3D and will be
    progressively aligned with the estimated trajectory, updating
    approximately every 10-30 frames. As more frames are processed, the
    alignment between the ground truth and estimated trajectory becomes more
    accurate. After about 20 frames, if the button is pressed, a window will
    appear showing the Cartesian alignment errors along the main axes (i.e.,
    $e_x, e_y, e_z$ and the history of the total $RMSE$ between
    the ground truth and the aligned estimated trajectories.
\end{itemize}

\hypertarget{monitor-the-logs-for-tracking-local-mapping-and-loop-closing-simultaneously}{%
\subsection{Monitor the logs for tracking, local mapping, and loop
closing
simultaneously}\label{monitor-the-logs-for-tracking-local-mapping-and-loop-closing-simultaneously}}

The logs generated by the modules \texttt{local\_mapping.py},
\texttt{loop\_closing.py}, \texttt{loop\_detecting\_process.py},
\texttt{global\_bundle\_adjustments.py}, and
\texttt{volumetric\ integrator\_\textless{}X\textgreater{}.py} are
collected in the files \texttt{local\_mapping.log},
\texttt{loop\_closing.log}, \texttt{loop\_detecting.log},
\texttt{gba.log}, and \texttt{volumetric\_integrator.log}, which are all
stored in the folder \texttt{logs}. 

For debugging, you can monitor one
of the them in parallel by running the following command in a separate
shell:\\
\begin{codeblock}
\codeline{tail\ -f\ logs/\textless{}log\ file\ name\textgreater{}}
\end{codeblock}
Otherwise, to check all parallel logs with tmux, run:\\
\begin{codeblock}
\codeline{./scripts/launch\_tmux\_logs.sh}
\end{codeblock}
To launch slam and check all logs in a single tmux, run:\\
\begin{codeblock}
\codeline{./scripts/launch\_tmux\_slam.sh}
\end{codeblock}
Press \texttt{CTRL+A} followed by \texttt{CTRL+Q} to exit from
\texttt{tmux} environment.

\hypertarget{scene-from-views-usage}{%
\subsection{Scene from Views}\label{scene-from-views-usage}}

The \textit{Scene from Views} module provides a unified interface for feed-forward 3D scene reconstruction from multiple
views, following a modular factory pattern architecture. The pipeline
provides a consistent API across different reconstruction models while
maintaining model-specific optimizations. See the main script
\texttt{main\_scene\_from\_views.py}.

\begin{itemize}
\tightlist
\item
  Feed-forward multi-view reconstruction (poses + fused 3D geometry
  directly from images): \texttt{SceneFromViewsDust3r},
  \texttt{SceneFromViewsMast3r}, \texttt{SceneFromViewsMvdust3r},
  \texttt{SceneFromViewsVggt}, \texttt{SceneFromViewsVggtRobust}, \texttt{SceneFromViewsFast3r}, \texttt{SceneFromViewsSolv3r}.
\item
  Views added over time: \texttt{create\_session("incremental")} or
  \texttt{create\_session("streaming")} on any reconstructor
  (native in \texttt{SceneFromViewsSolv3r}, generic for the other models)
\item
  Single-view depth-first pipeline with optional poses/intrinsics:
  \texttt{SceneFromViewsDepthAnythingV3}
\item
  Global alignment optimization stage for merging views:
  \texttt{SceneFromViewsDust3r} (dense alignment) and
  \texttt{SceneFromViewsMast3r} (sparse alignment variant)
\item
  Robust view filtering / outlier rejection:
  \texttt{SceneFromViewsVggtRobust} (anchor-based attention + cosine
  scoring that discards low-confidence views before reconstruction)
\end{itemize}

Note that \texttt{DUSt3R} and \texttt{MASt3R} are pairwise models: they
take two images at a time. They are feed-forward per pair; multi-view
reconstruction is then obtained by global alignment / optimization over
all pairwise pointmaps. \texttt{MV-DUSt3R}, \texttt{VGGT}, and
\texttt{Fast3R} are multi-view models that process all images
simultaneously in a single forward pass. \texttt{SOLV3R} is a multi-view
solver that refines one state per view over a number of solver calls.

All models return standardized \texttt{SceneFromViewsResult} output
containing point clouds, meshes, camera poses, depth maps, and
intrinsics in a consistent format. The factory pattern allows easy
switching between models while maintaining the same interface.

Further details \href{https://github.com/luigifreda/pyslam/blob/master/pyslam/scene_from_views/README.md}{here}.

\hypertarget{configuration}{%
\section{Configuration}\label{configuration}}

\hypertarget{main-configuration-file}{%
\subsection{Main configuration file}\label{main-configuration-file}}

Refer to
\protect\hyperlink{selecting-a-dataset-and-different-configuration-parameters}{this
section} for how to update the main configuration file
\href{https://github.com/luigifreda/pyslam/blob/master/config.yaml}{config.yaml} and affect the configuration
parameters in \href{https://github.com/luigifreda/pyslam/blob/master/pyslam/config_parameters.py}{config\_parameters.py}.

\hypertarget{datasets}{%
\subsection{Datasets}\label{datasets}}

The following datasets are supported:

\begin{longtable}[]{@{}ll@{}}
\toprule
Dataset & type in \texttt{config.yaml}\tabularnewline
\midrule
\endhead
\href{http://www.cvlibs.net/datasets/kitti/eval_odometry.php}{KITTI
odometry data set} \cite{Geiger2012CVPR} &
\texttt{type:\ KITTI\_DATASET} \tabularnewline
\href{https://vision.in.tum.de/data/datasets/rgbd-dataset/download}{TUM
dataset} \cite{sturm12iros} & \texttt{type:\ TUM\_DATASET} \tabularnewline
\href{https://cvg.cit.tum.de/data/datasets/visual-inertial-dataset}{TUM-VI
dataset} \cite{schubert2018tumvi} & \texttt{type:\ TUM\_VI\_DATASET} \tabularnewline
\href{https://www.doc.ic.ac.uk/~ahanda/VaFRIC/iclnuim.html}{ICL-NUIM
dataset} \cite{handa:etal:ICRA2014} & \texttt{type:\ ICL\_NUIM\_DATASET}\tabularnewline
\href{https://projects.asl.ethz.ch/datasets/euroc-mav/}{EUROC
dataset} \cite{Burri25012016} & \texttt{type:\ EUROC\_DATASET}\tabularnewline
\href{https://github.com/facebookresearch/Replica-Dataset}{REPLICA
dataset} \cite{replica19arxiv} & \texttt{type:\ REPLICA\_DATASET}\tabularnewline
\href{https://theairlab.org/tartanair-dataset/}{TARTANAIR dataset} \cite{tartanair2020iros} &
\texttt{type:\ TARTANAIR\_DATASET}\tabularnewline
\href{https://www.microsoft.com/en-us/research/project/rgb-d-dataset-7-scenes/}{Seven Scenes dataset} \cite{seven_scenes} &
\texttt{type:\ SEVEN\_SCENES\_DATASET}\tabularnewline
\href{https://github.com/dazinovic/neural-rgbd-surface-reconstruction}{Neural RGBD dataset} \cite{Azinovic_2022_CVPR} &
\texttt{type:\ NEURAL\_RGBD\_DATASET}\tabularnewline
\href{https://iis-esslingen.github.io/rover/}{Rover dataset} \cite{schmidt2025rover} &
\texttt{type:\ ROVER\_DATASET}\tabularnewline
\href{http://www.scan-net.org/}{ScanNet dataset} \cite{dai2017scannet} &
\texttt{type:\ SCANNET\_DATASET}\tabularnewline
\href{https://github.com/MIT-SPARK/Clio}{CLIO dataset} \cite{kim2024clio} &
\texttt{type:\ CLIO\_DATASET}\tabularnewline
\href{https://wiki.ros.org/Bags}{ROS1 bags} &
\texttt{type:\ ROS1BAG\_DATASET}\tabularnewline
\href{https://docs.ros.org/en/foxy/Tutorials/Beginner-CLI-Tools/Recording-And-Playing-Back-Data/Recording-And-Playing-Back-Data.html}{ROS2
bags} & \texttt{type:\ ROS2BAG\_DATASET}\tabularnewline
\href{https://foxglove.dev/blog/introducing-the-mcap-file-format}{MCAP
file} & \texttt{type:\ MCAP\_DATASET}\tabularnewline
Video file & \texttt{type:\ VIDEO\_DATASET}\tabularnewline
Folder of images & \texttt{type:\ FOLDER\_DATASET}\tabularnewline
\bottomrule
\end{longtable}

Use the download scripts available in the folder \texttt{scripts} to
download some of the following datasets.

File-based datasets prefetch decoded images in a background thread, using the optional C++ \texttt{data\_thread\_reader} module, so that image decoding overlaps with tracking. Prefetching can be disabled with the environment variable \texttt{PYSLAM\_DISABLE\_IMAGE\_PREFETCH=1}. Supported readers, build instructions, and memory behavior are described \href{https://github.com/luigifreda/pyslam/blob/master/cpp/modules/data_thread_reader/README.md}{here}.

\hypertarget{kitti-datasets}{%
\subsubsection{KITTI Datasets}\label{kitti-datasets}}

pySLAM code expects the following structure in the specified KITTI path
folder (specified in the section \texttt{KITTI\_DATASET} of the file
\texttt{config.yaml}). :

\begin{scriptsize}
\begin{Shaded}
\begin{Highlighting}[]
\NormalTok{+-- sequences}
\NormalTok{|   +-- 00}
\NormalTok{|   ...}
\NormalTok{|   +-- 21}
\NormalTok{+-- poses}
\NormalTok{    +-- 00.txt}
\NormalTok{    ...}
\NormalTok{    +-- 10.txt}
\end{Highlighting}
\end{Shaded}
\end{scriptsize}

\begin{enumerate}
\def\labelenumi{\arabic{enumi}.}
\item
  Download the dataset (grayscale images) from
  http://www.cvlibs.net/datasets/kitti/eval\_odometry.php and prepare
  the KITTI folder as specified above
\item
  Select the corresponding calibration settings file (section
  \texttt{KITTI\_DATASET:\ settings:} in the file \texttt{config.yaml})
\end{enumerate}

\hypertarget{tum-datasets}{%
\subsubsection{TUM Datasets}\label{tum-datasets}}

pySLAM code expects a file \texttt{associations.txt} in each TUM dataset
folder (specified in the section \texttt{TUM\_DATASET:} of the file
\texttt{config.yaml}).

\begin{enumerate}
\def\labelenumi{\arabic{enumi}.}
\tightlist
\item
  Download a sequence from
  \href{http://vision.in.tum.de/data/datasets/rgbd-dataset/download}{vision.in.tum.de/data/datasets/rgbd-dataset/download} and
  uncompress it.
\item
  Associate RGB images and depth images using the python script
  \href{http://vision.in.tum.de/data/datasets/rgbd-dataset/tools}{associate.py}.
  You can generate your \texttt{associations.txt} file by executing:\\
  \texttt{\$\ python\ associate.py\ PATH\_TO\_SEQUENCE/rgb.txt\ PATH\_TO\_SEQUENCE/depth.txt\ \textgreater{}\ associations.txt\ \ \ \ \ \ \#\ pay\ attention\ to\ the\ order!}
\item
  Select the corresponding calibration settings file (section
  \texttt{TUM\_DATASET:\ settings:} in the file \texttt{config.yaml}).
\end{enumerate}

\hypertarget{tum-vi-datasets}{%
\subsubsection{TUM-VI Datasets}\label{tum-vi-datasets}}

pySLAM reads the 512$\times$512 image sequences of TUM-VI~\cite{schubert2018tumvi}; IMU measurements are not used. Each sequence folder (the \texttt{name} field of the section \texttt{TUM\_VI\_DATASET} in \texttt{config.yaml}) is expected to contain \texttt{mav0/cam0/data.csv}, \texttt{mav0/cam0/data/<timestamp>.png}, the same two entries for \texttt{cam1}, and \texttt{mav0/mocap0/data.csv}.

\begin{enumerate}
\def\labelenumi{\arabic{enumi}.}
\tightlist
\item
  Download a 512$\times$512 sequence from
  \href{https://cvg.cit.tum.de/data/datasets/visual-inertial-dataset}{cvg.cit.tum.de/data/datasets/visual-inertial-dataset}
  and uncompress it under \texttt{TUM\_VI\_DATASET:\ base\_path}.
\item
  Set \texttt{DATASET.type:\ TUM\_VI\_DATASET}. Set \texttt{sensor\_type} to \texttt{stereo} or \texttt{mono}, and \texttt{stereo\_mode} to \texttt{native} (original Kannala--Brandt images) or \texttt{rectified} (fisheye images remapped to a pinhole stereo pair). For monocular fisheye tracking use \texttt{sensor\_type:\ mono} with \texttt{stereo\_mode:\ native}. See Sect.~\ref{camera-models} for the differences between the two modes.
\item
  Select \texttt{settings/TUM\_VI.yaml}. With \texttt{groundtruth\_file:\ auto}, the motion-capture poses are converted to the tracking camera frame. Calibration, rectification, and ground-truth conversion are detailed \href{https://github.com/luigifreda/pyslam/blob/master/docs/tum-vi.md}{here}.
\end{enumerate}

\hypertarget{icl-nuim-datasets}{%
\subsubsection{ICL-NUIM Datasets}\label{icl-nuim-datasets}}

Follow the same instructions provided for the TUM datasets.

\hypertarget{euroc-datasets}{%
\subsubsection{EuRoC Datasets}\label{euroc-datasets}}

\begin{enumerate}
\def\labelenumi{\arabic{enumi}.}
\tightlist
\item
  Download a sequence (ASL format) from
  http://projects.asl.ethz.ch/datasets/doku.php?id=kmavvisualinertialdatasets
  (check this direct
  \href{http://robotics.ethz.ch/~asl-datasets/ijrr_euroc_mav_dataset/}{link})
\item
  Use the script \texttt{io/generate\_euroc\_groundtruths\_as\_tum.sh}
  to generate the TUM-like groundtruth files
  \texttt{path\ +\ \textquotesingle{}/\textquotesingle{}\ +\ name\ +\ \textquotesingle{}/mav0/state\_groundtruth\_estimate0/data.tum\textquotesingle{}}
  that are required by the \texttt{EurocGroundTruth} class.
\item
  Select the corresponding calibration settings file (section
  \texttt{EUROC\_DATASET:\ settings:} in the file \texttt{config.yaml}).
\end{enumerate}

\hypertarget{replica-datasets}{%
\subsubsection{Replica Datasets}\label{replica-datasets}}

\begin{enumerate}
\def\labelenumi{\arabic{enumi}.}
\tightlist
\item
  You can download the zip file containing all the sequences by
  running:\\
  \texttt{\$\ wget\ https://cvg-data.inf.ethz.ch/nice-slam/data/Replica.zip}~\\
\item
  Then, uncompress it and deploy the files as you wish.
\item
  Select the corresponding calibration settings file (section
  \texttt{REPLICA\_DATASET:\ settings:} in the file
  \texttt{config.yaml}).
\end{enumerate}

\hypertarget{tartanair-datasets}{%
\subsubsection{Tartanair Datasets}\label{tartanair-datasets}}

\begin{enumerate}
\def\labelenumi{\arabic{enumi}.}
\tightlist
\item
  You can download the datasets from
  \href{https://theairlab.org/tartanair-dataset}{https://theairlab.org/tartanair-dataset}
\item
  Then, uncompress them and deploy the files as you wish.
\item
  Select the corresponding calibration settings file (section
  \texttt{TARTANAIR\_DATASET:\ settings:} in the file
  \texttt{config.yaml}).
\end{enumerate}

\hypertarget{scenes-dataset}{%
\subsubsection{7 Scenes dataset}\label{scenes-dataset}}

To dowload the dataset follow the instructions at
https://www.microsoft.com/en-us/research/project/rgb-d-dataset-7-scenes/.
Select the section \texttt{SEVEN\_SCENES\_DATASET:\ settings:} in the
file \texttt{config.yaml}.A calibration file
\texttt{settings/SEVEN\_SCENES.yaml} is already available.

\hypertarget{neural-rgbd-dataset}{%
\subsubsection{Neural RGBD dataset}\label{neural-rgbd-dataset}}

To dowload the dataset follow the instructions at
https://github.com/dazinovic/neural-rgbd-surface-reconstruction. Select
the section \texttt{NEURAL\_RGBD\_DATASET:\ settings:} in the file
\texttt{config.yaml}. A calibration file
\texttt{settings/NEURAL\_RGBD.yaml} is already available.

\hypertarget{rover-dataset}{%
\subsubsection{Rover dataset}\label{rover-dataset}}

To dowload the dataset follow the instructions at
https://iis-esslingen.github.io/rover/. Select the section
\texttt{ROVER\_DATASET:\ settings:} in the file \texttt{config.yaml}. A
calibration file for the \texttt{d435i} camera is already available:
\texttt{settings/ROVER\_d435i.yaml}. Other camera models will be
supported in the future.

\hypertarget{scannet-datasets}{%
\subsubsection{ScanNet Datasets}\label{scannet-datasets}}

\begin{enumerate}
\def\labelenumi{\arabic{enumi}.}
\tightlist
\item
  You can download the datasets following instructions in
  \href{http://www.scan-net.org}{http://www.scan-net.org}. You will need to request the dataset from
  the authors.
\item
  There are two versions you can download:
\begin{itemize}
\tightlist
\item
  A subset of pre-processed data termed as
  \texttt{tasks/scannet\_frames\_2k}: this version is smaller, and more
  generally available for training neural networks. However, it only
  includes one frame out of each 100, which makes it unusable for SLAM.
  The labels are processed by mapping them from the original Scannet
  label annotations to NYU40.
\item
  The raw data: this version is the one used for SLAM. You can download
  the whole dataset (TBs of data) or specific scenes. A common approach
  for evaluation of semantic mapping is to use the
  \texttt{scannetv2\_val.txt} scenes. For downloading and processing the
  data, you can use the following
  \href{https://github.com/dvdmc/scannet-processing}{repository} as the
  original Scannet repository is tested under Python 2.7 and does't
  support batch downloading of scenes.
\end{itemize}
\item
  Once you have the \texttt{color}, \texttt{depth}, \texttt{pose}, and
  (optional for semantic mapping) \texttt{label} folders, you should
  place them following
  \texttt{\{path\_to\_scannet\}/scans/\{scene\_name\}/{[}color,\ depth,\ pose,\ label{]}}.
  Then, configure the \texttt{base\_path} and \texttt{name} in the file
  \texttt{config.yaml}.
\item
  Select the corresponding calibration settings file (section
  \texttt{SCANNET\_DATASET:\ settings:} in the file
  \texttt{config.yaml}). NOTE: the RGB images are rescaled to match the
  depth image. The current intrinsic parametes in the existing
  calibration file reflect that.
\end{enumerate}

\hypertarget{clio-datasets}{%
\subsubsection{CLIO Datasets}\label{clio-datasets}}

pySLAM supports the RGB-D sequences released with
\href{https://github.com/MIT-SPARK/Clio}{Clio} \cite{kim2024clio}
(Hydra/Clio scene-graph mapping). Each scene is expected to follow this
layout under \texttt{\{base\_path\}/\{name\}/}:

\begin{scriptsize}
\begin{Shaded}
\begin{Highlighting}[]
\NormalTok{images/rgb\_\{id\}.jpg}
\NormalTok{depth/depth\_\{id\}.png}
\NormalTok{sparse/0/images.bin          \# COLMAP sparse model (optional reference poses)}
\NormalTok{*.bag                        \# ROS1 recording (fps auto-detection, metric odometry)}
\end{Highlighting}
\end{Shaded}
\end{scriptsize}

\begin{enumerate}
\def\labelenumi{\arabic{enumi}.}
\tightlist
\item
  Download or prepare a CLIO scene and set \texttt{base\_path} and
  \texttt{name} in the \texttt{CLIO\_DATASET} section of
  \texttt{config.yaml}.
\item
  Select the calibration settings file (\texttt{CLIO\_DATASET:\ settings:\
  settings/CLIO.yaml}). Depth maps are stored as \texttt{uint16} in
  \textbf{millimeters} (\texttt{DepthMapFactor:\ 1000.0}).
\item
  Set \texttt{sensor\_type:\ rgbd} (or \texttt{mono} if depth is
  unavailable).
\item
  Set \texttt{groundtruth\_file:\ auto} to load a \textbf{reference
  trajectory} for evaluation and visualization (not independent
  motion-capture GT). By default, pySLAM prefers metric poses from the
  scene ROS1 bag topic \texttt{/dominic/forward/colmap\_odom}; if the
  bag is missing, it falls back to COLMAP \texttt{sparse/0} poses
  (typically up-to-scale). Use \texttt{clio\_reference\_poses:\ auto\ |\ bag\ |\ sparse}
  to override this choice.
\item
  Playback rate is auto-detected from the bag when present
  (${\sim}7.5$~Hz on the office scene); you can override it with
  \texttt{fps} in \texttt{config.yaml} or \texttt{Camera.fps} in the
  settings file.
\end{enumerate}

When the reference trajectory is non-metric (COLMAP sparse), pySLAM
automatically uses Sim(3) alignment for the ground-truth overlay and ATE
metrics.

\hypertarget{ros1-bags}{%
\subsubsection{ROS1 bags}\label{ros1-bags}}

\begin{enumerate}
\def\labelenumi{\arabic{enumi}.}
\tightlist
\item
  Source the main ROS1 \texttt{setup.bash} after you have sourced the
  \texttt{pyslam} python environment.
\item
  Set the paths and \texttt{ROS1BAG\_DATASET:\ ros\_parameters} in the
  file \texttt{config.yaml}.
\item
  Select/prepare the correspoding calibration settings file (section
  \texttt{ROS1BAG\_DATASET:\ settings:} in the file
  \texttt{config.yaml}). See the available yaml files in the folder
  \texttt{Settings} as an example.
\end{enumerate}

\hypertarget{ros2-bags}{%
\subsubsection{ROS2 bags}\label{ros2-bags}}

\begin{enumerate}
\def\labelenumi{\arabic{enumi}.}
\tightlist
\item
  Source the main ROS2 \texttt{setup.bash} after you have sourced the
  \texttt{pyslam} python environment.
\item
  Set the paths and \texttt{ROS2BAG\_DATASET:\ ros\_parameters} in the
  file \texttt{config.yaml}.
\item
  Select/prepare the correspoding calibration settings file (section
  \texttt{ROS2BAG\_DATASET:\ settings:} in the file
  \texttt{config.yaml}). See the available yaml files in the folder
  \texttt{Settings} as an example.
\end{enumerate}

\hypertarget{mcap-file}{%
\subsubsection{MCAP file}\label{mcap-file}}

A concise introduction of the MCAP format is available at \href{https://foxglove.dev/blog/introducing-the-mcap-file-format}{https://foxglove.dev/blog/introducing-the-mcap-file-format}. These files can be also recorded and played back by using ROS2 tools.  

\hypertarget{video-and-folder-datasets}{%
\subsubsection{Video and Folder Datasets}\label{video-and-folder-datasets}}

You can use the \texttt{VIDEO\_DATASET} and \texttt{FOLDER\_DATASET}
types to read generic video files and image folders (specifying a glob
pattern), respectively. A companion ground truth file can be set in the
simple format type: Refer to the class \texttt{SimpleGroundTruth} in
\texttt{io/ground\_truth.py} and check the script
\texttt{io/convert\_groundtruth\_to\_simple.py}.
\hypertarget{camera-settings}{%
\subsection{Camera Settings}\label{camera-settings}}

The folder \texttt{settings} contains the camera settings files which
can be used for testing the code. These are the same used in the
framework \href{https://github.com/raulmur/ORB_SLAM2}{ORB-SLAM2} \cite{ORB_SLAM2}. You
can easily modify one of those files for creating your own new
calibration file (for your new datasets).

A settings file that does not set \texttt{Camera.type} uses the linear pinhole model: \texttt{Camera.k1}, \ldots, \texttt{Camera.k3} are used only to undistort keypoints. Setting \texttt{Camera.type} to \texttt{kannala\_brandt4}, \texttt{pinhole\_radtan8}, or \texttt{double\_sphere} selects the corresponding model of Sect.~\ref{camera-models}. Note that \texttt{Camera.k3} and \texttt{Camera.k4} are the higher-order polynomial coefficients in a Kannala--Brandt file, not the tangential coefficients. A non-pinhole stereo rig requires an explicit right camera (\texttt{Stereo.right}) and the relative pose \texttt{Stereo.T\_right\_left}. Parameters and pixel conventions of each model are detailed \href{https://github.com/luigifreda/pyslam/blob/master/cpp/modules/camera_models/README.md}{here}.

In order to calibrate your camera, you can use the scripts in the folder
\texttt{calibration}. In particular: 1. Use the script
\texttt{grab\_chessboard\_images.py} to collect a sequence of images
where the chessboard can be detected (set the chessboard size therein,
you can use the calibration pattern \texttt{calib\_pattern.pdf} in the
same folder) 2. Use the script \texttt{calibrate.py} to process the
collected images and compute the calibration parameters (set the
chessboard size therein)

For more information on the calibration process, see this
\href{https://learnopencv.com/camera-calibration-using-opencv/}{tutorial} \cite{learnopencv}
or this other
\href{https://docs.opencv.org/4.x/dc/dbb/tutorial_py_calibration.html}{link} \cite{opencv}.

If you want to \textbf{use your camera}, you have to: 
\begin{itemize}
\item Calibrate it and configure \href{https://github.com/luigifreda/pyslam/blob/master/settings/WEBCAM.yaml}{settings/WEBCAM.yaml} accordingly.
\item Record a video (for instance, by using \texttt{save\_video.py} in the folder \texttt{calibration}). 
\item Configure the \texttt{VIDEO\_DATASET} section of \texttt{config.yaml} in order to point to your recorded video.
\end{itemize}

\clearpage
\hypertarget{visual-models-appendix}{%
\section{Visual Models: Installation and Runtime Options}\label{visual-models-appendix}}

The visual models of Sect.~\ref{supported-components-and-models} are described by the catalog of Sect.~\ref{visual-model-catalog}. The generated catalog table, with dependency groups and licenses, and all the commands below are in \texttt{docs/visual\_features.md}.

\subsection{Installing Models}\label{visual-model-install}

Classic features need no installation beyond the standard build, and most learned integrations are installed by the standard install scripts. Models integrated through pinned sources (DeDoDe, RDD, RoMa, tiny-RoMa, SALAD, BoQ, MegaLoc, and the LoFTR weights) are prefetched with \texttt{./scripts/install\_visual\_model.sh}, or downloaded on first use. Installing a model constructs it once on the CPU, which fetches the pinned source snapshot and all its checkpoints into the standard caches, and then verifies every registered SHA-256 checksum. No package is installed and no pin changes; afterwards, the models also run offline (\texttt{PYSLAM\_OFFLINE=1}).

\subsection{Runtime}\label{visual-model-runtime}

The feature manager and feature matcher import learned and vendor wrappers (and kornia, LightGlue, XFeat, TensorFlow) only when the selected configuration constructs them; OpenCV and ORB2 features are always available. Compared with importing every wrapper eagerly, importing the feature stack takes about 0.3\,s instead of 4\,s, and benchmark process startup about 1.4\,s instead of 7.3\,s, with half the peak memory; outputs are bit-identical for the deterministic presets. On Linux, \texttt{pyslam/config.py} also sets glibc's heap trim and mmap thresholds, so that per-frame image buffers are reused instead of being returned to the kernel and page-faulted back on every frame (about 17k faults per SIFT call at 640$\times$480 with glibc defaults); \texttt{PYSLAM\_MALLOC\_TUNING=0} restores the defaults. Torch-based learned detectors run inference on one long-lived thread each. Stereo frames are extracted on threads created per frame, and PyTorch rebuilds its per-thread CUDA/cuDNN state on every call from a new thread: at KITTI resolution, ALIKED would take 45\,ms instead of 20\,ms per image and DeDoDe 2.8\,s instead of 175\,ms. With the persistent thread, the outputs are unchanged and the KITTI stereo median tracking time with ALIKED is 90\,ms per frame instead of 197\,ms.

\subsection{Optional Accelerations}\label{visual-model-accelerations}

Three accelerations are opt-in and off by default. They are enabled with environment variables, or the matching fields in \texttt{pyslam/config\_parameters.py}, and can change matches slightly; those differences are accepted when the option is on. \texttt{PYSLAM\_GPU\_NN\_MATCHER=1} matches float L2 descriptors on the GPU with the same one-to-one ratio test as the C++ path (about 5--10$\times$ faster per pair). Rare near-ties can flip, and FLANN's approximate search becomes exact. It applies to the Python \texttt{FeatureMatcher} only. \texttt{PYSLAM\_LIGHTGLUE\_BATCHED=1} matches one keyframe against every candidate of \texttt{compute\_frame\_matches} in one forward. On the measured GPU it was 10--20\% slower than per-pair matching, and one of twelve test pairs differed by a single match. \texttt{PYSLAM\_TORCH\_COMPILE=1} compiles SuperPoint, XFeat, ALIKED and LightGlue (\texttt{PYSLAM\_TORCH\_COMPILE\_BACKEND} is \texttt{inductor} or \texttt{tensorrt}). Extraction stays within $\pm$5\%, the first call spends several seconds compiling, and keypoint order can change. The TensorRT backend needs \texttt{torch\_tensorrt}, which pySLAM does not install, and any failure falls back to eager mode. Set \texttt{TORCHINDUCTOR\_COMPILE\_THREADS=1} so that the inductor workers stay within memory.

\subsection{Licenses}\label{visual-model-licenses}

The catalog records the license of each model's upstream code and weights, with the license files in \texttt{thirdparty/} as evidence, and the generated table in \texttt{docs/visual\_features.md} reports it. Licenses should be checked before any commercial use or redistribution: SuperPoint, R2D2, LF-Net, and MASt3R are restricted to non-commercial or research use; ContextDesc, SOSNet, and the L2-Net weights come from repositories without a license; and SURF is a patented algorithm. SuperPoint's terms also apply when it is paired with LightGlue.

\end{document}